\pdfoutput=1

\documentclass[preprint,12pt]{elsarticle}

\makeatletter
\def\ps@pprintTitle{%
 \let\@oddhead\@empty
 \let\@evenhead\@empty
 \def\@oddfoot{}%
 \let\@evenfoot\@oddfoot}
\makeatother

\usepackage{amssymb}
\usepackage{amsmath,epsfig}
\usepackage{mathptmx}
\usepackage[ruled,vlined]{algorithm2e}
\usepackage{float}
\usepackage{graphicx}
\usepackage{color}
\usepackage{placeins}
\usepackage{float}
\usepackage{tabularx,colortbl}
\usepackage{mathrsfs,amsmath}
\usepackage{bm}
\usepackage{enumitem}
\usepackage[caption=false,font=footnotesize]{subfig}
\usepackage{flushend}
\usepackage{fontenc}
\usepackage{epstopdf}
\usepackage{multirow}
\usepackage{changepage}
\usepackage{floatpag}
\usepackage{url}
\usepackage[table]{xcolor}
\usepackage{color,soul}
\soulregister\cite7
\soulregister\ref7
\soulregister\pageref7

\biboptions{sort&compress}

\usepackage[british]{babel}
\usepackage[flushleft]{threeparttable}

\usepackage{comment}
\usepackage{graphicx,lipsum}
\usepackage{fancyhdr,floatpag}

\floatpagestyle{fancy}

\begin{document}

\begin{frontmatter}



\begin{itemize}[align=parleft, labelsep=1cm]

\item[\textbf{Citation}]{D. Temel and G. AlRegib, Perceptual image 
quality assessment through spectral analysis of error representations, Signal Processing: Image Communication, Volume 70, 2019, Pages 37-46,ISSN 0923-5965}

\item[\textbf{DOI}]{https://doi.org/10.1016/j.image.2018.09.005}

\item[\textbf{Review}]{Submitted 15 August 2017, Accepted 10 September 2018}

\item[\textbf{Code}]{https://ghassanalregib.com/publication-3mt-videos/}

\item[\textbf{Bib}]{@article\{Temel2019\_SPIC,\\
title = "Perceptual image quality assessment through spectral analysis of error representations",\\
journal = "Signal Processing: Image Communication",\\
volume = "70",\\
pages = "37 - 46",\\
year = "2019",\\
issn = "0923-5965",\\
doi = "https://doi.org/10.1016/j.image.2018.09.005",\\
url = "http://www.sciencedirect.com/science/article/pii/S0923596518308531",\\
author = "Dogancan Temel and Ghassan AlRegib",
keywords = "Full-reference image quality assessment, Visual system, Error spectrum, Spectral analysis, Color perception, Multi-resolution"\}\\
}

\item[\textbf{Contact}]{alregib@gatech.edu~~~~~~~\url{https://ghassanalregib.com/}\\dcantemel@gmail.com~~~~~~~\url{http://cantemel.com/}}
\end{itemize}
\thispagestyle{empty}
\newpage
\clearpage
\setcounter{page}{1}

\title{Perceptual Image Quality Assessment through Spectral Analysis of Error Representations }


\author{Dogancan Temel and Ghassan AlRegib}

\address{Center for Signal and Information Processing \\ School of Electrical and Computer Engineering \\ Georgia Institute of Technology \\ Atlanta, GA 30332}

\begin{abstract}
In this paper, we analyze the statistics of error signals to assess the perceived quality of images. Specifically, we focus on the magnitude spectrum of error images obtained from the difference of reference and distorted images. Analyzing spectral statistics over grayscale images partially models interference in spatial harmonic distortion exhibited by the visual system but it overlooks color information, selective and hierarchical nature of visual system. To overcome these shortcomings, we introduce an image quality assessment algorithm 
based on the \underline{\texttt{S}}pectral \underline{\texttt{U}}nderstanding of \underline{\texttt{M}}ulti-scale and \underline{\texttt{M}}ulti-channel \underline{\texttt{E}}rror \underline{\texttt{R}}epresentations, denoted as \underline{\texttt{SUMMER}}. We validate the quality assessment performance over $3$ databases with around $30$ distortion types. These distortion types are grouped into $7$ main categories as compression artifact, image noise, color artifact, communication error, blur, global and local distortions. In total, we benchmark the performance of $17$ algorithms along with the proposed algorithm using $5$ performance metrics that measure linearity, monotonicity, accuracy, and consistency. In addition to experiments with standard performance metrics, we analyze the distribution of objective and subjective scores with histogram difference metrics and scatter plots. Moreover, we analyze the classification performance of quality assessment algorithms along with their statistical significance tests. Based on our experiments, \texttt{SUMMER} significantly outperforms majority of the compared methods in all benchmark categories.

\end{abstract}

\end{frontmatter}


\section{Introduction}\label{sec:intro}
\label{sec:related}
Recently, online platforms have been dominated by images because of the advances in capturing, storage, streaming, and display technologies. In order for these platforms to support the uploaded content, images should be formatted. The formatting of images can be considered as an optimization problem whose cost function is an image quality assessment algorithm. These quality assessment algorithms are grouped into three main categories as full-reference, reduced-reference, and no-reference. Design of these algorithms usually rely on visual system characteristics because their objective is estimating subjective quality. The characteristics of the human visual system include frequency sensitivity, luminance sensitivity, and masking effects \cite{WL+95}. Sensitivity of a visual system with respect to spatial frequency characteristics is considered under frequency sensitivity, just-noticeable intensity difference is studied under luminance sensitivity, and decreasing visibility of a signal in the presence of other signals is considered under masking. In \cite{Hegazy2014}, the authors do not directly investigate these characteristics individually but they introduce a quality assessment algorithm (COHERENSI) that captures perceptually correlated information from the phase and harmonic analysis of error signals.

The harmonics analysis in \cite{Hegazy2014} is based on the gradients of the error signals. Consecutive Fourier transforms are applied to measure the chaotic behavior in gradient representations. In this manuscript, we directly focus on the error signals without calculating their gradients. Instead of applying consecutive transforms, we apply it only once to stay in the Fourier domain and focus solely on the magnitude information. If we reconstruct images without their phase information, they are usually unrecognizable. This is because phase information includes the location of the image features that are critical for reconstructing the original image in the spatial domain. However, in this study, we do not need to reconstruct the images in the spatial domain. We do not utilize the phase information in the Fourier domain, but we use the location information while obtaining the difference of reference and compared images pixel-wise in the spatial domain. We analyze magnitude spectrums over each color channel in the RGB color space for multiple scales and use frequency-based weights to align quality scores. The main contributions of this manuscript compared to the baseline study is six folds.

\begin{itemize}

\item We analyzed the magnitude spectrums of error signals obtained from natural images and show that there is a general relationship between the magnitude spectums of error images and degradation levels. 

\item We eliminated the requirement for fine-tuned parameters that were utilized for fusion of phase and harmonic analysis as well as the scaling fraction parameter used in the multi-scale calculation.

\item We extended the baseline spectral analysis method with multi-scale and multi-channel error representations along with frequency-based weights, which significantly outperforms majority of the compared methods in all benchmark categories. 

\item We enlarged the test set from $1,625$ images to $3,000$ images in the TID 2013 database and added the Multiply Distorted LIVE database to the validation, which includes simultaneously applied distortions. 

\item We increased the number of validation metrics from two to five along with statistical significance tests for correlation metrics. We analyzed the distribution of scores with scatter plots and histogram difference metrics. Based on the overall validation, we showed that \texttt{SUMMER} is consistently among the top performing  methods.

\item We measured the capability of quality assessment algorithms to ($i$) distinguish statistically different and similar pairs and to ($ii$) identify the higher quality image and the lower quality image. We showed that \texttt{SUMMER} significantly outperforms all other top performing algorithms in task ($i$) and all other than one algorithm in task (ii).  
\end{itemize}

We briefly discuss the related work in Section \ref{sec:related} and describe the baseline spectral analysis method in Section \ref{sec:summer_core}. We extend the baseline method with multiple scales, multiple color channels, and frequency-based weights in Section \ref{sec:summer}. We describe the experimental setup in Section \ref{sec:setup} and report the results in Section \ref{sec:results}. Finally, we conclude our work in Section \ref{sec:conclusion}.

\section{Related Work}
\label{sec:related}
An intuitive approach to assess the quality of an image is to measure fidelity, which can be performed by comparing the image with its distortion-free version, if available. Mean Square Error (MSE) and Peak Signal-to-Noise Ratio (PSNR) are commonly utilized examples of fidelity-based full-reference methods. Wang \textit{et al.}~\cite{Wang2004} showed that human perception is more consistent with structural similarity as opposed to MSE and PSNR. Structural similarity methods such as SSIM~\cite{Wang2004} were shown to be more correlated with human error perception. Spatial domain-based single-scale structural similarity was further extended to multi-scale (MS-SSIM) \cite{Wang2003}, complex domain (CW-SSIM) \cite{Sampat2009}, and  information-weighted (IW-SSIM) \cite{Wang2011} versions. Instead of focusing on the structural similarity, Ponomarenko \textit{et al.} developed a series of quality estimators~\cite{EA+06, PS+07, Ponomarenko2011} that are based on extending fidelity with visual system characteristics. 

Daly \cite{Daly1992} introduced a visual model denoted as visual difference predictor (VDP), which is based on amplitude nonlinearity, contrast sensitivity, and a hierarchy of detection mechanisms. Through these mechanisms, VDP tries to measure visible differences that are caused by physical differences. Zhang and Li~\cite{Zhang12} modeled suppression mechanisms by spectral residual (SR-SIM), which is calculated in the frequency domain. Damera \textit{et al.}~\cite{Venkata2000} proposed a degradation model denoted as NQM, which mimics the human visual system by considering contrast sensitivity, local luminance, contrast interaction between spatial frequencies, and contrast masking effects. Chandler and Hemami~\cite{Chandler2007} formulated visual masking and summation through wavelet-based models to weight the SNR map. Other methods based on the frequency domain  were also used to analyze human visual system properties including \cite{Sampat2009, NiL+08, NaL+12}. Zhang \textit{et al.} developed FSIM, which mimics low-level feature perception through phase congruency and gradient magnitude. The majority of existing methods including FSIM measure quality using grayscale images or intensity channels. Intensity channels are usually preferred over chroma because human visual system is more sensitive to changes in intensity compared to color \cite{Lambrecht2001}. However, color channels still include information that is not part of intensity channels. FSIM was extended by introducing color information through pixel-wise fidelity over chroma channels. Temel and AlRegib utilized color information in the proposed methods PerSIM ~\cite{temel_15_persim} and CSV ~\cite{temel_16_csv}. To introduce color information into quality assessment, PerSIM uses pixel-wise fidelity whereas CSV utilizes color difference equations and color name distances.

The aforementioned quality estimators are based on handcrafting quality attributes. However, data-driven approaches can also be used to obtain quality estimators \cite{Tang2011,Mittal2012,Moorthy2010,Saad2012,Temel_UNIQUE,Kang2014}
. Existing data-driven methods are commonly based on natural scene statistics, support vector machines, various types of neural networks, dictionary generation, and filter learning. Tang \textit{et al.}~\cite{Tang2011} proposed measuring quality through features based on natural image statistics, distortion texture statistics, and blur/noise statistics. These statistical features are mapped to quality scores by support vector regression. Mittal \textit{et al.}~\cite{Mittal2012} introduced BRISQUE, which is based on natural scene statistics in the spatial domain that are regressed to obtain quality estimates. Natural scene statistics-based methods were extended to frequency domain as in~\cite{Moorthy2010, Saad2012}. Temel \textit{et al.}~\cite{Temel_UNIQUE} introduced an unsupervised approach, in which a linear decoder architecture is used to obtain quality-aware sparse representations whereas Kang \textit{et al.}~\cite{Kang2014} proposed a supervised approach based on Convolutional Neural Networks.

In this manuscript, we follow an alternative approach by performing a frequency domain analysis of error representations. Qadri \textit{et al.}~\cite{QT+11} developed a full-reference method based on harmonic analysis for blockiness artifacts and a reduced-reference method based on harmonic gain and loss. On contrary to \cite{QT+11}, we focus on error images instead of compared images and our approach is not limited to blockiness artifacts and generalized to numerous distortions including compression artifact, image noise, color artifact, communication error, blur, global and local distortions.

\section{Spectral Analysis of Error Representations}
\label{sec:summer_core}
Spatial frequency masking is a characteristic that was observed in various biological visual systems~\cite{AD81}. Albrecht and De Valois~\cite{AD81} showed that striate cortex cells tuned to a spatial fundamental frequency respond to harmonics only if fundamental frequency component exists simultaneously. In addition to the visual system, the auditory system also possess similar masking characteristics. Alphei \textit{et al.}~\cite{AP+87} observed masking of temporal harmonics in the auditory cortex. Because changes in harmonics have the potential to interfere with the masking mechanisms, this interference can affect perceived quality. \textit{Based on the observations related to sensory systems, we hypothesize that changes in frequency domain characteristics can be correlated to changes in perception, specifically perceived quality.} To test our hypothesis, we developed a full-reference image quality assessment algorithm and validated its perceived quality assessment capability in multiple databases. The core of the proposed method is based on the spectral analysis of error representations. 

\begin{figure}[htbp!]
\begin{minipage}[b]{0.305\linewidth}
  \centering
\includegraphics[width=\linewidth]{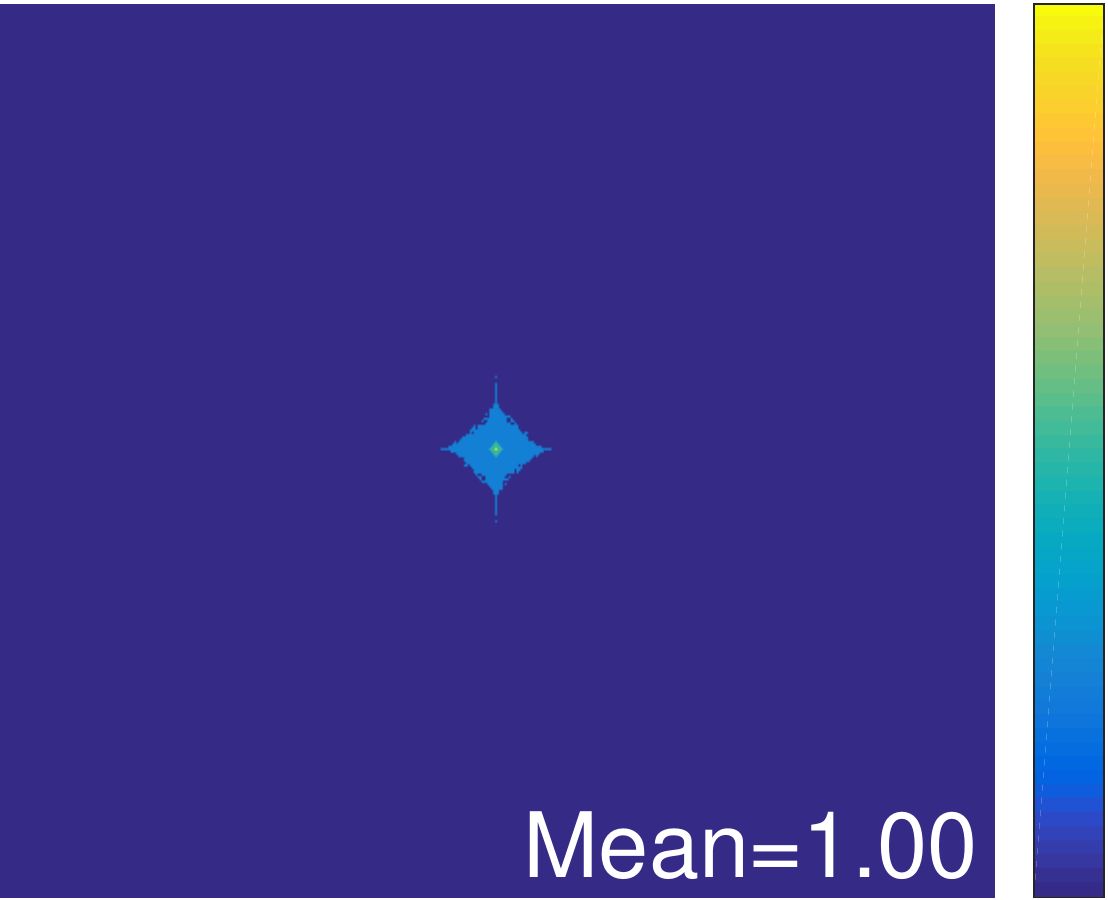}
  \vspace{0.03cm}
  \centerline{\scriptsize{(a)Level 1 Distortion}}
\end{minipage}
\hfill
 \vspace{0.05cm}
\begin{minipage}[b]{0.305\linewidth}
  \centering
\includegraphics[width=\linewidth]{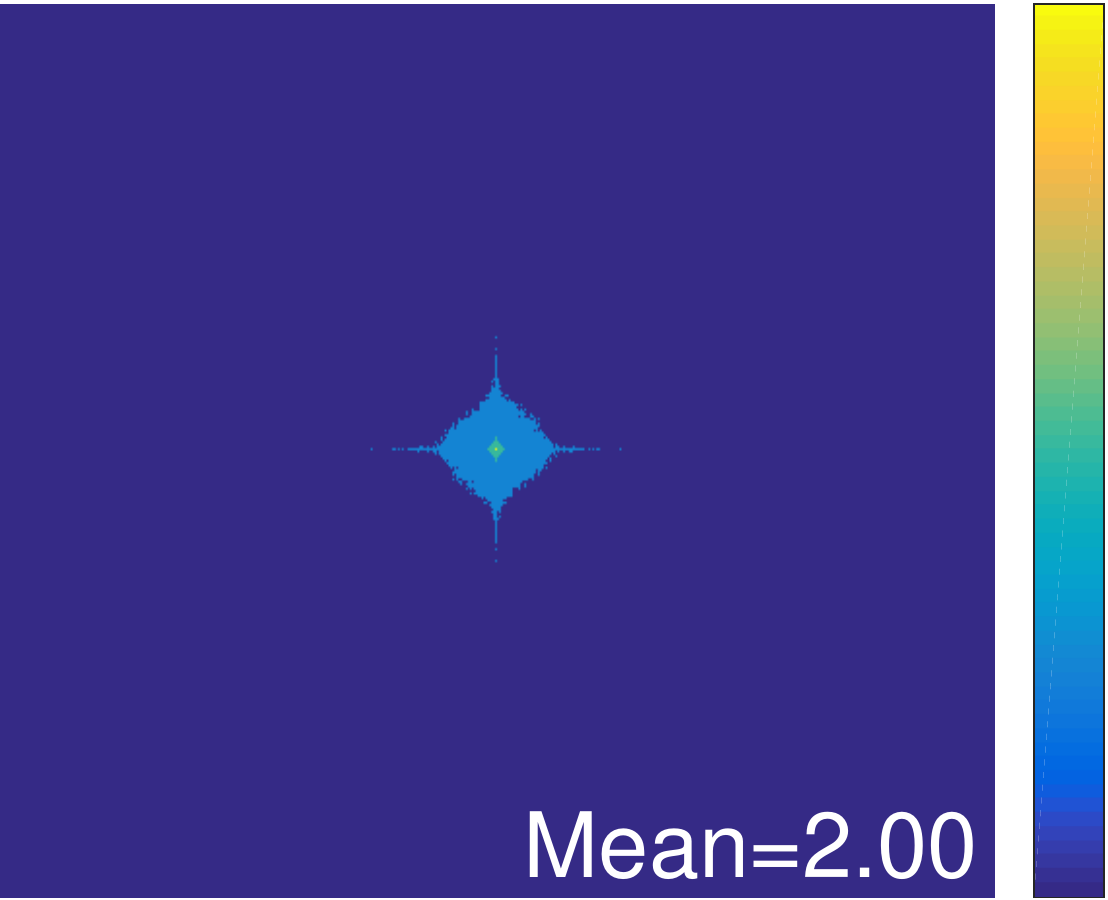}
  \vspace{0.03 cm}
  \centerline{\scriptsize{(b)Level 3 Distortion} }
\end{minipage}
\hfill
 \vspace{0.05cm}
\begin{minipage}[b]{0.305\linewidth}
  \centering
\includegraphics[width=\linewidth]{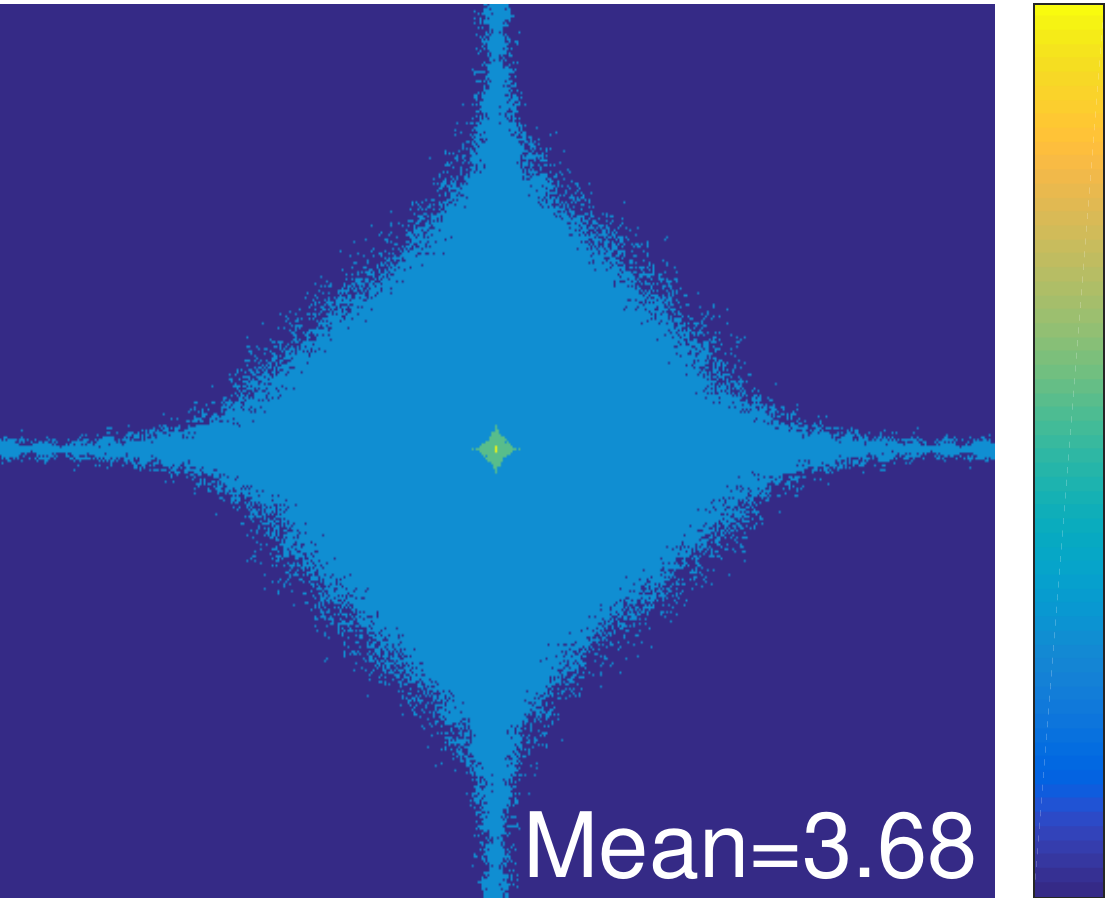}
  \vspace{0.03 cm}
  \centerline{\scriptsize{(c)Level 5 Distortion} }
\end{minipage}
 \caption{Average magnitude spectrum of error signals based on different distortion levels. Each level includes 25 images distorted with 24 degradation types in the TID 2013 database, which corresponds to 600 images (25 images x 24 distortion types) per distortion level.}
\label{fig:spectrums_average}
\end{figure}

Van der Schaaf and Van Hateren \cite{VANDERSCHAAF1996} analyzed the power spectrum of natural images and showed that even though total power and its spatial frequency dependency vary considerably between images, they still follow a common characteristic for natural images. In \cite{Torralba2003}, Torralba and Oliva used the average power spectrum of images to extract information related to naturalness and  openness of the images, semantic category of the scene, recognition of objects in the image, and depth of the scene. Since the power spectrum of the image depend on the context of the scene, directly extracting information related to quality may not be feasible. This is because changes in the context can affect the power spectrum more than the image quality in certain conditions. However, if we calculate the power spectrum of the error signal, we can limit the effect of context and focus more on measuring the degradations.  To test the relationship between the magnitude spectrum of error and the level of distortion, we have conducted an experiment over 1,800 images in the TID 2013 database \cite{tid13}. We obtained the error images by taking the pixelwise difference between reference and compared images. In each distortion level, there are 600 images (25 x 24), which corresponds to 25 reference images distorted with 24 degradation types. We obtained the magnitude information by taking the DFT of the error images and calculating the log magnitude of the transformed images. Fourier images were shifted to display the low frequency components in the central region. Finally, we averaged the magnitude spectrum of the images corresponding to different challenge levels and quantize them to obtain the average magnitude spectrums in Fig.~\ref{fig:spectrums_average}. We calculated the mean value of the spectrums and divided them by the mean value of level 1 spectrum to show the relative change in the mean values. 


\begin{figure}[htbp!]
\begin{minipage}[b]{0.245\linewidth}
  \centering
\includegraphics[width=\linewidth, trim= 0mm 7mm 0mm 0mm]{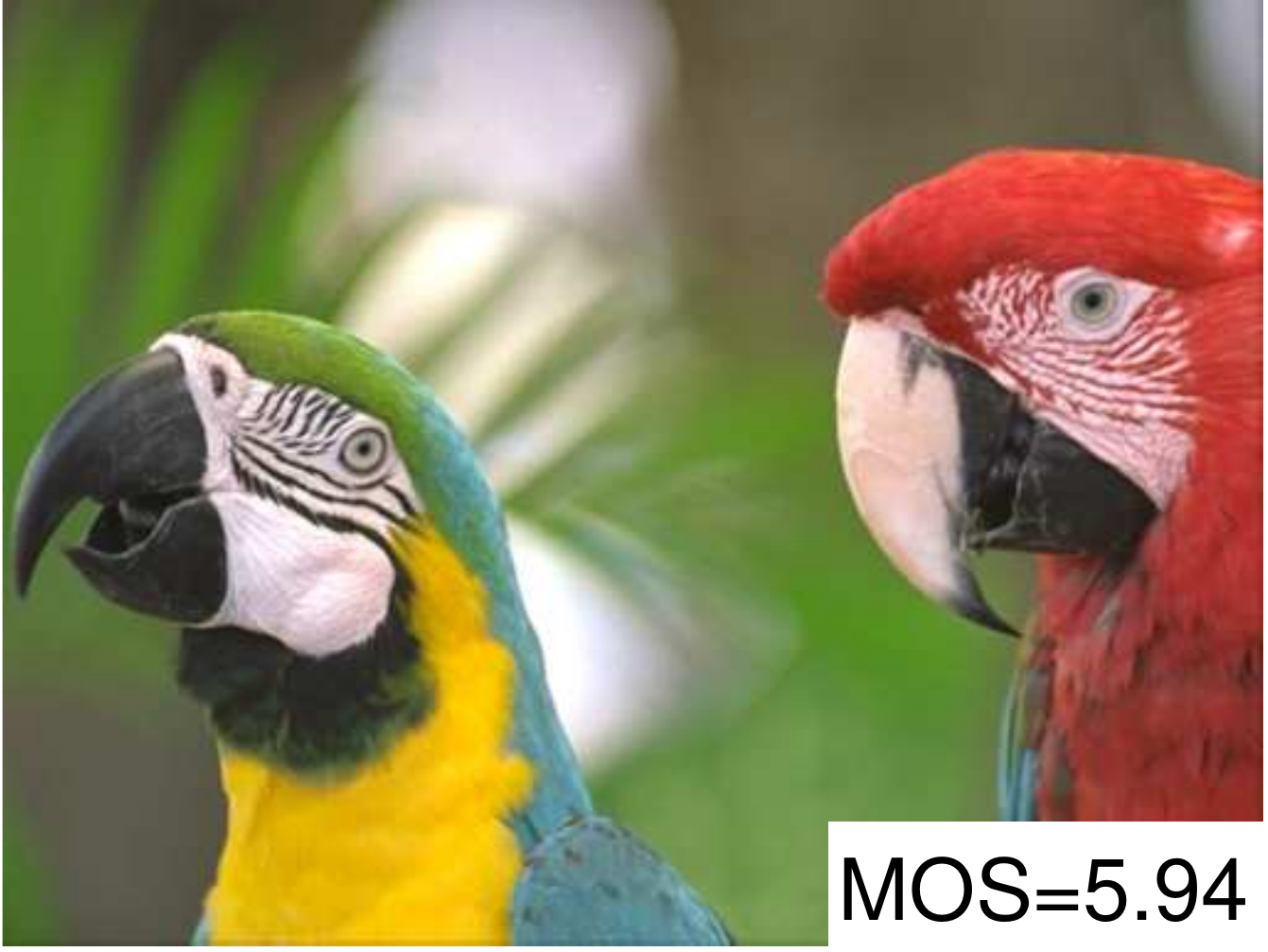}
  \vspace{0.03cm}
  \centerline{\scriptsize{(a)Blur:Level 1 Image}}
\end{minipage}
 \vspace{0.05cm}
\begin{minipage}[b]{0.245\linewidth}
  \centering
\includegraphics[width=\linewidth, trim= 0mm 7mm 0mm 0mm]{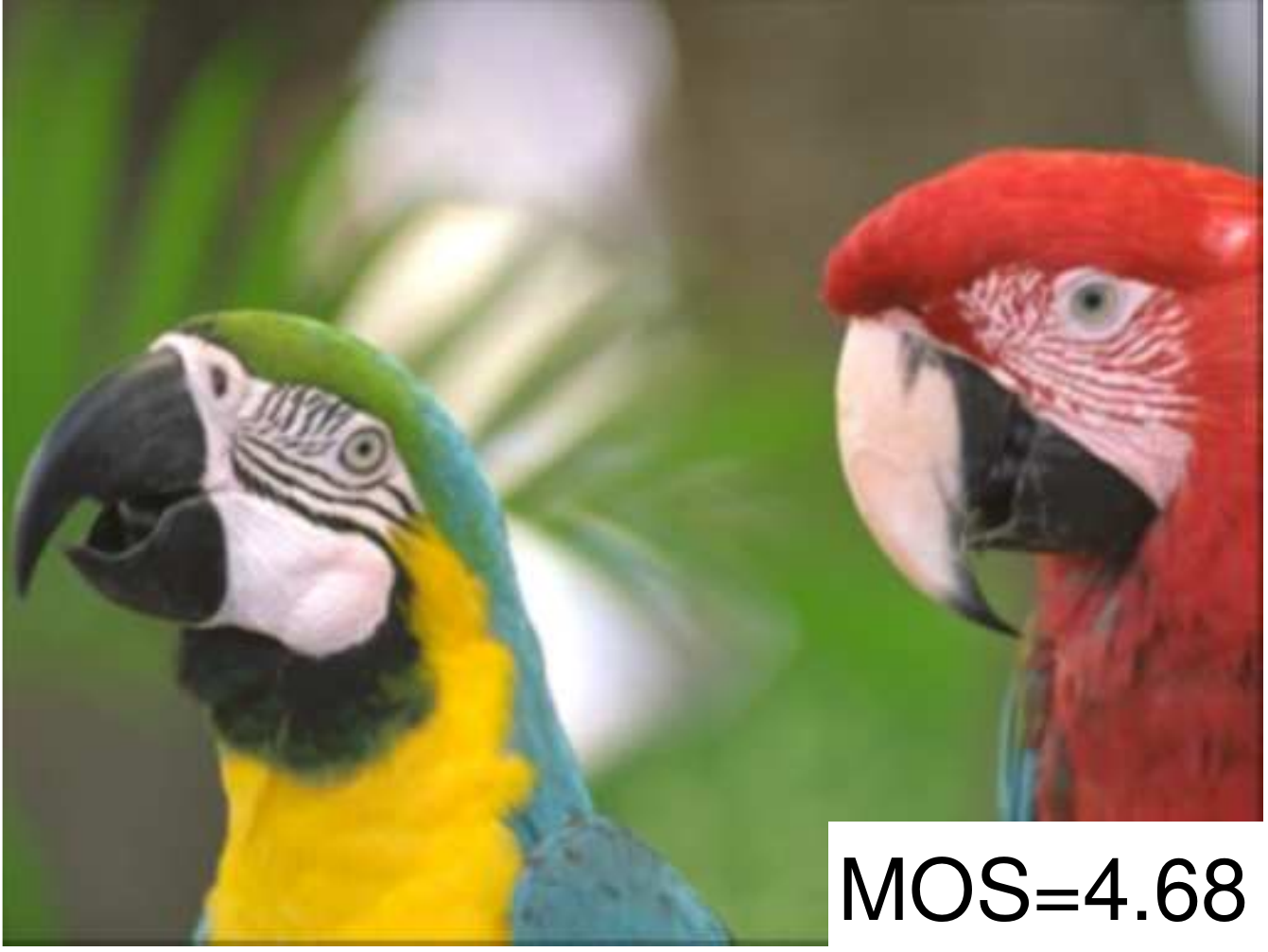}
  \vspace{0.03 cm}
  \centerline{\scriptsize{(b)Blur:Level 3 Image} }
\end{minipage}
 \vspace{0.05cm}
\begin{minipage}[b]{0.245\linewidth}
  \centering
\includegraphics[width=\linewidth, trim= 0mm 7mm 0mm 0mm]{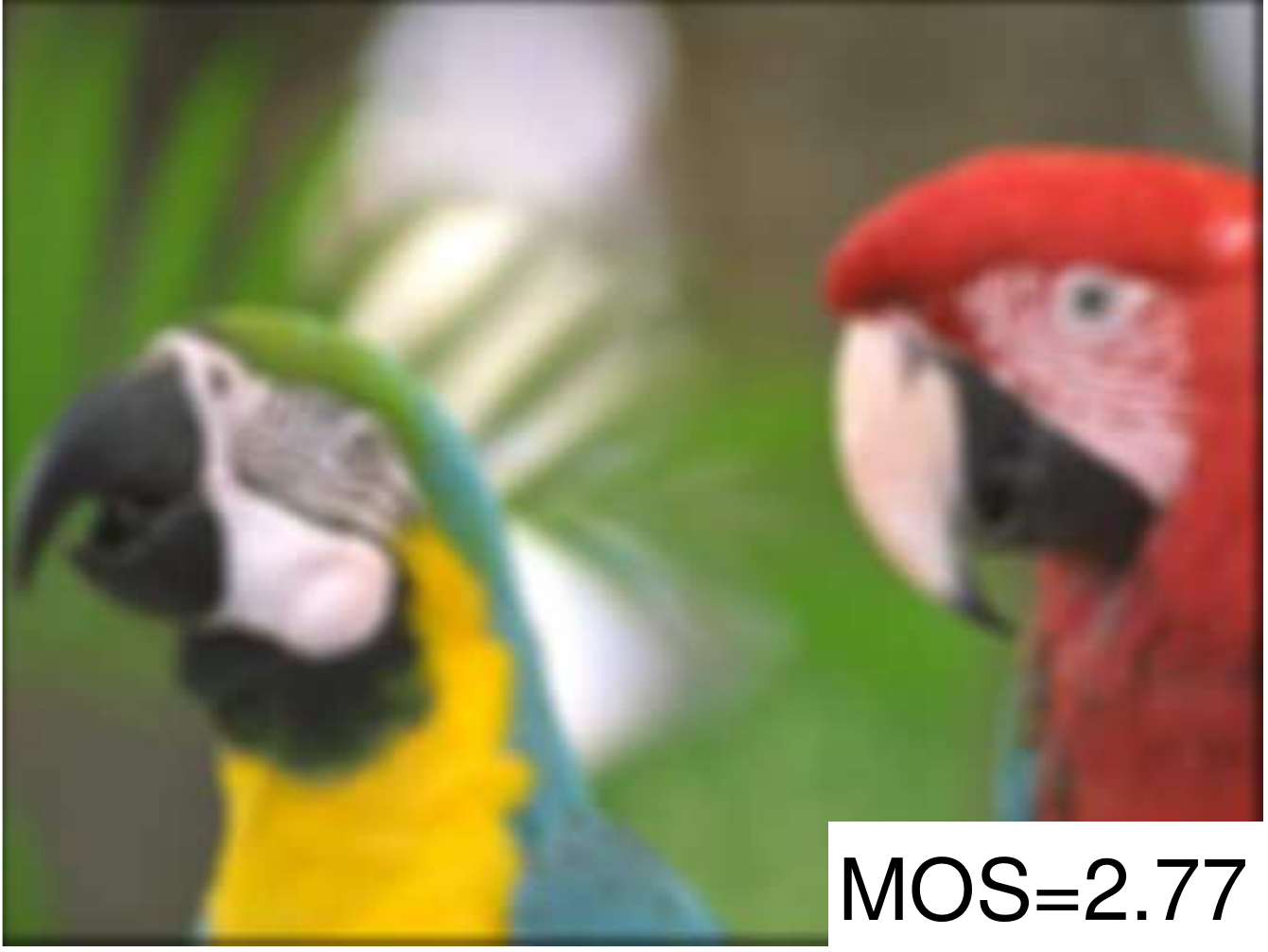}
  \vspace{0.03 cm}
  \centerline{\scriptsize{(c)Blur:Level 5 Image} }
\end{minipage}
 \vspace{0.05cm}
\begin{minipage}[b]{0.255\linewidth}
  \centering
\includegraphics[width=\linewidth, trim= 0mm 7mm 0mm 0mm]{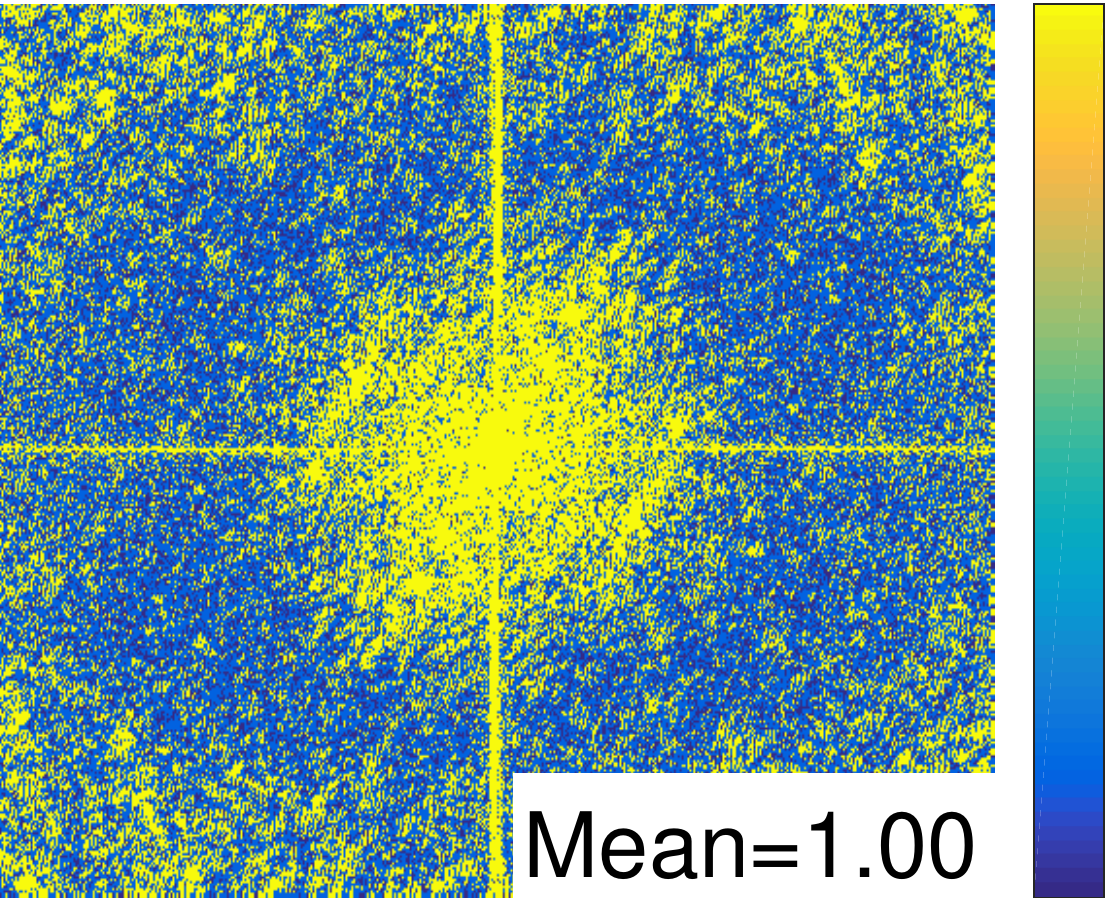}
  \vspace{0.03cm}
  \centerline{\scriptsize{(d)Blur:Level 1 Map}}
\end{minipage}
 \vspace{0.05cm}
\begin{minipage}[b]{0.255\linewidth}
  \centering
\includegraphics[width=\linewidth, trim= 0mm 7mm 0mm 0mm]{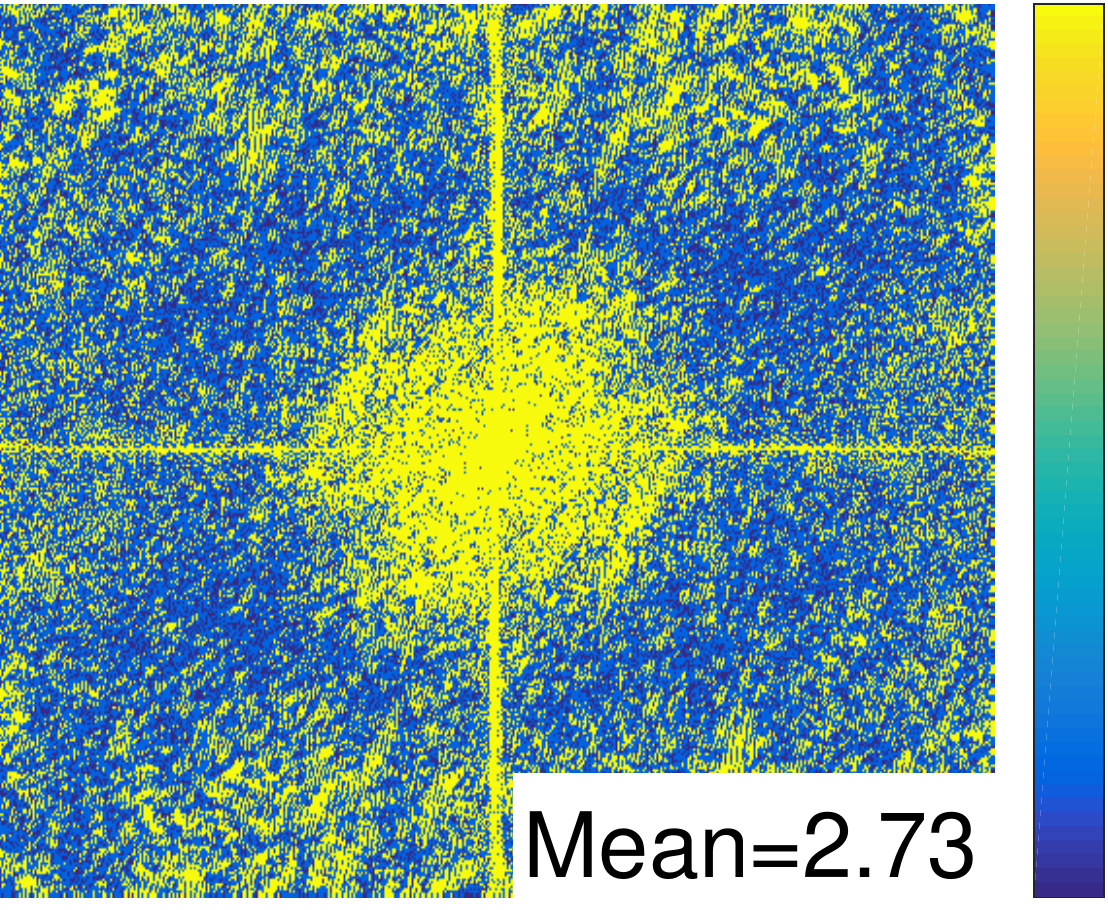}
  \vspace{0.03cm}
  \centerline{\scriptsize{(e)Blur:Level 3 Map}}
\end{minipage}
 \vspace{0.05cm}
\begin{minipage}[b]{0.255\linewidth}
  \centering
\includegraphics[width=\linewidth, trim= 0mm 7mm 0mm 0mm]{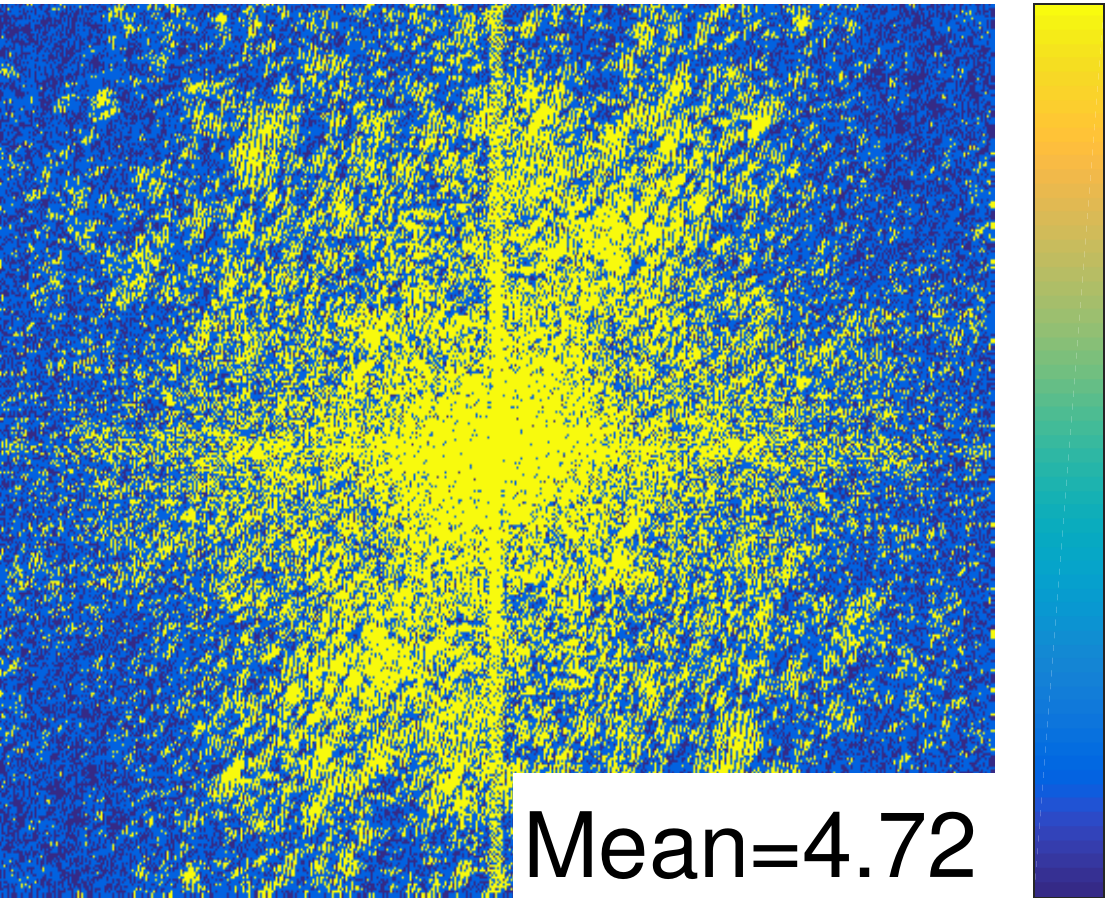}
  \vspace{0.03 cm}
  \centerline{\scriptsize{(f)Blur:Level 5 Map} }
\end{minipage}
 \vspace{0.05cm}
\begin{minipage}[b]{0.245\linewidth}
  \centering
\includegraphics[width=\linewidth, trim= 0mm 7mm 0mm 0mm]{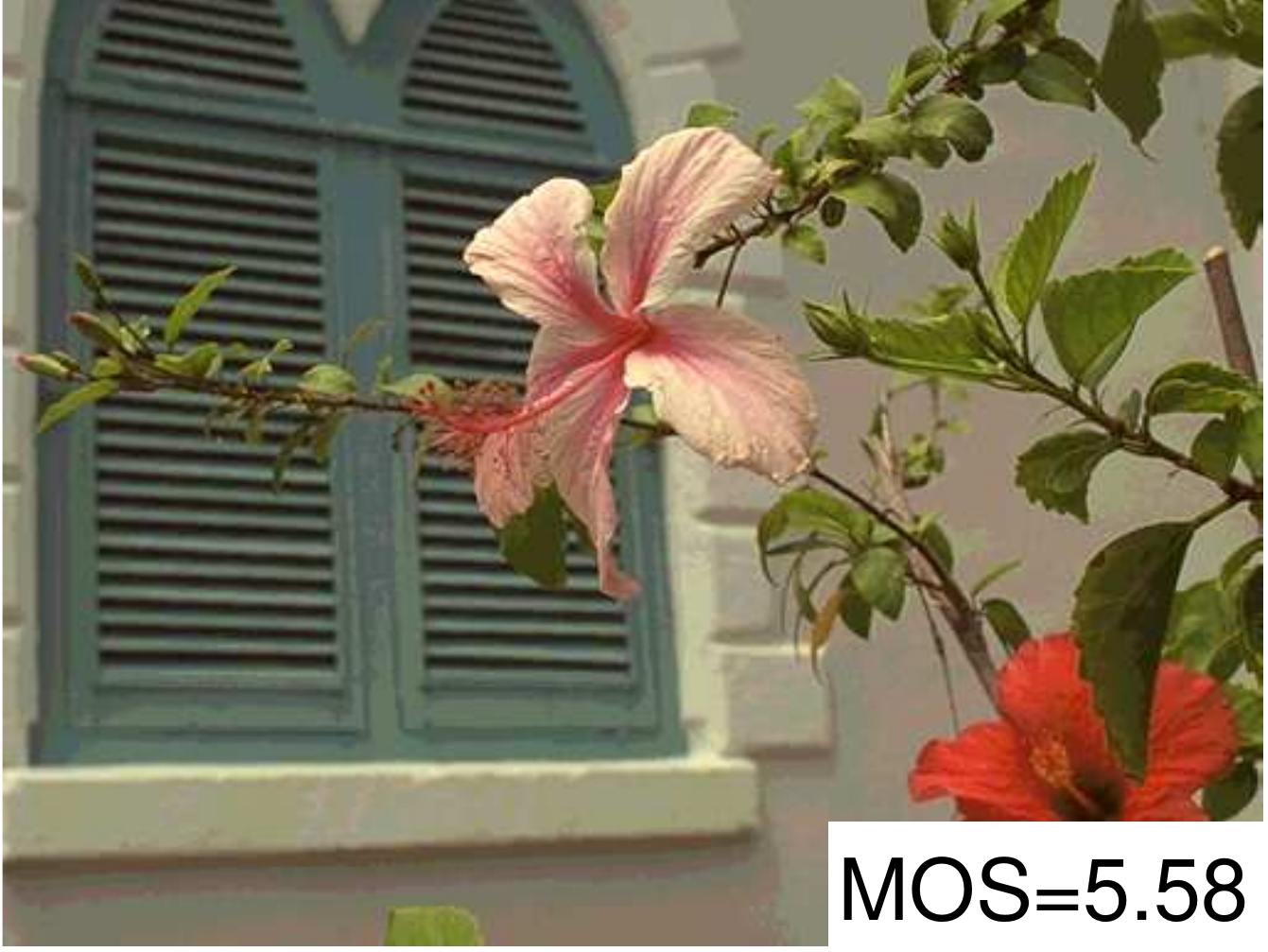}
  \vspace{0.03 cm}
  \centerline{\scriptsize{(g)Quantization:Level 1 Image} }
\end{minipage}
 \vspace{0.05cm}
\begin{minipage}[b]{0.245\linewidth}
  \centering
\includegraphics[width=\linewidth, trim= 0mm 7mm 0mm 0mm]{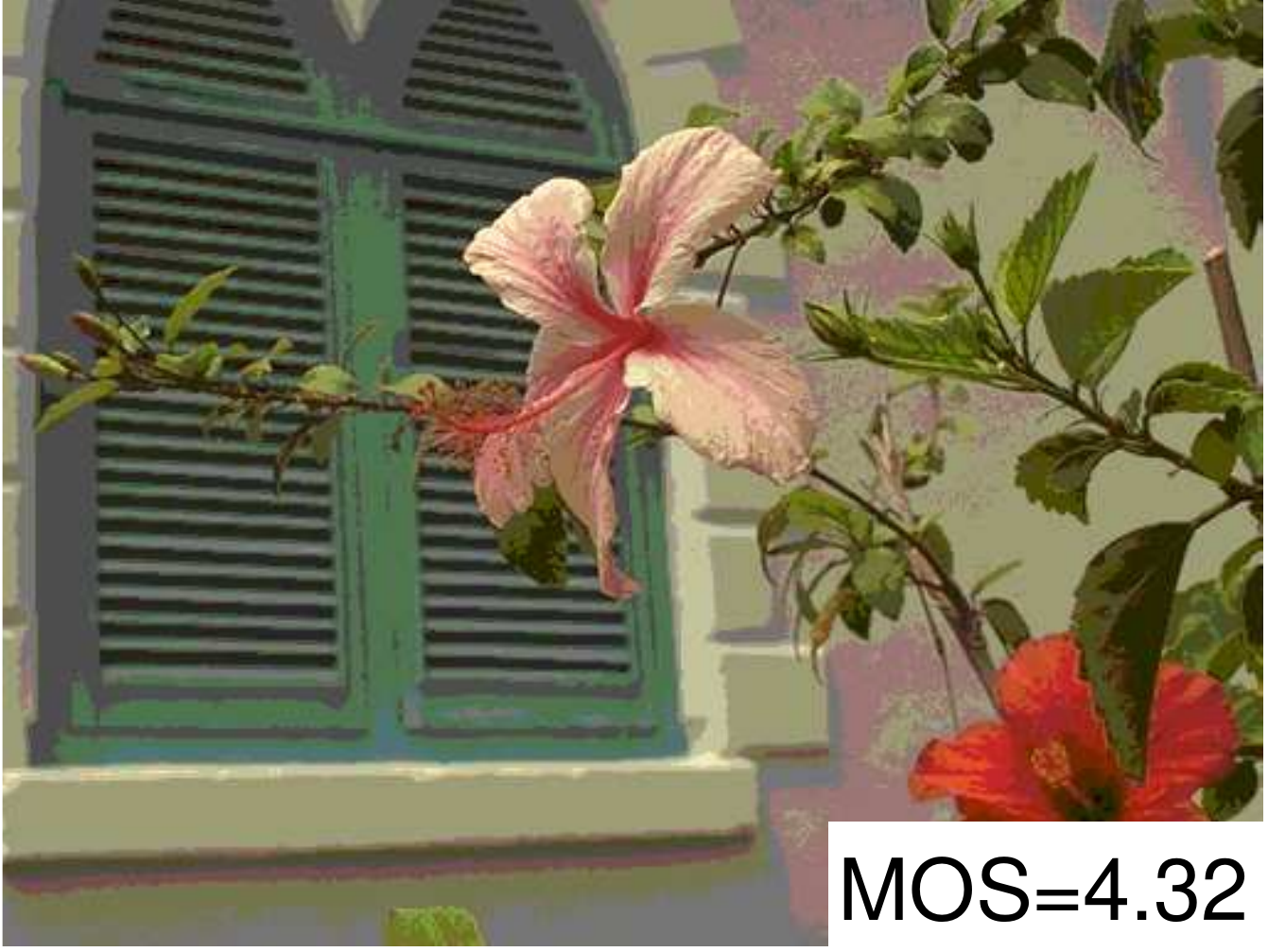}
  \vspace{0.03cm}
  \centerline{\scriptsize{(h)Quantization:Level 3 Image}}
\end{minipage}
 \vspace{0.05cm}
\begin{minipage}[b]{0.245\linewidth}
  \centering
\includegraphics[width=\linewidth, trim= 0mm 7mm 0mm 0mm]{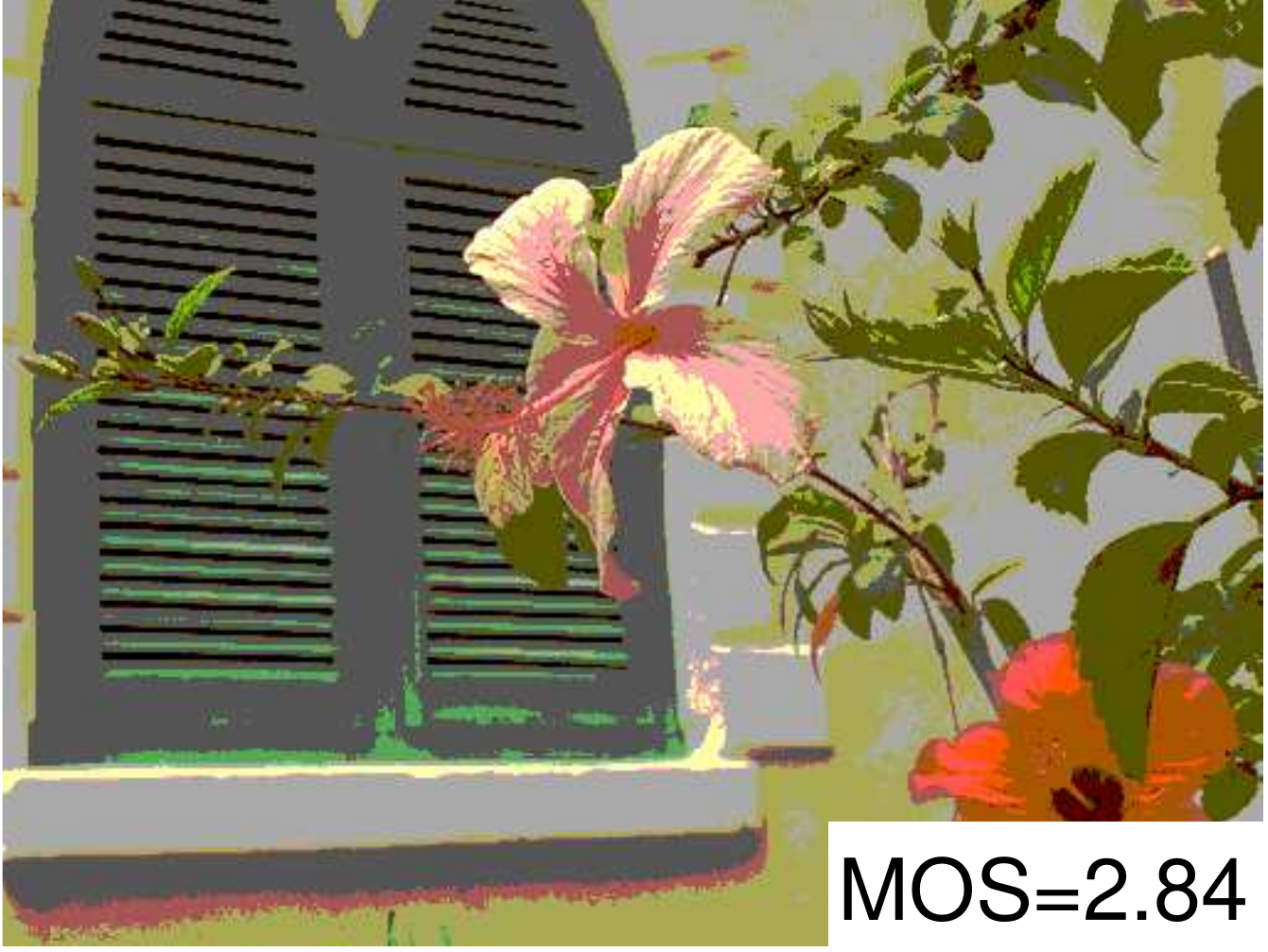}
  \vspace{0.03cm}
  \centerline{\scriptsize{(i)Quantization:Level 5 Image}}
\end{minipage}
 \vspace{0.05cm}
\begin{minipage}[b]{0.255\linewidth}
  \centering
\includegraphics[width=\linewidth, trim= 0mm 7mm 0mm 0mm]{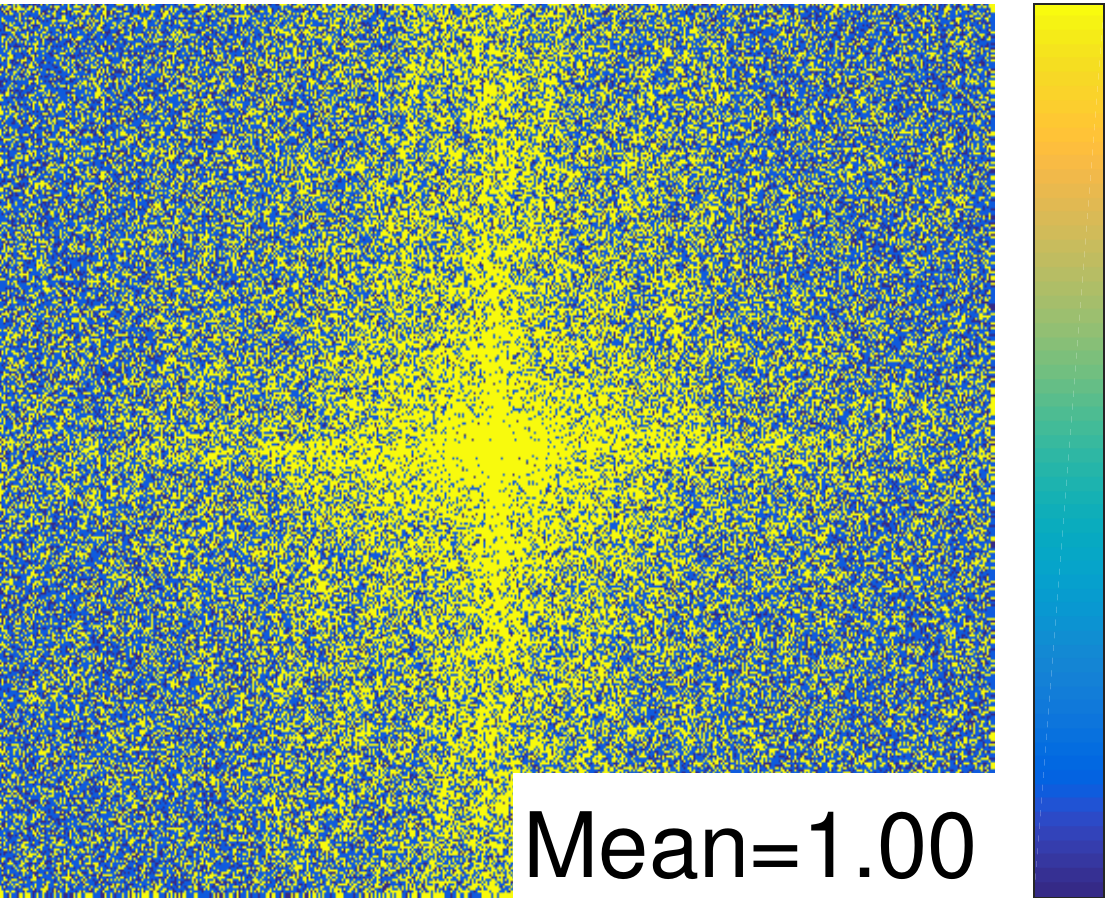}
  \vspace{0.03 cm}
  \centerline{\scriptsize{(j)Quantization:Level 1 Map} }
\end{minipage}
 \vspace{0.05cm}
\begin{minipage}[b]{0.255\linewidth}
  \centering
\includegraphics[width=\linewidth, trim= 0mm 7mm 0mm 0mm]{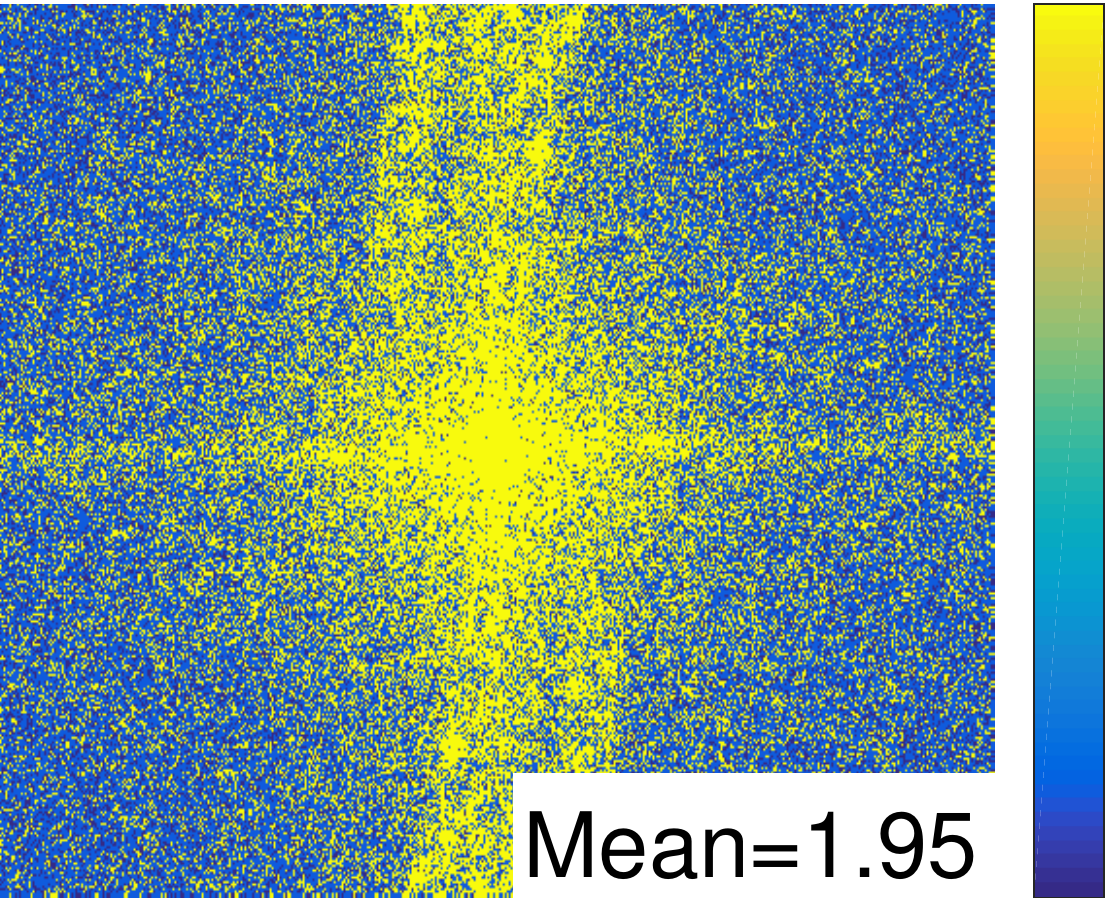}
  \vspace{0.03 cm}
  \centerline{\scriptsize{(k)Quantization:Level 3 Map} }
\end{minipage}
 \vspace{0.05cm}
\begin{minipage}[b]{0.255\linewidth}
  \centering
\includegraphics[width=\linewidth, trim= 0mm 7mm 0mm 0mm]{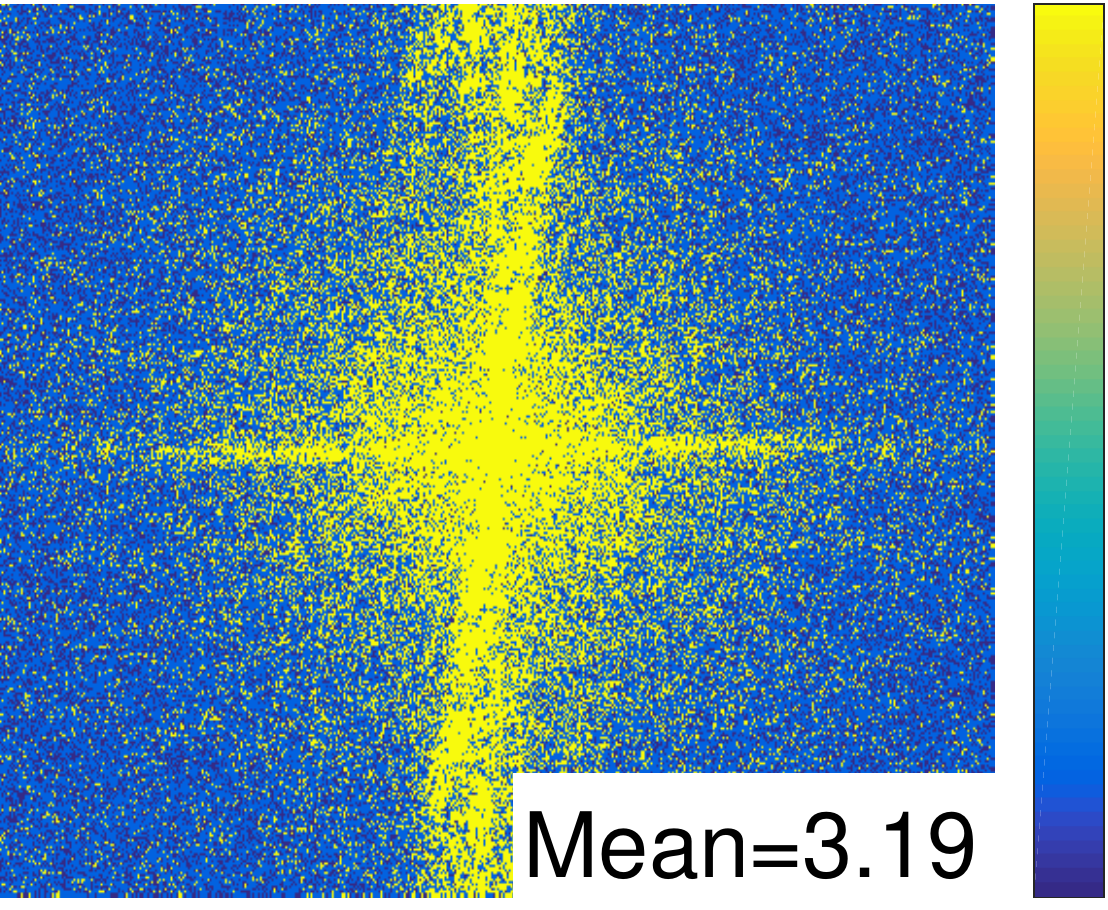}
  \vspace{0.03 cm}
  \centerline{\scriptsize{(l)Quantization:Level 5 Map}}
\end{minipage}
 \vspace{0.05cm}
 \begin{minipage}[b]{0.245\linewidth}
  \centering
\includegraphics[width=\linewidth, trim= 0mm 7mm 0mm 0mm]{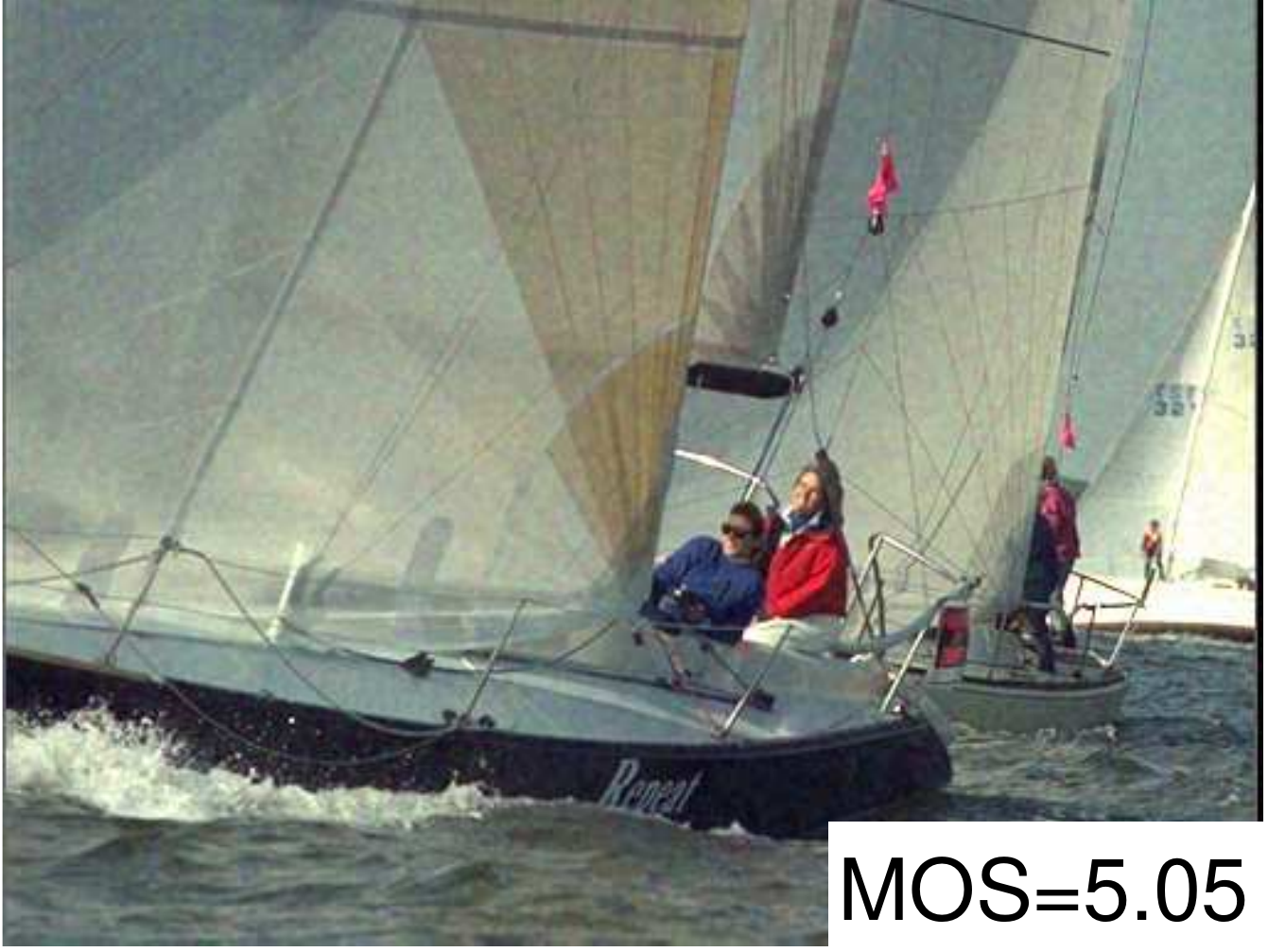}
  \vspace{0.03 cm}
  \centerline{\scriptsize{(m)Noise:Level 1 Image} }
\end{minipage}
 \vspace{0.05cm}
\begin{minipage}[b]{0.245\linewidth}
  \centering
\includegraphics[width=\linewidth, trim= 0mm 7mm 0mm 0mm]{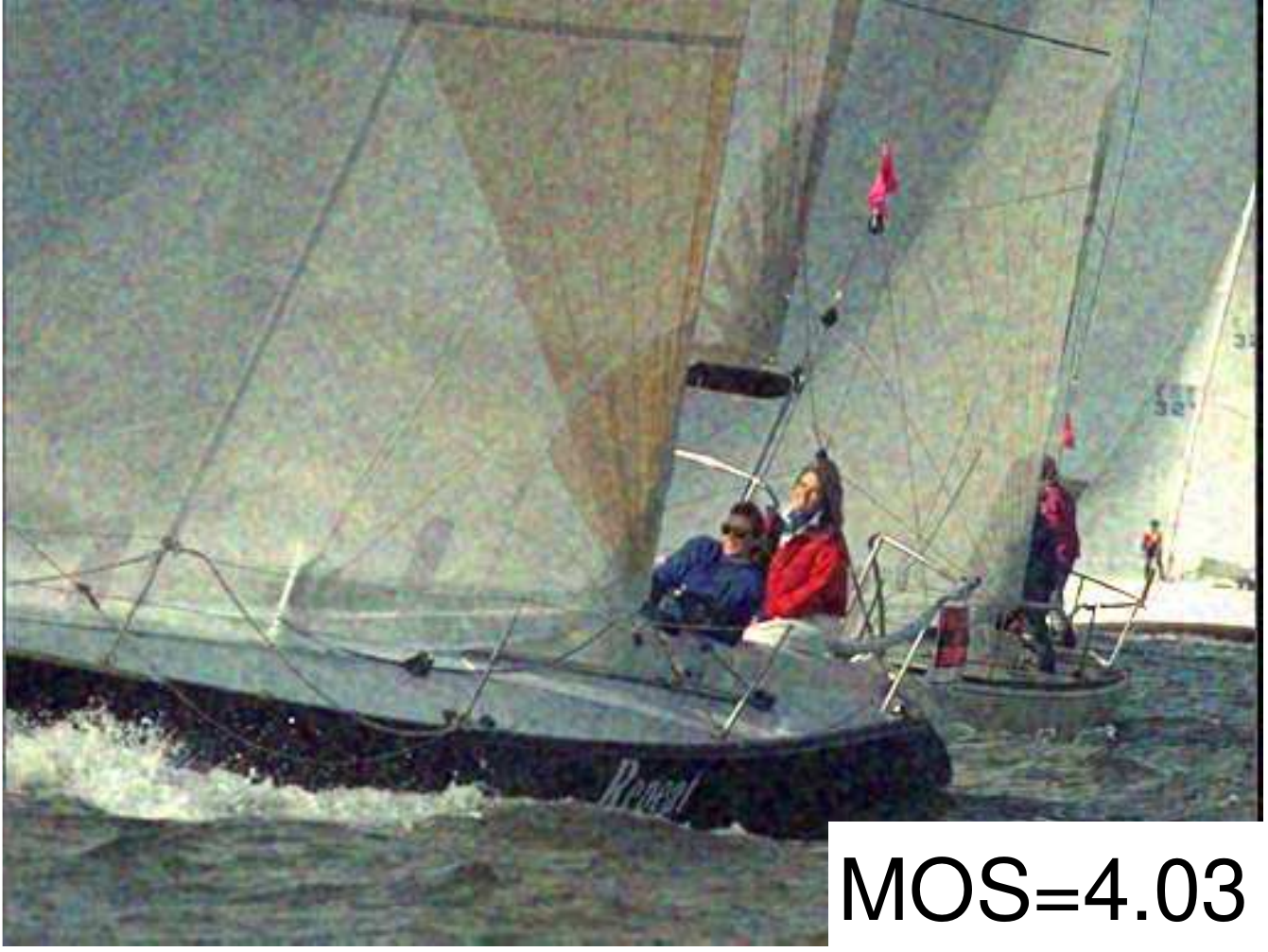}
  \vspace{0.03 cm}
  \centerline{\scriptsize{(n)Noise:Level 3 Image} }
\end{minipage}
 \vspace{0.05cm}
\begin{minipage}[b]{0.245\linewidth}
  \centering
\includegraphics[width=\linewidth, trim= 0mm 7mm 0mm 0mm]{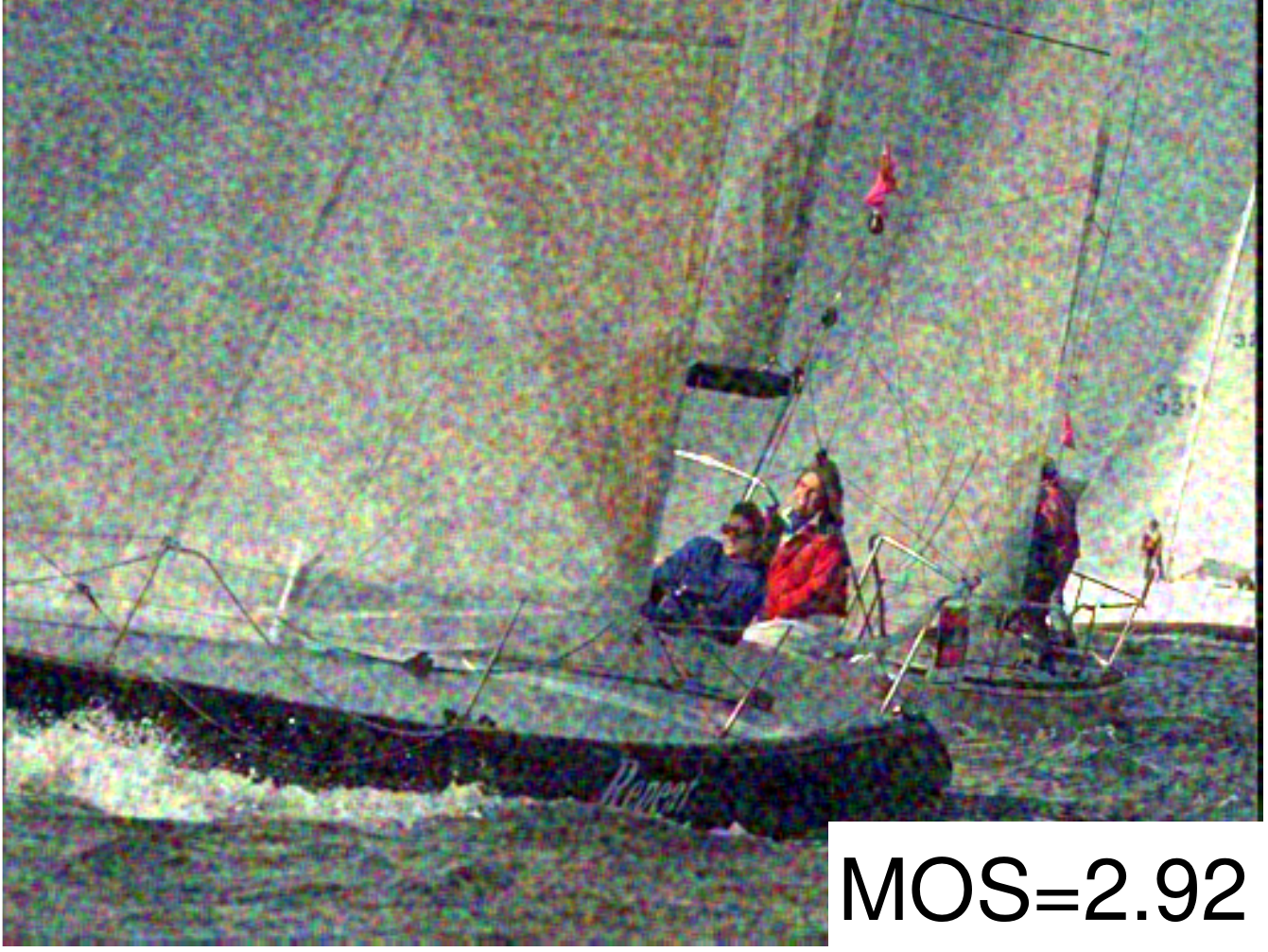}
  \vspace{0.03 cm}
  \centerline{\scriptsize{(o)Noise:Level 5 Image}}
\end{minipage}
 \vspace{0.05cm}
 \begin{minipage}[b]{0.255\linewidth}
  \centering
\includegraphics[width=\linewidth, trim= 0mm 7mm 0mm 0mm]{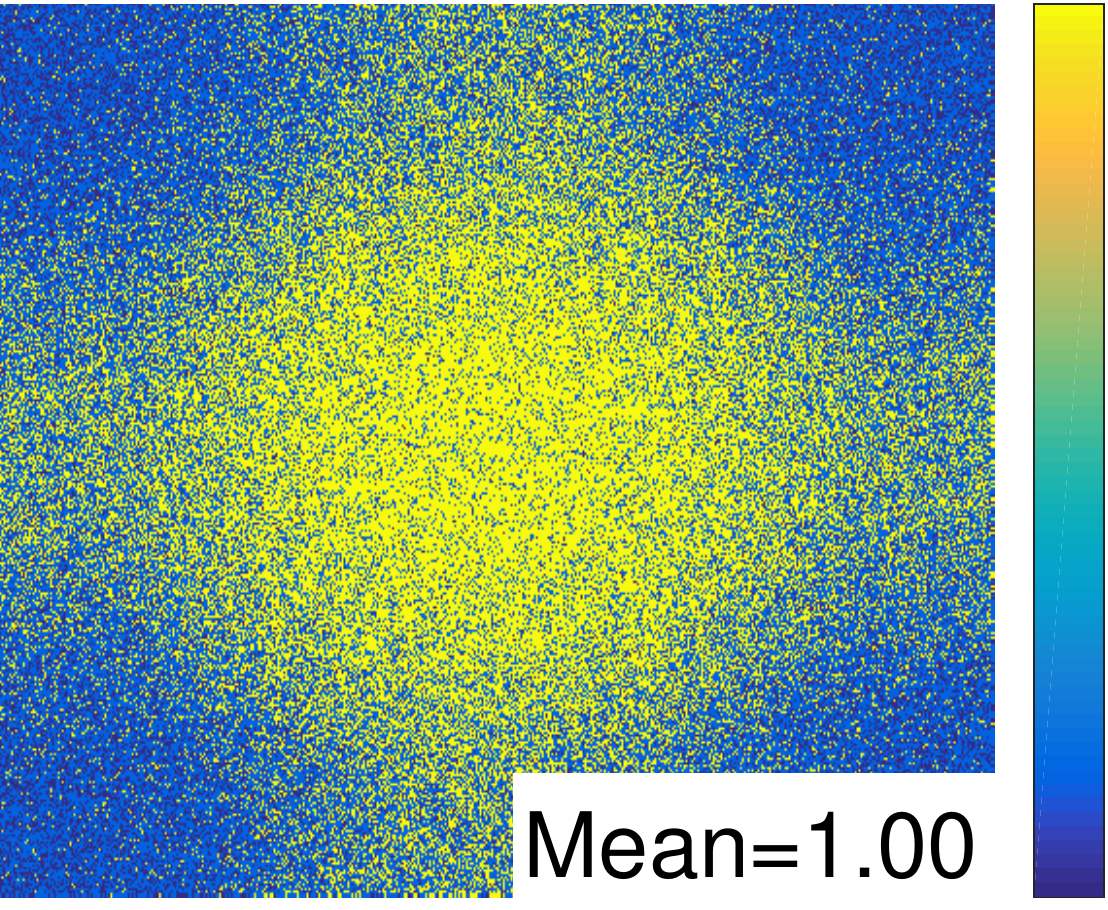}
  \vspace{0.03 cm}
  \centerline{\scriptsize{(p)Noise:Level 1 Map} }
\end{minipage}
\hfill
 \vspace{0.05cm}
\begin{minipage}[b]{0.255\linewidth}
  \centering
\includegraphics[width=\linewidth, trim= 0mm 7mm 0mm 0mm]{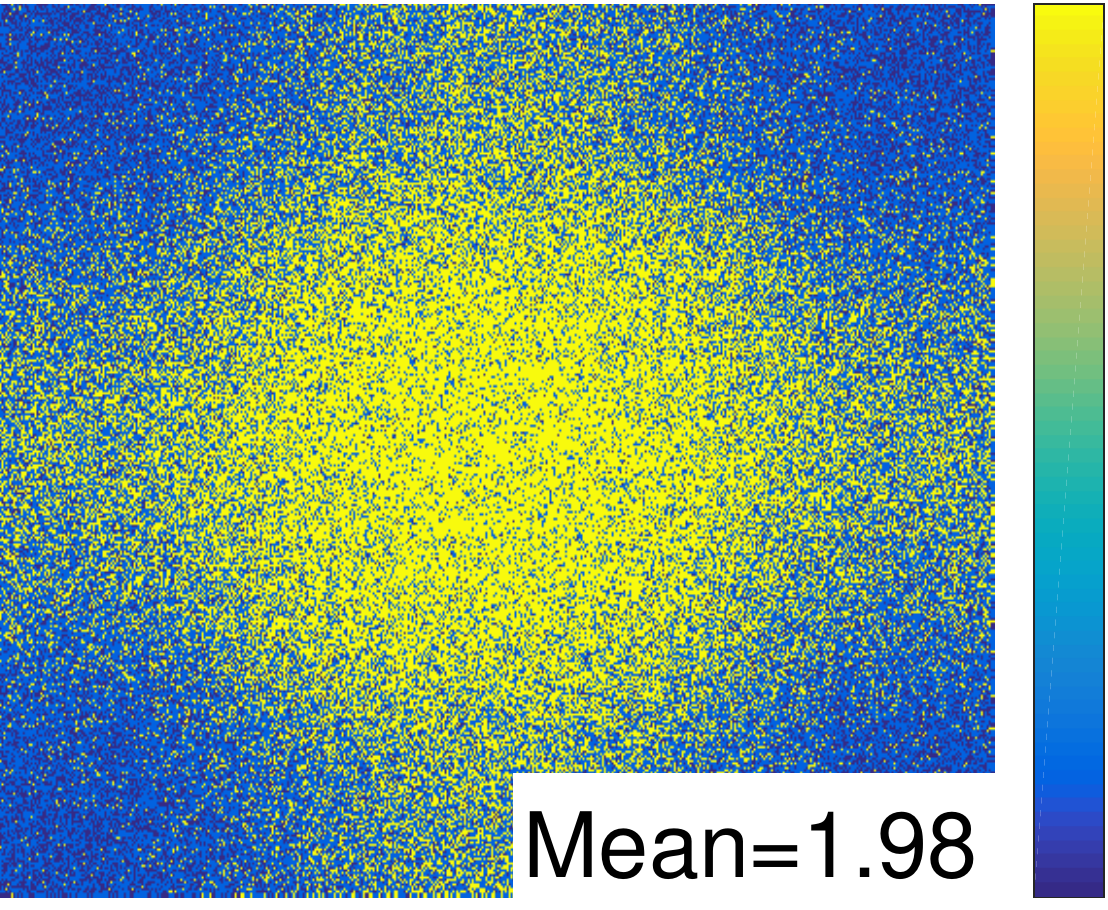}
  \vspace{0.03 cm}
  \centerline{\scriptsize{(q)Noise:Level 3 Map} }
\end{minipage}
\hfill
 \vspace{0.05cm}
\begin{minipage}[b]{0.255\linewidth}
  \centering
\includegraphics[width=\linewidth, trim= 0mm 7mm 0mm 0mm]{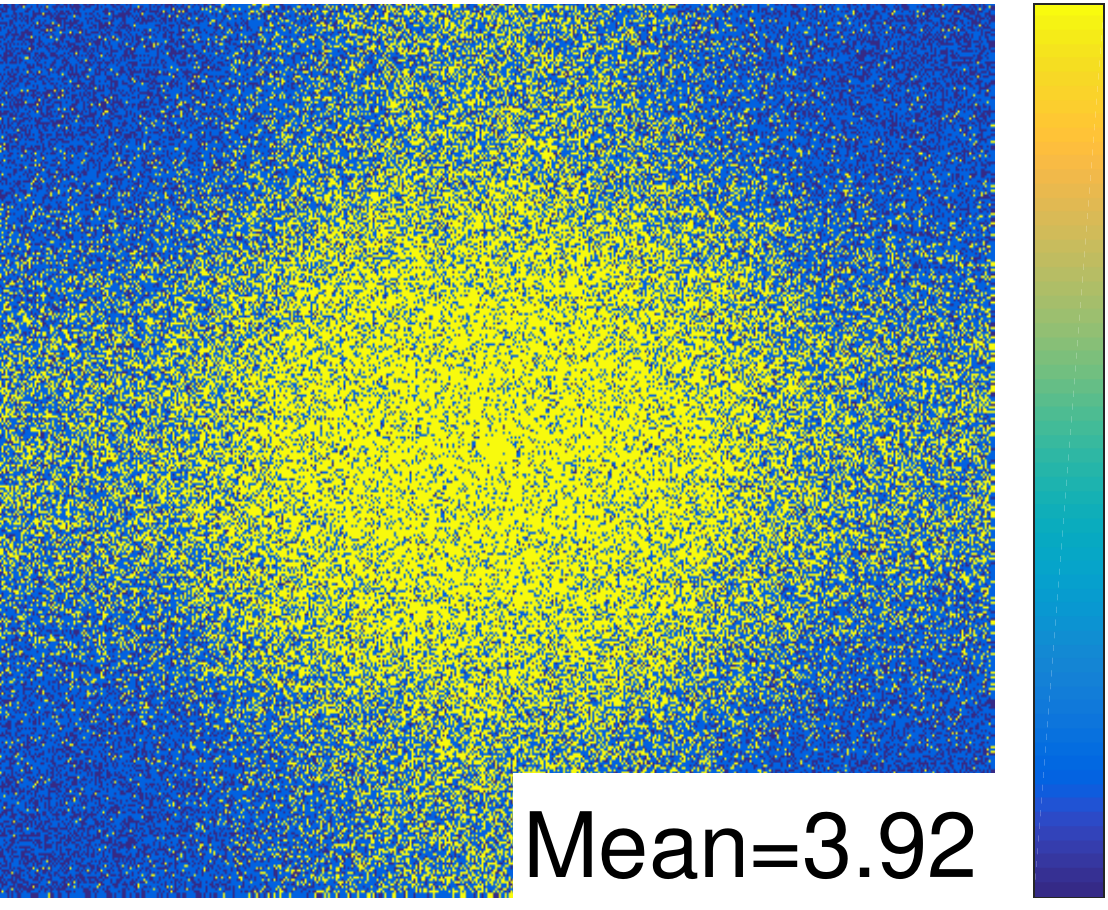}
  \vspace{0.03 cm}
  \centerline{\scriptsize{(r)Noise:Level 5 Map}}
\end{minipage}
\vspace{-0.5cm} 
 \caption{Sample distorted images and magnitude spectrums of corresponding error maps.}
\label{fig:spectrums_samples}
\end{figure}

In these figures, the center of the image corresponds to low frequencies and corners correspond to high frequencies. Pixels are color coded according to the provided color legend based on the intensity levels of each spectrum component. The minimum distortion level is 1 and the maximum level is 5. As distortion level increases, degradations spread over the spectrum and intensify, which corresponds to an increase in the mean value of the spectrums. Therefore, the mean magnitude spectrum provides information related to the distortion level. The average spectrum analysis shows that the magnitude spectrum of error signals can be used to quantify degradations. As observed in Fig.~\ref{fig:spectrums_average}, there is a spectral pattern followed by each distortion level when they are averaged over multiple images. To understand the structure of magnitude spectrums for individual natural images, we analyzed sample images from the TID 2013 database as shown in Fig.~\ref{fig:spectrums_samples}. As sample images, we used images of parrots with an out of focus natural scene in the background, a flower with a house in the background, and a sailboat with another sailboat behind in an ocean. As distortion, we included blur, quantization, and spatially correlated noise with a level of 1 (min), 3 (mid), and 5 (max). We show distorted images in one row and corresponding power spectrums in the following row. On the lower right side, we show the mean opinion scores for the distorted images and the mean values for the normalized magnitude spectrums. 

In case of blur degradation, high frequency components are filtered out and images are smoother. As the degradation level increases, additional lower frequency components get filtered out and error becomes more centralized in the spectrum as observed in Fig.~\ref{fig:spectrums_samples}(d-f). There is a sharp horizontal line and a vertical line, which correspond to the regular patterns in the images. In the quantization degradation, pixels with similar color and texture characteristics converge to similar values because of the loss of details in the quantization stage. Therefore, the error spectrum is more concentrated as the degradation levels increase as observed in Fig.~\ref{fig:spectrums_samples}(j-l). In the noise degradation, pixels are corrupted with a spatially correlated degradation, which leads to local pointy distortions all over the images. Because of the spatial correlation, the magnitude spectrum is more symmetric and continuous as observed in Fig.~\ref{fig:spectrums_samples}(p-r). The shapes of the spectrums are different from each other and from the rhombus shape we obtained for the average spectrums. Therefore, it is not straightforward to pursue a shape-based measurement that is correlated with quality. However, we can pursue a global measurement to quantify the general behavior, which is the mean of the spectrum in this study. To obtain a distortion score, we calculate the mean of the log magnitude spectrum as     
\begin{equation}
 Mean\left\{ Log \left\{\left| \mathscr{F} \left\{\left| E \right| \right\}  \right| \right\} \right\}  ,
\label{eq:harmonics}
\end{equation}
where $E$ is the error map, $|$  $|$ is the absolute value operator, $Log$ is the logarithm, ${\cal F}$ is the $2$-$D$ discrete Fourier transform, and $Mean$ corresponds to the $2$-$D$ mean pooling operation.

\section{Multi-Scale and Multi-Channel Spectral Analysis}
Spectral analysis of error signals is calculated over a single scale in the original resolution of the images. However, multi-scale representations and transforms can be considered as partial visual system models  because neural responses in a visual cortex include scale-space orientation decomposition. Multiple scale and resolution approaches enable visual representations that can support different abstraction levels, which are commonly used in the image quality assessment literature \cite{Wang2003,Sampat2009,Wang2011,EA+06,PS+07,Ponomarenko2011,Zhang12,Venkata2000,Chandler2007,
temel_15_persim,Mittal2012}. To extend the single-scale baseline method to multi-scale, we perform a spectral analysis over multiple resolutions. Specifically, we downsample error maps by factors of $2^i$, where $i$ is the scale index varying from $1$ to $4$. The number of scales can be adjusted based on image characteristics. If the proposed quality assessment algorithm is used for higher resolution images, baseline scale or number of scales can be increased.

Color information is also overlooked in the baseline spectral analysis method. To naively utilize color information, we perform a multi-scale spectral analysis over each color channel in the RGB color space. In the RGB color space, color and intensity information is mixed. Therefore, we can utilize the same operator over each channel. However, in more perceptually correlated color spaces, color channels are more decorrelated and spectral analysis should not be performed over these channels in an identical fashion. Nevertheless, in the algorithm development process, we switched RGB with CIEXYZ, CIELa*b*, YCbCr, and HSV color spaces and they all underperformed.       

\begin{figure}[htb!]
\centering
\includegraphics[width=\textwidth] {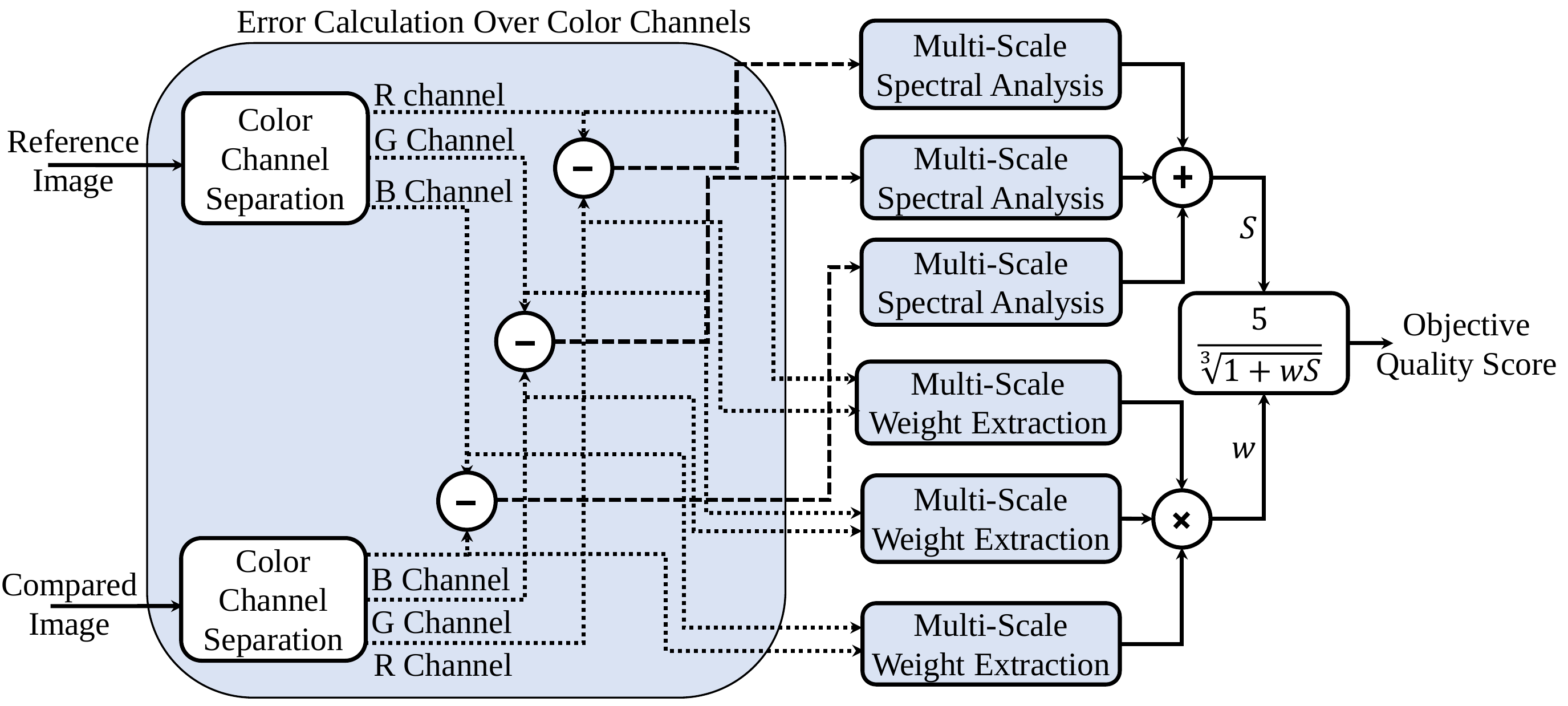}
\caption{Block diagram of objective quality estimation based on multi-scale and multi-channel spectral analysis.}
\label{fig:fm_coherensi}
\end{figure}

When we measure quality solely based on error maps, quality estimation ranges and monotonic behaviors vary with the distortion type. These variations lead to misalignments that degrade the overall quality estimation performance. To eliminate these misalignments, we calculate frequency-based weights as 
\begin{equation}
 Log\left\{ Mean \left\{ \left| \mathscr{F} \left\{ I \right\} /  \mathscr{F} \left\{ J \right\}  \right| \right\} \right\}  ,
\label{eq:harmonics}
\end{equation}
where $I$ is the reference image, $J$ is the compared image, ${\cal F}$ is the $2$-$D$ discrete Fourier transform, $|$  $|$ is the absolute value, $Log$ is the logarithm, and $Mean$ corresponds to the $2$-$D$ mean pooling. Instead of calculating frequency-based weights over all scales, we only compute them for the final two smallest scales. This is because frequency-based weights calculated over high resolution images are sensitive to minor changes that do not necessarily correspond to perceived degradations. We multiply the multi-scale weights to obtain a single weight (w), which is further multiplied with the the spectral analysis-based score (S) as shown in Fig.~\ref{fig:fm_coherensi}. We set the maximum objective score to five and divide it by the cube root of one plus the weighted score (wS) to obtain the final quality score. As the compared image gets similar to the reference image, final score converges to five.

\label{sec:summer}

\section{Experimental Setup}
\label{sec:setup}
 \subsection{Databases}
In order to validate the performance of image quality estimators, we need to use comprehensive databases that include a high variety of distortion types. To satisfy this requirement, we utilize the TID 2013 (TID13) database \cite{tid13}, which is one of the most comprehensive image quality assessment databases with reference images in the literature in terms of distortion types. In addition to the TID13 database, we utilize the LIVE database \cite{Sheikh2006b}, which is one of the most commonly used image quality databases. Even though TID13 and LIVE cover a wide range of distortion types, they do not include simultaneously applied distortion. To consider simultaneous distortions in the validation, we utilize also the LIVE Multiply Distorted (MULTI) database \cite{Jayaraman2012}. LIVE database experiments were conduced in an office environment with normal indoor illumination levels in which subjects viewed a 21 inch CRT monitor that displayed mostly $768\times512$ pixel images from an approximate viewing distance of 2-2.5 screen height. The illumination conditions of the test protocol are not explicitly stated by the authors in \cite{Wang2004}. MULTI database experiments were performed in a workspace environment under normal illumination levels in which subjects viewed a monitor that displayed  $1280\times720$ pixel images from an approximate distance of 4 times screen height. TID13 database experiments were conduced in laboratory conditions as well as through internet in which subjects were recommended to use a convenient distance to their monitors. We utilize databases with different setups because it is not possible restrict users in real life and we need to develop generic visual quality estimators that should operate in diverse platforms and conditions.

There are $5$ distortion types in the LIVE database, $3$ types in the MULTI database, and $24$ types in the TID13 database. Even though specific distortion types are different from each other, they can be grouped into common categories according to their high-level characteristics. In this study, individual distortion types are grouped into $7$ main categories as follows: The \texttt{Compression} category includes JPEG, JPEG 2000, and lossy compression of noisy images. The \texttt{Noise} category includes additive Gaussian noise, additive noise in chroma channels, impulse noise, spatially correlated noise, masked noise, high frequency noise, quantization noise, image denoising artifacts, multiplicative Gaussian noise, comfort noise, lossy compression of noisy images and white noise. The \texttt{Communication} category includes Rayleigh fast-fading channel error, JPEG and JPEG2000 transmission errors. The \texttt{Blur} category includes Gaussian blur and sparse sampling and reconstruction error. The \texttt{Color} category contains color saturation change and color quantization with dither and chromatic aberrations. The \texttt{Global} category includes intensity shift and contrast change. The \texttt{Local} category includes non-eccentricity pattern and local block-wise distortion of different intensity. The number of images in each category is summarized in Table \ref{tab_db}. 

\begin{table}[htbp!]
\scriptsize
\centering
\caption{The number of images per degradation category in each database.}
\label{tab_db}
\begin{tabular}{|c|c|c|c|c|}
\hline
                    & {\bf LIVE \cite{Sheikh2006b}} & {\bf MULTI \cite{Jayaraman2012}} & {\bf TID13 \cite{tid13}} & {\bf Total} \\ \hline
{\bf Compression}   & 460        & 180         & 375         & 1015        \\ \hline
{\bf Noise}         & 174        & 180         & 1375        & 1729        \\ \hline
{\bf Communication} & 174        & -           & 250         & 424         \\ \hline
{\bf Blur}          & 174        & 315         & 250         & 739         \\ \hline
{\bf Color}         & -          & -           & 375         & 375         \\ \hline
{\bf Global}        & -          & -           & 250         & 250         \\ \hline
{\bf Local}         & -          & -           & 250         &250          \\ \hline
\end{tabular}
\end{table}

\vspace{4.0mm}

\subsection{Validation Setup}
We utilize the Pearson linear correlation coefficient (PLCC) to measure linearity, Spearman rank-order correlation coefficient (SRCC) and the Kendall rank-order correlation coefficient (KRCC) to measure monotonicity, root mean squared error (RMSE) to measure accuracy, and outlier ratio (OR) to measure consistency of quality estimates \cite{ITU-T,Kendal1945}. We regress quality scores before computing validation metrics as in \cite{Sheikh2006b}, which can formulated as
\begin{equation}
\label{eq:nonlinreg}
Q=\beta_1 \left ( \frac{1}{1}-\frac{1}{2+\exp(\beta_2(Q_0 -\beta_3 ))} \right )+\beta_4 Q_0 +\beta_5,
\end{equation}
where $Q_0$ is the objective score, $Q$ is the regressed objective score, and $\beta$s are the parameters that are tuned based on the relationship between objective and subjective scores. We utilized the \texttt{fitnlm} function in MATLAB and initialized  regression coefficients to [0.0, 0.1, 0.0, 0.0, 0.0], from which a nonlinear model started its search for optimal coefficients. Reported performances of existing methods can vary from the literature because of the differences in regression curves and initialization
coefficients.

We use statistical tests suggested in ITU-T Rec.P.1401 \cite{ITU-T} to evaluate the significance of difference between correlation coefficients. There are two main hypothesis in a statistical significance test. The first one ($H_0$) claims that there is no significant difference between compared correlation coefficients and the second one ($H_1$ ) claims that there is a significant difference between compared correlation coefficients. In order to verify whether $H_0$ is true or not, at first, we assume that $H_0$ is true.  Then, we calculate Fisher-z transforms of compared correlation coefficients, compute mean and standard deviation of Fisher-z transform values, and obtain the significance value as in \cite{ITU-T}. If the significance value is below the two-tailed t-distribution value, $H_0$ is true, otherwise $H_1$ is true. We use the tabulated t-distribution values for the $95\%$ significance level of the two tailed test. 

To analyze the distribution of subjective scores versus objective scores of best performing quality estimators, we provide scatter plots, whose x-axis corresponds to quality estimates and whose y-axis corresponds to mean opinion scores (MOS) or differential mean opinion scores (DMOS). An ideal quality estimator leads to a scatter plot in which scores should be located on a linear curve. Moreover, to further analyze the difference between subjective and objective scores, we calculate the difference between normalized histograms of subjective scores and regressed quality estimates as in \cite{temel_16_csv}. We utilize the common histogram differences metrics including Earth Mover's Distance (EMD), Kullback-Leibler (KL) divergence, Jensen-Shannon (JS) divergence, histogram intersection (HI), and $l_2$ norm.

\begin{table*}[htb!]
\tiny
\begin{adjustwidth}{-2.0cm}{}

\centering
\caption{Overall performance of image quality estimators.}
\label{tab_results_all}

\begin{threeparttable}

\begin{tabular}{p{1.3cm}p{0.45cm}p{0.45cm}p{0.45cm}p{0.45cm}p{0.45cm}p{0.45cm}p{0.45cm}p{0.45cm}p{0.45cm}p{0.45cm}p{0.45cm}p{0.45cm}p{0.45cm}p{0.45cm}p{0.45cm}p{0.45cm}p{0.55cm}p{0.80cm}}
\hline


\multirow{3}{*}{{\bf Databases }}                 &\bf PSNR  & \bf PSNR  &\bf PSNR  &\bf SSIM  &\bf MS  &\bf CW &\bf IW  &\bf SR  &\bf FSIM  &\bf FSIMc  &\bf BRIS  & \bf BIQI &\bf BLII &\bf Per &\bf CSV &\bf UNI &\bf COHER &\bf SUMMER \\
& &\bf HA &\bf HMA & &\bf SSIM  &\bf SSIM  &\bf SSIM  &\bf SIM  & &&\bf QUE & &\bf NDS2 &\bf SIM & & \bf QUE & \bf ENSI & \\ & &\cite{Ponomarenko2011} & \cite{Ponomarenko2011} &\cite{Wang2004} &\cite{Wang2003} &\cite{Sampat2009}&\cite{Wang2011}&\cite{Zhang12} &\cite{Zhang2011} &\cite{Zhang2011} &\cite{Mittal2012} &\cite{Moorthy2010} &\cite{Saad2012}&\cite{temel_15_persim} & \cite{temel_16_csv}&\cite{Temel_UNIQUE}  &\cite{Hegazy2014} &
\\ \hline




            \textbf{}   & \multicolumn{18}{c}{\textbf{Average Computation Time per Image }}                                                                                                                                                                        \\ \hline
 
\textbf{All}    & 0.02 & 1.75 & 1.75 & 0.02 & 0.05 & 1.09 & 0.35 & 0.03 & 0.20 & 0.20 & - & - & - & 0.45 & 0.65 & 0.29 & 0.05 & 0.07          \\ \hline
         
                & \multicolumn{18}{c}{\textbf{Outlier Ratio (OR)}}                                                                                                                                                                        \\ \hline
\textbf{MULTI}  
& 0.009 & 0.013 & 0.009 & 0.016 & 0.013 & 0.093 & 0.013 &\cellcolor{blue!10} \bf 0.000 & 0.018 & 0.016 & 0.067 & 0.024 & 0.078 & 0.004 &\cellcolor{blue!10} \bf 0.000 &\cellcolor{blue!10} \bf 0.000 & 0.031 &\cellcolor{blue!10} \bf 0.000 
\\                            
\textbf{TID13}  
& 0.725 &\cellcolor{blue!10} \bf 0.615 & 0.670 & 0.734 & 0.743 & 0.856 & 0.701 & 0.632 & 0.742 & 0.728 & 0.851 & 0.856 & 0.852 & 0.655 & 0.687 & 0.640 & 0.833 & \cellcolor{blue!10} \bf 0.620
 \\ 
 \hline

                & \multicolumn{18}{c}{\textbf{Root Mean Square Error (RMSE)}}                                                                                                                                                                        \\ \hline

\textbf{LIVE} &  8.613 & 6.935 & 6.581 & 7.527 & 7.440 & 11.299 & 7.114 & 7.546 & 7.282 & 7.205 & 8.572 & 10.852 & 9.050 & 6.807 &\cellcolor{blue!10} \bf 5.845 & 6.770 & 8.989 &\cellcolor{blue!10} \bf 5.915 \\

\textbf{MULTI}  
& 12.738 & 11.320 & 10.785 & 11.024 & 11.275 & 18.862 & 10.049 &\cellcolor{blue!10} \bf 8.686 & 10.866 & 10.794 & 15.058 & 12.744 & 17.419 & 9.898 & 9.895 & 9.258 & 14.806 &\cellcolor{blue!10} \bf 8.212 \\

\textbf{TID13}    & 0.879 & 0.652 & 0.697 & 0.762 & 0.702 & 1.207 & 0.688 &\cellcolor{blue!10} \bf 0.619 & 0.710 & 0.687 & 1.100 & 1.108 & 1.092 & 0.643 & 0.647 &\cellcolor{blue!10} \bf 0.615 & 1.049  & 0.630 
 \\ \hline

               & \multicolumn{18}{c}{\textbf{Pearson Linear Correlation Coefficient (PLCC)}}                                                                                                                                                                        \\ \hline

{\bf LIVE} 
& 0.928 & 0.954 & 0.959 & 0.945 & 0.947 & 0.872 & 0.951 & 0.945 & 0.949 & 0.950 & 0.929 & 0.883 & 0.920 & 0.956 &\cellcolor{blue!10} \bf 0.968 & 0.956 & 0.921 &\cellcolor{blue!10} \bf 0.967 \\ 
& -1 & -1 & -1 & -1 & -1 & -1 & -1 & -1 & -1 & -1 & -1 & -1 & -1 & -1 & 0 & -1 & -1  &  

 \\

\multirow{2}{*}{{\bf MULTI}}                                  
& 0.739 & 0.801 & 0.821 & 0.813 & 0.803 & 0.380 & 0.847 &  \cellcolor{blue!10} \bf 0.888 & 0.818 & 0.821 & 0.605 & 0.739 & 0.389 & 0.852 & 0.852 &  0.872 & 0.622 &\cellcolor{blue!10} \bf  0.901 \\                                          
& -1 & -1 & -1 & -1 & -1 & -1 & -1 & 0 & -1 & -1 & -1 & -1 & -1 & -1 & -1 & -1 & -1  & 
 \\

\multirow{2}{*}{{\bf TID13}}                               
& 0.705 & 0.851 & 0.827 & 0.789 & 0.830 & 0.227 & 0.832 &\cellcolor{blue!10} \bf 0.866 & 0.820 & 0.832 & 0.461 & 0.449 & 0.473 & 0.855 & 0.853 & \cellcolor{blue!10} \bf 0.869 & 0.533  &  0.861\\
& -1 & 0 & -1 & -1 & -1 & -1 & -1 & 0 & -1 & -1 & -1 & -1 & -1 & 0 & 0 & 0 & -1 &   \\ \hline

\textbf{}      & \multicolumn{18}{c}{\textbf{Spearman's Rank Correlation Coefficient (SRCC)}}                                                                                                                                                                        \\ \hline

 {\bf LIVE} & 0.909 & 0.938 & 0.944 & 0.950 & 0.951 & 0.903 & \cellcolor{blue!10} \bf 0.960 & 0.956 &\cellcolor{blue!10} \bf 0.961 & 0.960 & 0.940 & 0.897 & 0.923 & 0.950 & 0.959 & 0.952 & 0.886  & 0.959  \\
& -1 & -1 & -1 & -1 & -1 & -1 & 0 & 0 & 0 & 0 & -1 & -1 & -1 & -1 & 0 & 0 & -1  &  

\\
       
\multirow{2}{*}{{\bf MULTI}}                                          
& 0.677 & 0.715 & 0.743 & 0.860 & 0.836 & 0.631 & \cellcolor{blue!10} \bf 0.884 & 0.867 & 0.864 & 0.867 & 0.598 & 0.611 & 0.386 & 0.818 & 0.849 &  0.867 & 0.554  & \cellcolor{blue!10} \bf 0.884 \\   
& -1 & -1 & -1 & 0 & -1 & -1 &  0 & 0 & 0 & 0 & -1 & -1 & -1 & -1 & -1 & 0 & -1  &  \\

\multirow{2}{*}{{\bf TID13}}
& 0.701 & 0.847 & 0.817 & 0.742 & 0.786 & 0.563 & 0.778 & 0.807 & 0.802 & 0.851 & 0.414 & 0.393 & 0.396 &0.854 & 0.846 & \cellcolor{blue!10} \bf 0.860 & 0.649  & \cellcolor{blue!10} \bf 0.856 \\
& -1 & 0 & -1 & -1 & -1 & -1 & -1 & -1 & -1 & 0 & -1 & -1 & -1 & 0 & 0 & 0 & -1 &  \\ \hline

\textbf{}      & \multicolumn{18}{c}{\textbf{Kendall's Rank Correlation Coefficient (KRCC)}}                                                                                                                                                                        \\ \hline

\multirow{2}{*}{{\bf LIVE}}
& 0.748 & 0.791 & 0.803 & 0.815 & 0.818 & 0.732 &\cellcolor{blue!10} \bf 0.838 & 0.819 & \cellcolor{blue!10} \bf 0.838 &0.837 & 0.786 & 0.720 & 0.761 & 0.816 & 0.834 & 0.819 & 0.719 &  0.833                                         \\ & -1 & -1 & -1 & 0 & 0 & -1 & 0 & 0 & 0 & 0 & -1 & -1 & -1 & 0 & 0 & 0 & -1  &  
 \\

\multirow{2}{*}{{\bf MULTI}}                                              
& 0.500 & 0.532 & 0.559 & 0.669 & 0.644 & 0.457 & \cellcolor{blue!10} \bf 0.702 & 0.678 & 0.673 & 0.677 & 0.420 & 0.440 & 0.268 & 0.624 & 0.655 &  0.679 & 0.399 & \cellcolor{blue!10} \bf 0.698 \\                                         
& -1 & -1 & -1 & 0 & 0 & -1 & 0 & 0 & 0 & 0 & -1 & -1 & -1 & -1 & 0 & 0 & -1 & \\  

\multirow{2}{*}{{\bf TID13}}                                          
& 0.516 & 0.666 & 0.630 & 0.559 & 0.605 & 0.404 & 0.598 & 0.641 & 0.629 &\cellcolor{blue!10} \bf 0.667 & 0.286 & 0.270 & 0.277 &\cellcolor{blue!10} \bf 0.678 & 0.654 & \cellcolor{blue!10} \bf 0.667 & 0.474   &\cellcolor{blue!10} \bf 0.667 \\
& -1 & 0 & -1 & -1 & -1 & -1 & -1 & 0 & -1 & 0 & -1 & -1 & -1 & 0 & 0 & 0 & -1  &  
\\ \hline

\end{tabular}

\begin{tablenotes}
\item[] \textbf{[Sources Codes]} \textbf{PerSIM, CSV, UNIQUE, COHERENSI, SUMMER:} \url{https://ghassanalregib.com/publications/}, \textbf{PSNR-HA,PSNR-HMA:} \url{http://www.ponomarenko.info/psnrhma.m}, \textbf{SSIM, MS-SSIM, BRISQUE, BIQI, BLIINDS2:} \url{http://live.ece.utexas.edu/research/quality/
index.htm} , \textbf{CW-SSIM:} \url{https://www.mathworks.com/matlabcentral/fileexchange/
43017-complex-wavelet-structural-similarity-index-cw-ssim}, \textbf{IW-SSIM:} \url{https://ece.uwaterloo.ca/
~z70wang/research/iwssim/}, \textbf{SR-SIM:} \url{https://github.com/Netflix/vmaf/blob/master/matlab/strred/
SR_SIM.m}, \textbf{FSIM,FSIMc:} \url{http://www4.comp.polyu.edu.hk/
~cslzhang/IQA/FSIM/FSIM.htm}.   
\end{tablenotes}
\end{threeparttable}

\end{adjustwidth}{}
\end{table*}

\section{Results}
\label{sec:results}
We report the overall performance in Section \ref{subsec:results_overall} and distortion-based performance in Section \ref{subsec:results_distortion}. In Section \ref{subsec:dist_scatter}, we analyze the distributional difference between subjective and objective scores as well as their scatter plot characteristics. We analyze the classification performance of top quality estimators in Section \ref{subsec:stat_analysis}. Finally, in Section \ref{subsec:comp_time}, we report the average computation time of quality estimators and discuss possible approaches to accelerate the execution. 

\begin{figure}[htbp!]
\begin{minipage}[b]{0.49\linewidth}
  \centering
\includegraphics[width=\linewidth, trim= 35mm 85mm 35mm 90mm]{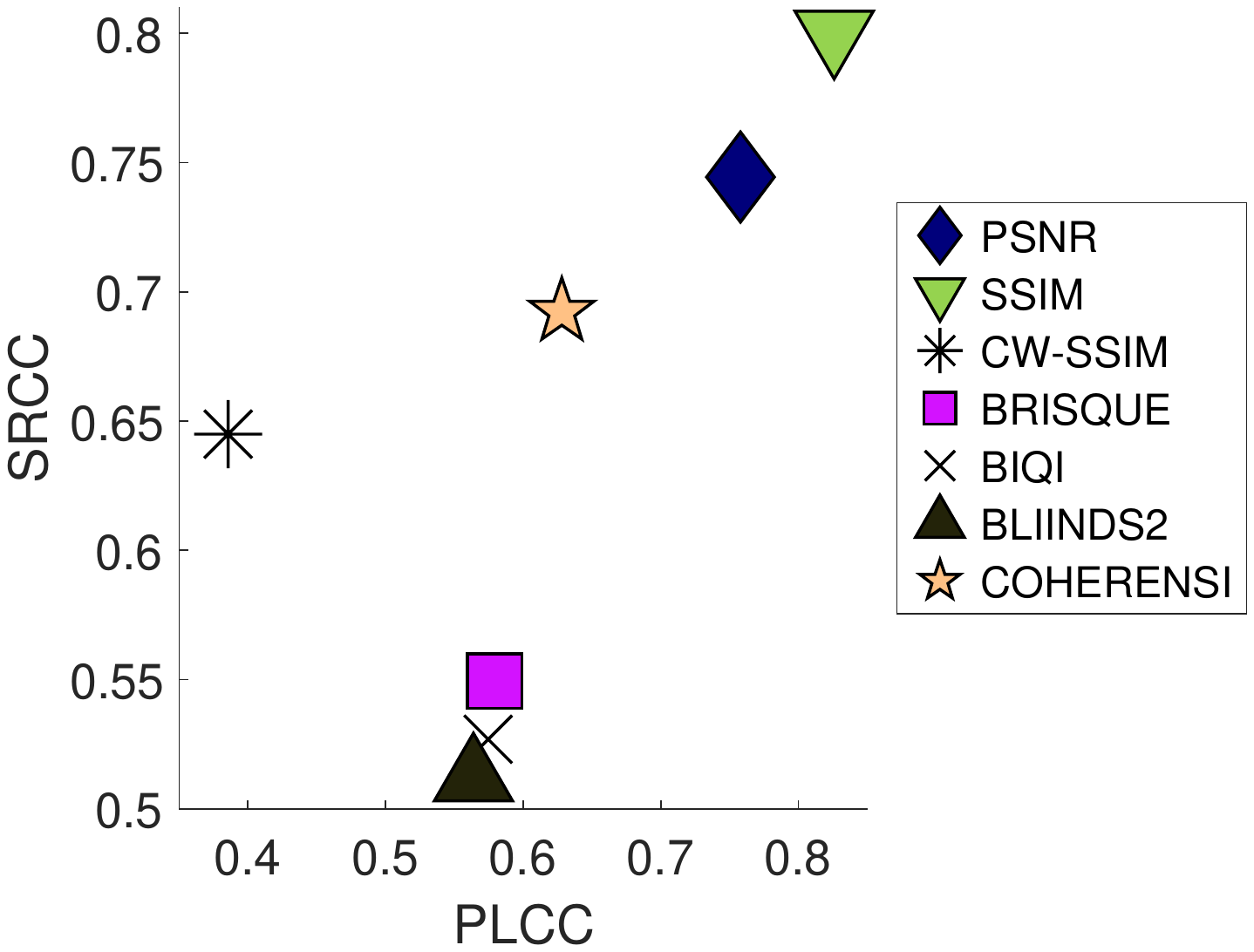}
  \vspace{0.03cm}
  \centerline{\scriptsize{(a)Quality assessment algorithms - lower quadrant}}
\end{minipage}
 \vspace{0.05cm}
\hfill
\begin{minipage}[b]{0.49\linewidth}
  \centering
\includegraphics[width=\linewidth, trim= 35mm 85mm 35mm 90mm]{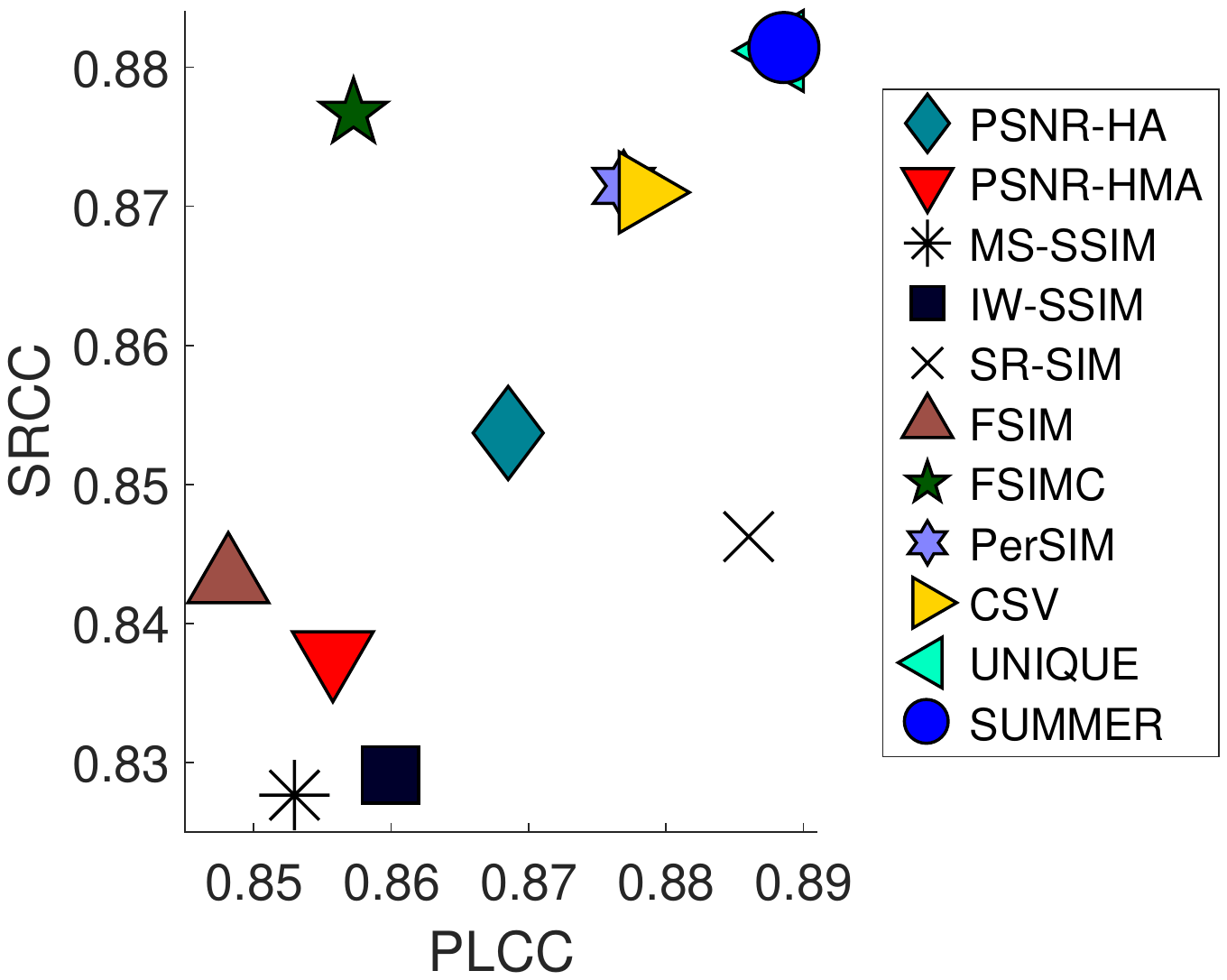}
  \vspace{0.03 cm}
  \centerline{\scriptsize{(b)Quality assessment algorithms - upper quadrant} }
\end{minipage}

\caption{Overall performance of image quality estimators in terms of Pearson (PLCC) and Spearman (SRCC) correlation in all tested databases (weighted average).}
\label{fig:scatter_all_weighted}
\end{figure}

\subsection{Overall}
\label{subsec:results_overall}
The performance of $18$ quality estimators including \texttt{SUMMER} over three databases is
summarized in Table \ref{tab_results_all}. We highlighted top two methods with a bold typeset and a light blue background. In case of  performance equivalence, we include all equivalent methods. Out of $14$ total categories, highlighted methods include \texttt{SUMMER} in  $10$ categories, SR-SIM and UNIQUE in $5$ categories, CSV in $3$ categories, PerSIM and PSNR-HA in $1$ category. We also measure the statistical significance of the difference between the performance of \texttt{SUMMER} and benchmarked methods in terms of correlation. We report the results of these statistical significance tests under correlation values of existing methods. A $0$ corresponds to statistically similar performance, $-1$ implies that compared method is statistically inferior, and $+1$ means that compared method is statistically superior. \texttt{SUMMER} statistically outperforms all the quality estimators in at least two categories and none of these methods statistically outperform \texttt{SUMMER} in any category. We observe that \texttt{SUMMER} statistically outperforms COHERENSI in all correlation categories. 

To illustrate the relative performance of image quality estimators, we computed weighted averages of their performance in terms of Pearson and Spearman correlations as shown in Fig.\ref{fig:scatter_all_weighted} in which the x-axis corresponds to the Pearson correlation and the y-axis corresponds to the Spearman correlation. Weighted averages were obtained by calculating the estimated quality in each database and weighing database performance values with the number of images in each database divided by total number of images in the validation. It was not possible to distinguish markers clearly when all quality estimators were shown in a single scatter plot. Therefore, we separated them into two scatter plots as upper quadrant and lower quadrant. Fig.\ref{fig:scatter_all_weighted}(a) includes PSNR, SSIM, CW-SSIM, BRISQUE, BIQI, BLIINDS2, and COHERENSI whereas Fig.\ref{fig:scatter_all_weighted}(b) includes PSNR-HA, PSNR-HMA, MS-SSIM, IW-SSIM, SR-SIM, FSIM, FSIMc, PerSIM, CSV, UNIQUE, and  \texttt{SUMMER}. COHERENSI is close to the center of the lower quadrant whereas \texttt{SUMMER} in on the top right of the higher quadrant.

\subsection{Distortion Categories}
\label{subsec:results_distortion}

\begin{table*}[htb!]
\tiny
\begin{adjustwidth}{-2.0cm}{}

\centering
\caption{Distortion category-based performance of image quality estimators.}
\label{tab_dist_all}

\begin{tabular}{p{1.3cm}p{0.45cm}p{0.45cm}p{0.45cm}p{0.45cm}p{0.45cm}p{0.45cm}p{0.45cm}p{0.45cm}p{0.45cm}p{0.45cm}p{0.45cm}p{0.45cm}p{0.45cm}p{0.45cm}p{0.45cm}p{0.45cm}p{0.55cm}p{0.80cm}}
\hline


\multirow{2}{*}{{\bf Databases}}                 &\bf PSNR  & \bf PSNR  &\bf PSNR  &\bf SSIM  &\bf MS  &\bf CW &\bf IW  &\bf SR  &\bf FSIM  &\bf FSIMc  &\bf BRIS  & \bf BIQI &\bf BLII &\bf Per &\bf CSV &\bf UNI &\bf COHER &\bf SUMMER \\& &\bf HA &\bf HMA & &\bf SSIM  &\bf SSIM  &\bf SSIM  &\bf SIM  & &&\bf QUE & &\bf NDS2 &\bf SIM & & \bf QUE & \bf ENSI &\bf   \\\hline

                & \multicolumn{18}{c}{\textbf{ Outlier Ratio (OR)}}                                                                                                                                                                        \\ \hline
{\bf Compression} & 0.475 & \cellcolor{blue!10} \bf 0.369 & 0.402 & 0.477 & 0.490 & 0.637 & 0.463 & 0.400 & 0.492 & 0.490 & 0.587 & 0.591 & 0.612 &  0.396 & 0.434 & 0.405 & 0.622  &\cellcolor{blue!10} \bf 0.359 \\                                       
{\bf Noise} & 0.616 & \cellcolor{blue!10} \bf 0.508 & 0.562 & 0.620 & 0.644 & 0.765 & 0.603 & 0.511 & 0.623 & 0.603 & 0.745 & 0.746 & 0.740 & 0.515 & 0.562 & 0.534 & 0.701  &\cellcolor{blue!10} \bf 0.454 
\\
{\bf Communication} & 0.860 & \cellcolor{blue!10} \bf 0.680 & 0.700 & 0.716 & 0.700 & 0.864 & 0.712 & 0.716 & 0.748 & 0.744 & 0.856 & 0.880 & 0.868 & 0.796 & 0.828 & \cellcolor{blue!10} \bf 0.676 & 0.812 & 0.812 
\\ 
{\bf Blur} & 0.292 & \cellcolor{blue!10} \bf 0.246 & 0.254 & 0.299 & 0.323 & 0.416 & 0.285 &\cellcolor{blue!10} \bf 0.246 & 0.335 & 0.321 & 0.389 & 0.359 & 0.395 & 0.265 & 0.313 & 0.293 & 0.382 & 0.272 
\\ 
{\bf Color} & 0.688 & \cellcolor{blue!10} \bf  0.672 & 0.704 & 0.747 & 0.773 & 0.840 & 0.736 & 0.736 & 0.784 & 0.789 & 0.861 & 0.840 & 0.867 & 0.728 & 0.675 &\cellcolor{blue!10} \bf 0.635 & 0.904  &  0.712 
\\ 
{\bf Global} & 0.820 &\cellcolor{blue!10} \bf 0.608 & 0.684 & 0.804 & 0.724 & 0.888 & 0.676 & 0.676 & 0.732 & 0.724 & 0.868 & 0.880 & 0.852 &\cellcolor{blue!10} \bf 0.656 & 0.700 & 0.700 & 0.752  & 0.688 
\\ 
{\bf Local} &\cellcolor{blue!10} \bf 0.712 & 0.808 & 0.872 & 0.880 & 0.832 & 0.780 & 0.776 &\cellcolor{blue!10} \bf 0.680 & 0.848 & 0.836 & 0.904 & 0.900 & 0.908 & 0.840 & 0.784 &\cellcolor{blue!10} \bf 0.712 & 0.900 & 0.772 
\\ \hline

                & \multicolumn{18}{c}{\textbf{ Root Mean Square Error (RMSE)}}                                                                                                                                                                        \\ \hline
{\bf Compression} & 6.740 & 5.470 & 5.143 & 5.509 & 5.647 & 9.850 & 5.202 & 5.050 & 5.372 & 5.329 & 6.485 & 7.931 & 7.577 &\cellcolor{blue!10} \bf 4.752 & 4.892 & 4.987 & 7.615  &\cellcolor{blue!10} \bf 4.283 
\\                                       
{\bf Noise} & 2.717 & 2.267 & 2.230 & 2.273 & 2.279 & 3.708 & 2.197 & 2.113 & 2.345 & 2.296 & 3.443 & 3.180 & 3.549 & 2.121 & \cellcolor{blue!10} \bf 1.970 & 1.996 & 2.965  &\cellcolor{blue!10} \bf 1.832 
\\
{\bf Communication} & 3.613 & 2.896 &\cellcolor{blue!10} \bf 2.662 & 3.891 & 3.781 & 4.544 & 3.211 & 2.924 & 3.172 & 3.319 & 3.733 & 4.591 & 3.809 & 3.392 & 2.750 & 3.484 & 4.105  &\cellcolor{blue!10} \bf 2.725\\
 
{\bf Blur} & 7.944 & 6.945 & 6.712 & 6.992 & 6.994 & 11.284 & 6.464 & 5.984 & 6.987 & 6.873 & 9.360 & 8.568 & 10.638 & 6.348 & 5.832 &\cellcolor{blue!10} \bf 5.785 & 8.846  &\cellcolor{blue!10} \bf 5.262 \\ 

{\bf Color} & 0.674 &\cellcolor{blue!10} \bf  0.642 & 0.687 & 0.948 & 0.886 & 1.103 & 0.951 & 0.958 & 0.899 & 0.818 & 1.094 & 1.046 & 1.025 & 0.732 &\cellcolor{blue!10} \bf 0.611 & 0.647 & 1.269 & 0.719\\
 
{\bf Global} & 1.656 & 0.749 & 0.871 & 1.056 & 0.935 & 1.401 &\cellcolor{blue!10} \bf 0.865 & 0.898 & 0.898 & 0.880 & 1.242 & 1.234 & 1.209 & 0.944 & 0.944 &\cellcolor{blue!10} \bf 0.847 & 1.084 & 0.878\\ 

{\bf Local} & 0.837 & 1.145 & 1.210 & 1.131 & 0.804 & 0.957 & 0.820 &\cellcolor{blue!10} \bf 0.633 & 0.871 & 0.883 & 1.161 & 1.157 & 1.133 & 0.993 & 0.809 &\cellcolor{blue!10} \bf 0.743 & 1.304 & 0.988\\ \hline

                   & \multicolumn{18}{c}{\textbf{ Pearson Linear Correlation Coefficient (PLCC)}}                                                                                                                                                                        \\ \hline
{\bf Compression} & 0.879 & 0.932 & 0.941 & 0.920 & 0.921 & 0.619 & 0.931 &\cellcolor{blue!10} \bf 0.948 & 0.918 & 0.920 & 0.869 & 0.767 & 0.742 & 0.947 & 0.944 & 0.940 & 0.815 &\cellcolor{blue!10} \bf 0.953 
\\                                       
{\bf Noise} & 0.793 &\cellcolor{blue!10} \bf 0.909 & 0.904 & 0.849 & 0.849 & 0.502 & 0.853 & 0.898 & 0.849 & 0.857 & 0.475 & 0.456 & 0.408 & 0.903 & 0.866 & 0.879 & 0.779  &\cellcolor{blue!10} \bf 0.932 
\\
{\bf Communication} & 0.851 & 0.882 & 0.876 &\cellcolor{blue!10} \bf 0.928 & 0.925 & 0.472 & 0.915 &\cellcolor{blue!10} \bf 0.934 & 0.887 & 0.888 & 0.505 & 0.507 & 0.595 & 0.901 & 0.898 & 0.922 & 0.919  & 0.896 
\\ 

{\bf Blur} & 0.831 & 0.878 & 0.890 & 0.885 & 0.889 & 0.466 & 0.904 &\cellcolor{blue!10} \bf 0.924 & 0.881 & 0.883 & 0.750 & 0.768 & 0.591 & 0.902 & 0.916 & 0.918 & 0.750 &\cellcolor{blue!10} \bf 0.936 
\\ 

{\bf Color} & 0.844 & 0.841 & 0.814 & 0.670 & 0.682 & 0.356 & 0.676 & 0.672 & 0.674 & 0.727 & 0.484 & 0.448 & 0.495 & 0.801 &\cellcolor{blue!10} \bf 0.887 &\cellcolor{blue!10} \bf 0.846 & 0.232 &0.800 
\\ 

{\bf Global} & 0.343 & 0.744 &\cellcolor{blue!10} \bf 0.763 & 0.561 & 0.665 & 0.653 & 0.652 & 0.571 & 0.649 & 0.647 & 0.008 & 0.174 & 0.028 & 0.586 & 0.410 & 0.512 & 0.402 &\cellcolor{blue!10} \bf 0.835 
\\ 

{\bf Local} & 0.648 & 0.688 & 0.733 & 0.236 & 0.642 & 0.375 & 0.698 &\cellcolor{blue!10} \bf 0.831 & 0.701 & 0.705 & 0.040 & 0.057 & 0.143 & 0.447 &\cellcolor{blue!10} \bf 0.872 & 0.720 & 0.147 & 0.794 
\\ \hline
 
                    & \multicolumn{18}{c}{\textbf{  Spearman's Rank Correlation Coefficient (SRCC)}}                                                                                                                                                                        \\ \hline
{\bf Compression} & 0.871 & 0.910 & 0.918 & 0.938 & 0.937 & 0.841 & 0.943 &\cellcolor{blue!10} \bf 0.950 & 0.948 &\cellcolor{blue!10} \bf 0.950 & 0.854 & 0.743 & 0.728 & 0.935 & 0.938 & 0.934 & 0.805  & 0.946 
\\                                       
{\bf Noise} & 0.774 & 0.896 & 0.885 & 0.862 & 0.865 & 0.798 & 0.871 & 0.902 & 0.893 & 0.901 & 0.481 & 0.451 & 0.378 &\cellcolor{blue!10} \bf 0.917 & 0.854 & 0.875 & 0.790 &\cellcolor{blue!10} \bf 0.927
\\
{\bf Communication} & 0.863 & 0.891 & 0.886 & 0.931 &  0.934 & 0.786 & 0.916 &\cellcolor{blue!10} \bf 0.940 & 0.930 & 0.934 & 0.536 & 0.610 & 0.655 & 0.918 & 0.906 & 0.933 &\cellcolor{blue!10} \bf 0.944  & 0.899 
\\ 
{\bf Blur} & 0.803 & 0.841 & 0.856 & 0.905 & 0.901 & 0.773 &\cellcolor{blue!10} \bf 0.917 & 0.913 & 0.912 & 0.913 & 0.740 & 0.722 & 0.618 & 0.892 & 0.909 & 0.909 &  0.731&\cellcolor{blue!10} \bf 0.919 
\\ 
{\bf Color} & 0.815 & 0.805 & 0.778 & 0.235 & 0.239 & 0.372 & 0.234 & 0.243 & 0.241 & 0.629 & 0.401 & 0.328 & 0.368 & 0.762 &\cellcolor{blue!10} \bf 0.886 &\cellcolor{blue!10} \bf 0.909 & 0.238& 0.749 
\\ 
{\bf Global} & 0.340 &  0.612 &\cellcolor{blue!10} \bf 0.672 & 0.499 & 0.458 & 0.326 & 0.453 & 0.393 & 0.441 & 0.440 & 0.036 & 0.264 & 0.041 & 0.409 & 0.341 & 0.356 & 0.322  &\cellcolor{blue!10} \bf 0.670 
\\ 
{\bf Local} & 0.543 & 0.592 & 0.635 & 0.288 & 0.645 & 0.699 & 0.612 &\cellcolor{blue!10} \bf 0.810 & 0.702 & 0.705 & 0.043 & 0.048 & 0.247 & 0.471 &\cellcolor{blue!10} \bf 0.787 & 0.645 & 0.137& 0.724
\\ \hline

                    & \multicolumn{18}{c}{\textbf{Kendall's Rank Correlation Coefficient (KRCC)}}                                                                                                                                                                        \\ \hline
{\bf Compression} & 0.701 & 0.764 & 0.774 & 0.796 & 0.798 & 0.667 & 0.806 &\cellcolor{blue!10} \bf 0.817 &\cellcolor{blue!10} \bf 0.818 & 0.821 & 0.687 & 0.570 & 0.569 & 0.798 & 0.803 & 0.790 & 0.627 & 0.810 
\\                                       
{\bf Noise} & 0.576 & 0.730 & 0.710 & 0.671 & 0.676 & 0.607 & 0.683 & 0.727 & 0.713 & 0.725 & 0.341 & 0.318 & 0.272 &\cellcolor{blue!10} \bf 0.753 & 0.675 & 0.691 & 0.595&\cellcolor{blue!10} \bf 0.764 
\\
{\bf Communication} & 0.721 & 0.750 & 0.748 & 0.780 & 0.788 & 0.651 & 0.758 &\cellcolor{blue!10} \bf 0.795 & 0.776 & 0.784 & 0.466 & 0.504 & 0.525 & 0.775 & 0.765 &\cellcolor{blue!10} \bf 0.795 &\cellcolor{blue!10} \bf 0.807  & 0.751 
\\ 
{\bf Blur} & 0.632 & 0.681 & 0.696 & 0.739 & 0.738 & 0.599 &\cellcolor{blue!10} \bf 0.759 & 0.752 & 0.752 & 0.754 & 0.562 & 0.539 & 0.461 & 0.729 & 0.752 & 0.745 & 0.560 &\cellcolor{blue!10} \bf 0.759
\\ 
{\bf Color} & 0.614 & 0.615 & 0.589 & 0.174 & 0.179 & 0.259 & 0.170 & 0.183 & 0.179 & 0.456 & 0.273 & 0.229 & 0.261 & 0.573 &\cellcolor{blue!10} \bf 0.712 &\cellcolor{blue!10} \bf 0.727 & 0.178 & 0.558
\\ 
{\bf Global} & 0.221 & 0.452 &\cellcolor{blue!10} \bf  0.503 & 0.350 & 0.369 & 0.229 & 0.364 & 0.318 & 0.359 & 0.357 & 0.024 & 0.185 & 0.031 & 0.334 & 0.264 & 0.262 & 0.259 &\cellcolor{blue!10} \bf  0.490 
\\ 
{\bf Local} & 0.371 & 0.409 & 0.447 & 0.200 & 0.462 & 0.494 & 0.426 &\cellcolor{blue!10} \bf 0.614 & 0.506 & 0.509 & 0.030 & 0.036 & 0.178 & 0.333 &\cellcolor{blue!10} \bf 0.582 & 0.457 & 0.055& 0.528 
\\ \hline

\end{tabular}
\end{adjustwidth}{}
\end{table*}

We report the distortion category-based performance of image quality estimators in Table \ref{tab_dist_all}. Distortion-based algorithmic performances were obtained from weighted averages of performances over different distortions in which weights were proportional to the number of images in each category. \texttt{SUMMER} is the best performing method in terms of all performance metrics in noise category. It is also the best method in all other categories other than color and local distortion in terms of at least one performance metric. There are $7$ main distortion categories and $5$ performance metrics. When we analyze these main distortion categories and performance metrics, there are $35$ main categories.  In each category, we highlighted top two methods with a bold typeset and a light blue background. In case of  performance equivalence, we include all equivalent methods. Out of total $35$ categories, highlighted methods include \texttt{SUMMER} in  $16$ categories, SR-SIM in $13$ categories, UNIQUE in $10$ categories, CSV and PSNR-HA in $8$ categories, PerSIM and PSNR-HMA in $4$ categories, COHERENSI in $2$ categories, FSIMc, FSIM, SSIM, and PSNR $1$ category.

\begin{table*}[htb!]
\tiny
\begin{adjustwidth}{-2.0cm}{}

\centering
\caption{Distributional difference between subjective score and objective scores.}
\label{tab_hist_dist}


\begin{tabular}{p{1.3cm}p{0.45cm}p{0.45cm}p{0.45cm}p{0.45cm}p{0.45cm}p{0.45cm}p{0.45cm}p{0.45cm}p{0.45cm}p{0.45cm}p{0.45cm}p{0.45cm}p{0.45cm}p{0.45cm}p{0.45cm}p{0.45cm}p{0.55cm}p{0.80cm}}
\hline


\multirow{3}{*}{{\bf Databases }}                 &\bf PSNR  & \bf PSNR  &\bf PSNR  &\bf SSIM  &\bf MS  &\bf CW &\bf IW  &\bf SR  &\bf FSIM  &\bf FSIMc  &\bf BRIS  & \bf BIQI &\bf BLII &\bf Per &\bf CSV &\bf UNI &\bf COHER &\bf SUMMER \\
& &\bf HA &\bf HMA & &\bf SSIM  &\bf SSIM  &\bf SSIM  &\bf SIM  & &&\bf QUE & &\bf NDS2 &\bf SIM & & \bf QUE & \bf ENSI& 
\\ \hline

                & \multicolumn{18}{c}{\textbf{ Earth Mover's Distance (EMD)}}                                                                                                                                                                        \\ \hline
\textbf{LIVE}   
& 0.26 & 0.24 &0.23 & 0.28 & 0.32 & 0.49 & 0.30 & 0.33 & 0.28 & 0.27 & 0.50 & 0.50 & 0.50 & 0.35 &\cellcolor{blue!10} \bf  0.20 & 0.24 & 0.31  &\cellcolor{blue!10} \bf 0.19 \\
\textbf{MULTI} & 0.32 & 0.43 & 0.44 & 0.44 & 0.42 & 0.83 & 0.42 & 0.40 & 0.47 & 0.46 & 0.32 & 0.66 & 0.62 & 0.46 &\cellcolor{blue!10} \bf 0.26 &\cellcolor{blue!10} \bf 0.31 & 0.35 & 0.47 
 \\
\textbf{TID13} & 0.55 &0.33 & 0.36 & 0.47 & 0.50 & 0.95 & 0.50 & 0.51 & 0.70 & 0.69 & 0.68 & 0.56 & 0.61 & 0.50 &\cellcolor{blue!10} \bf 0.17 & 0.41 & 0.69 & \cellcolor{blue!10} \bf 0.29
 \\ \hline

                & \multicolumn{18}{c}{\textbf{ Kullback-Leibler Divergence (KL)}}                                                                                                                                                                        \\ \hline
\textbf{LIVE}   
& 0.25 & 0.25 &0.21 & 0.30 & 0.35 & 1.00 & 0.33 & 0.39 & 0.30 & 0.30 & 0.92 & 1.00 & 0.86 & 0.58 &\cellcolor{blue!10} \bf 0.19 & 0.26 & 0.45  &\cellcolor{blue!10} \bf 0.15 \\ 
\textbf{MULTI} & 0.29 & 0.43 & 0.46 & 0.48 & 0.46 & 1.69 & 0.47 & 0.42 & 0.54 & 0.52 & 0.31 & 1.41 & 0.84 & 0.65 &\cellcolor{blue!10} \bf 0.16 &\cellcolor{blue!10} \bf 0.23 & 0.32& 0.67  \\
\textbf{TID13} & 1.37 & 0.82 & 0.93 & 1.25 & 1.56 & 5.94 & 1.68 & 1.62 & 2.67 & 2.54 & 2.62 & 1.63 & 2.09 & 1.25 &\cellcolor{blue!10} \bf 0.17 & 0.95 & 1.39 & \cellcolor{blue!10} \bf 0.43 
 \\ \hline              
                   & \multicolumn{18}{c}{\textbf{ Jensen-Shannon Divergence (JS)}}                                                                                                                                                                        \\ \hline
\textbf{LIVE}   
 & 0.06 & 0.06 & 0.05 & 0.07 & 0.08 & 0.17 & 0.07 & 0.09 & 0.07 & 0.07 & 0.18 & 0.19 & 0.15 & 0.11 & \cellcolor{blue!10} \bf 0.04 & 0.06 & 0.10 &\cellcolor{blue!10} \bf 0.04  \\ 
\textbf{MULTI}  & 0.06 & 0.10 & 0.10 & 0.10 & 0.10 & 0.30 & 0.10 & 0.09 & 0.12 & 0.11 & 0.06 & 0.22 & 0.18 & 0.12 &\cellcolor{blue!10} \bf 0.04 &\cellcolor{blue!10} \bf 0.05 & 0.07  & 0.13 \\
\textbf{TID13}  & 0.21 & 0.11 & 0.12 & 0.16 & 0.19 & 0.57 & 0.20 & 0.19 & 0.32 & 0.31 & 0.32 & 0.22 & 0.26 & 0.17 &\cellcolor{blue!10} \bf 0.03 & 0.13 & 0.29 & \cellcolor{blue!10} \bf 0.08 
 \\ \hline

                    & \multicolumn{18}{c}{\textbf{ Histogram Intersection (HI)}}                                                                                                                                                                        \\ \hline
\textbf{LIVE}   
 & 0.26 & 0.24 & 0.23 & 0.28 & 0.32 & 0.49 & 0.30 & 0.33 & 0.28 & 0.27 & 0.50 & 0.50 & 0.50 & 0.35 & \cellcolor{blue!10} \bf 0.20 & 0.24 & 0.31  &\cellcolor{blue!10} \bf 0.19 \\ 
\textbf{MULTI}  & 0.32 & 0.43 & 0.44 & 0.44 & 0.42 & 0.83 & 0.42 & 0.40 & 0.47 & 0.46 & 0.32 & 0.66 & 0.62 & 0.46 & \cellcolor{blue!10} \bf 0.26 &\cellcolor{blue!10} \bf 0.31 & 0.35  & 0.47 \\
\textbf{TID13}  & 0.55 & 0.33 & 0.36 & 0.47 & 0.50 & 0.95 & 0.50 & 0.51 & 0.70 & 0.69 & 0.68 & 0.56 & 0.61 & 0.50 &\cellcolor{blue!10} \bf 0.17 & 0.41 & 0.69 & \cellcolor{blue!10} \bf 0.29
 \\ \hline
 
                    & \multicolumn{18}{c}{\textbf{ L2 Norm (L2)}}                                                                                                                                                                        \\ \hline
\textbf{LIVE}   & 0.07 & 0.07 & 0.07 & 0.07 & 0.08 & 0.25 & 0.08 & 0.09 & 0.07 & 0.07 & 0.23 & 0.23 & 0.20 & 0.11 &\cellcolor{blue!10} \bf  0.06 & 0.07 & 0.09 &\cellcolor{blue!10} \bf  0.06 
 \\ 
\textbf{MULTI}  & 0.09 & 0.13 & 0.14 & 0.11 & 0.11 & 0.25 & 0.11 & 0.11 & 0.12 & 0.12 & 0.09 & 0.28 & 0.16 & 0.15 &\cellcolor{blue!10} \bf 0.07 &\cellcolor{blue!10} \bf 0.08 & 0.10  & 0.15 
  \\
\textbf{TID13}  & 0.15 & 0.11 & 0.12 & 0.14 & 0.17 & 0.83 & 0.18 & 0.17 & 0.26 & 0.24 & 0.28 & 0.17 & 0.23 & 0.14 &\cellcolor{blue!10} \bf 0.06 & 0.12 & 0.17 &\cellcolor{blue!10} \bf 0.09 
\\ \hline

\end{tabular}
\end{adjustwidth}{}
\end{table*}

\begin{figure}[htbp!]
\begin{minipage}[b]{0.31\linewidth}
  \centering
\includegraphics[width=\linewidth, trim= 40mm 90mm 40mm 85mm]{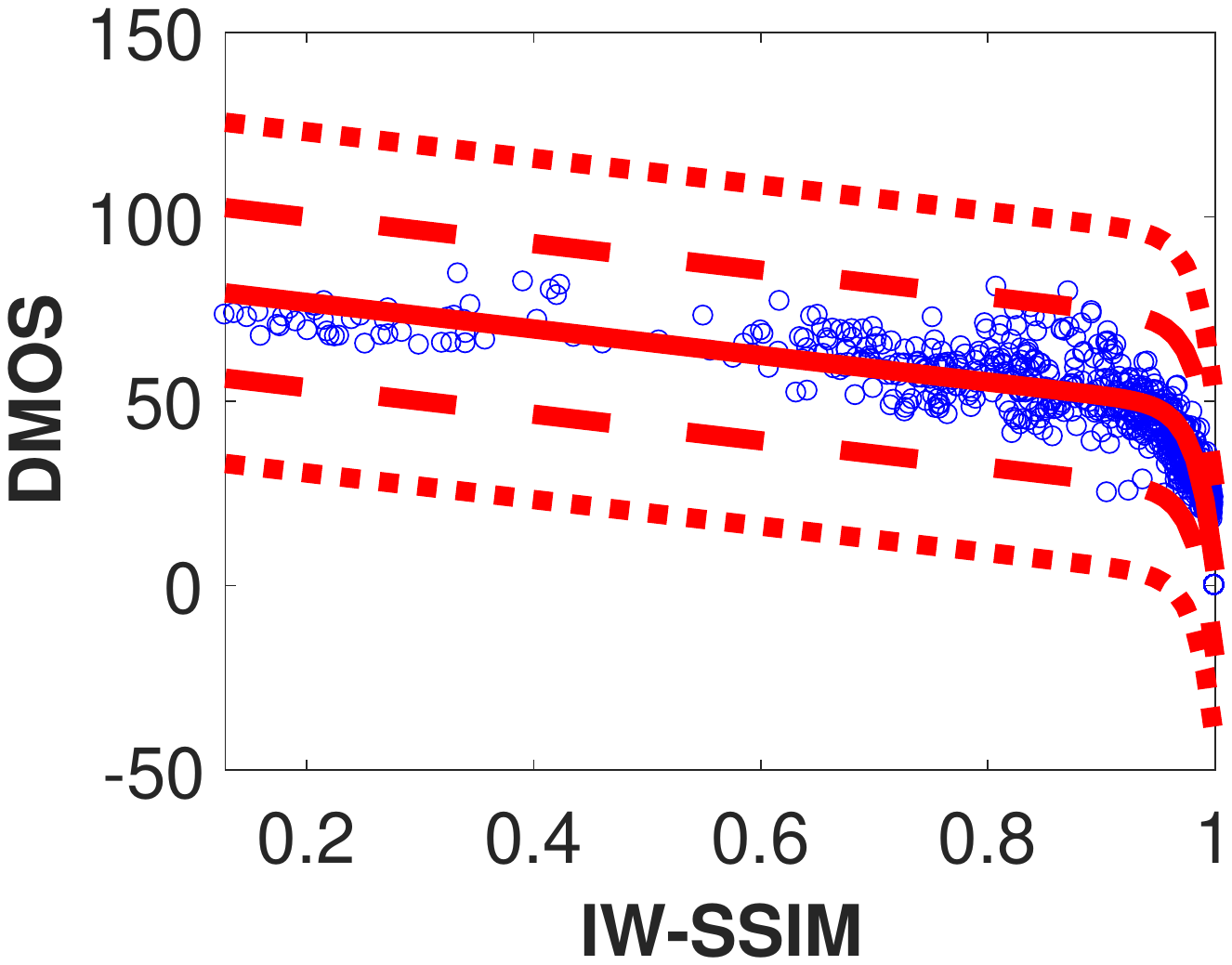}
  \vspace{0.03cm}
  \centerline{\scriptsize{(a)LIVE:IW-SSIM}}
\end{minipage}
 \vspace{0.05cm}
\begin{minipage}[b]{0.31\linewidth}
  \centering
\includegraphics[width=\linewidth, trim= 40mm 90mm 40mm 90mm]{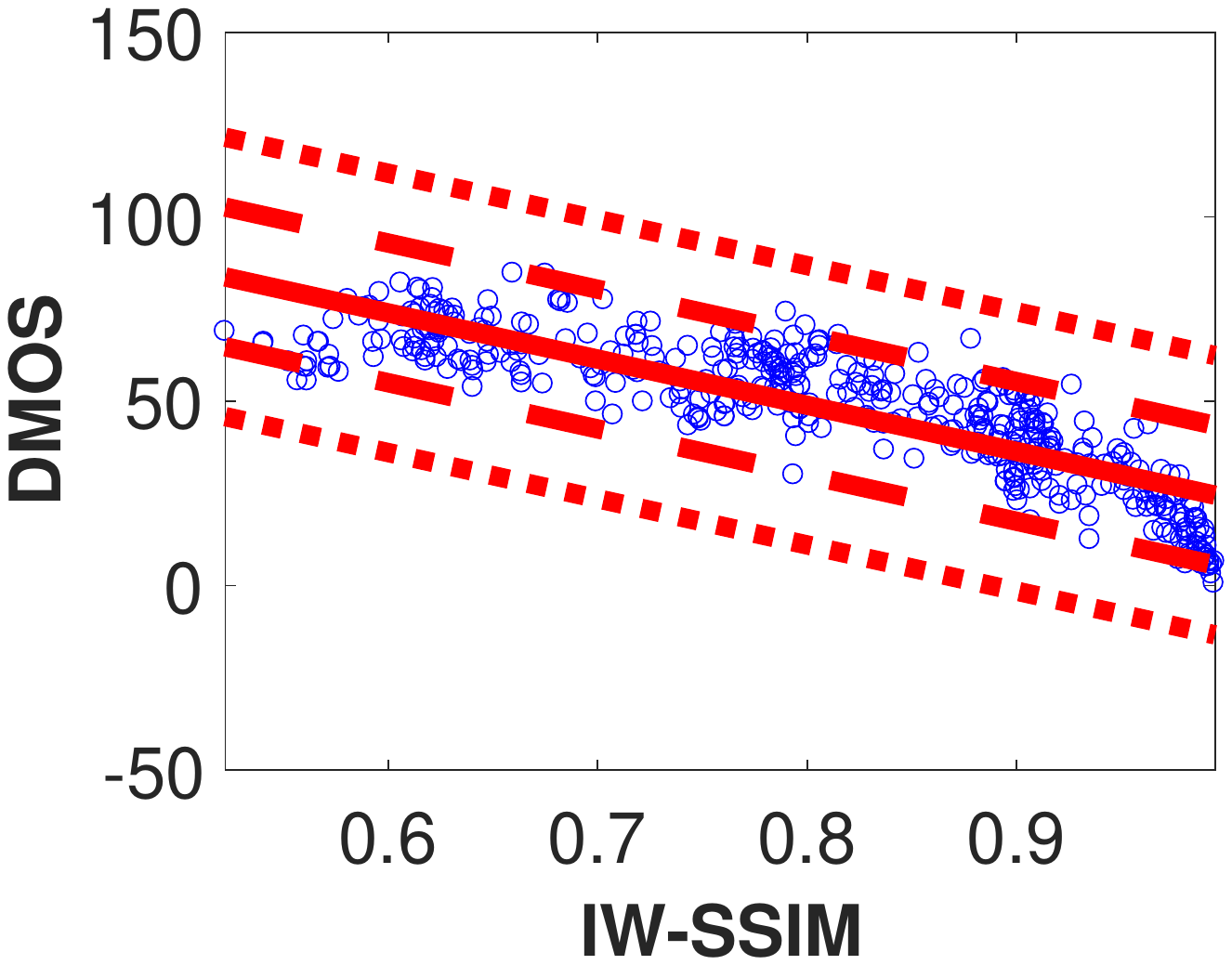}
  \vspace{0.03 cm}
  \centerline{\scriptsize{(b)MULTI:IW-SSIM} }
\end{minipage}
 \vspace{0.05cm}
\begin{minipage}[b]{0.31\linewidth}
  \centering
\includegraphics[width=\linewidth, trim= 40mm 90mm 40mm 90mm]{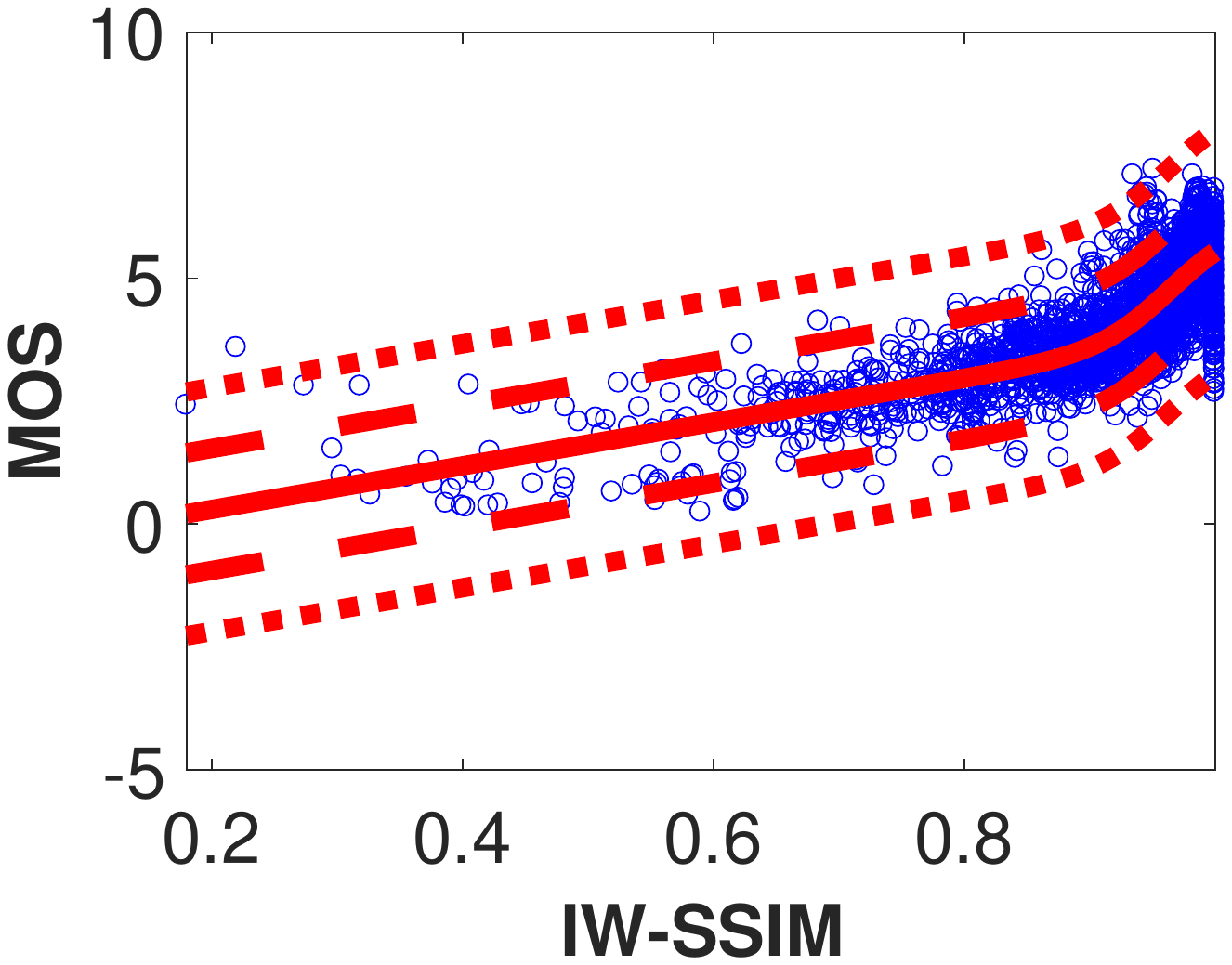}
  \vspace{0.03 cm}
  \centerline{\scriptsize{(c)TID13:IW-SSIM} }
\end{minipage}
 \vspace{0.05cm}
\begin{minipage}[b]{0.31\linewidth}
  \centering
\includegraphics[width=\linewidth, trim= 40mm 90mm 40mm 90mm]{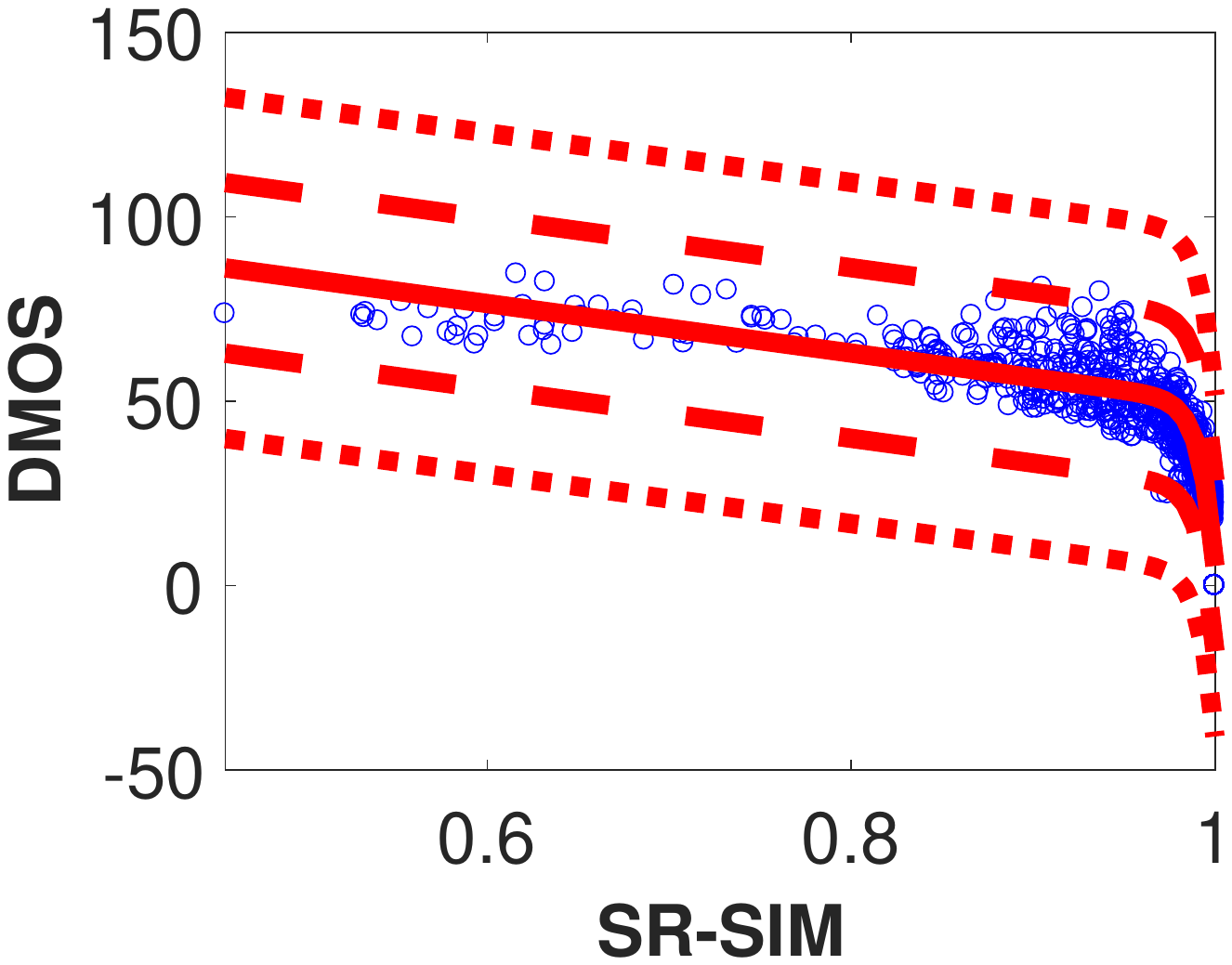}
  \vspace{0.03cm}
  \centerline{\scriptsize{(d)LIVE:SR-SIM}}
\end{minipage}
 \vspace{0.05cm}
\begin{minipage}[b]{0.31\linewidth}
  \centering
\includegraphics[width=\linewidth, trim= 40mm 90mm 40mm 90mm]{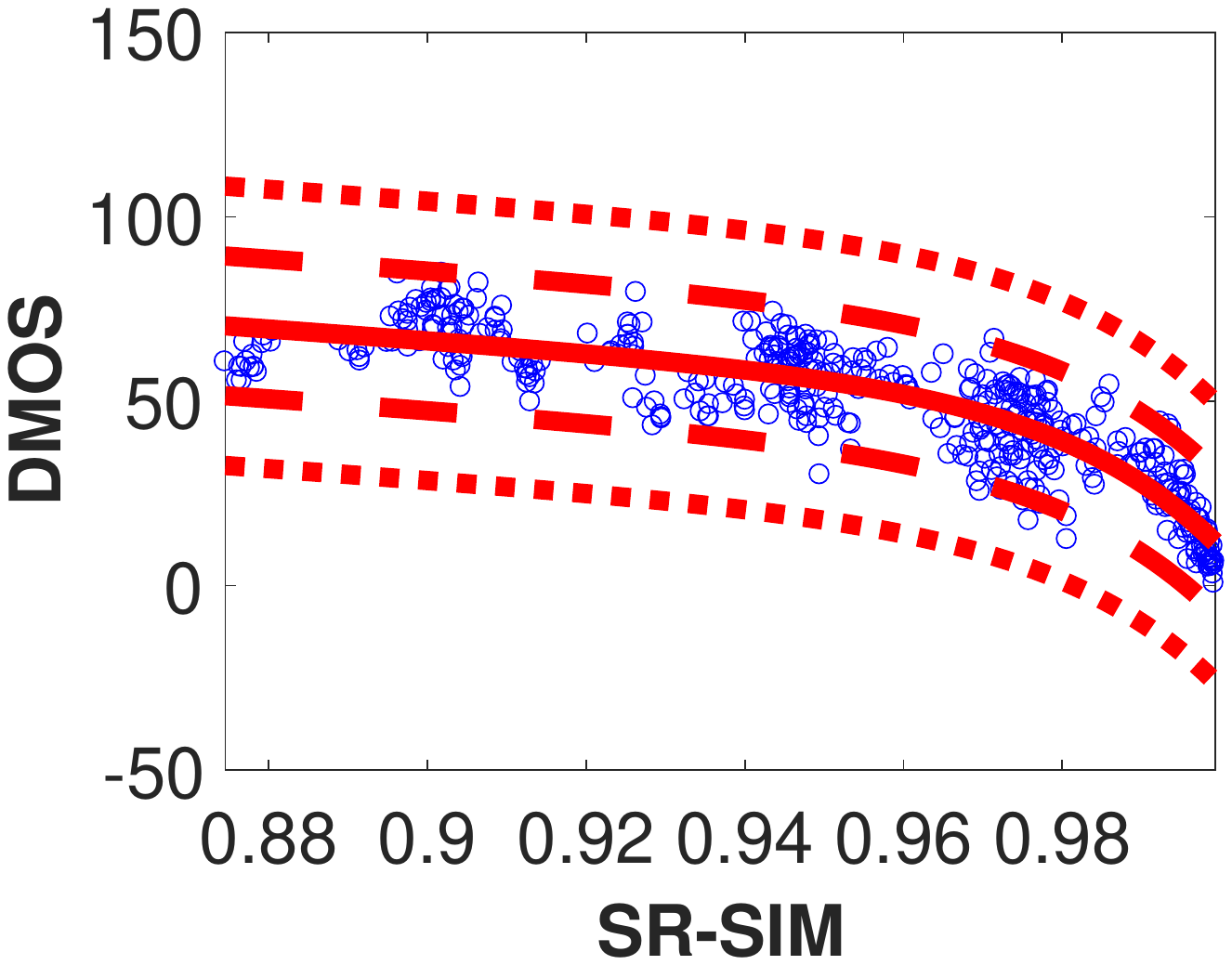}
  \vspace{0.03cm}
  \centerline{\scriptsize{(e)MULTI:SR-SIM}}
\end{minipage}
 \vspace{0.05cm}
\begin{minipage}[b]{0.31\linewidth}
  \centering
\includegraphics[width=\linewidth, trim= 40mm 90mm 40mm 90mm]{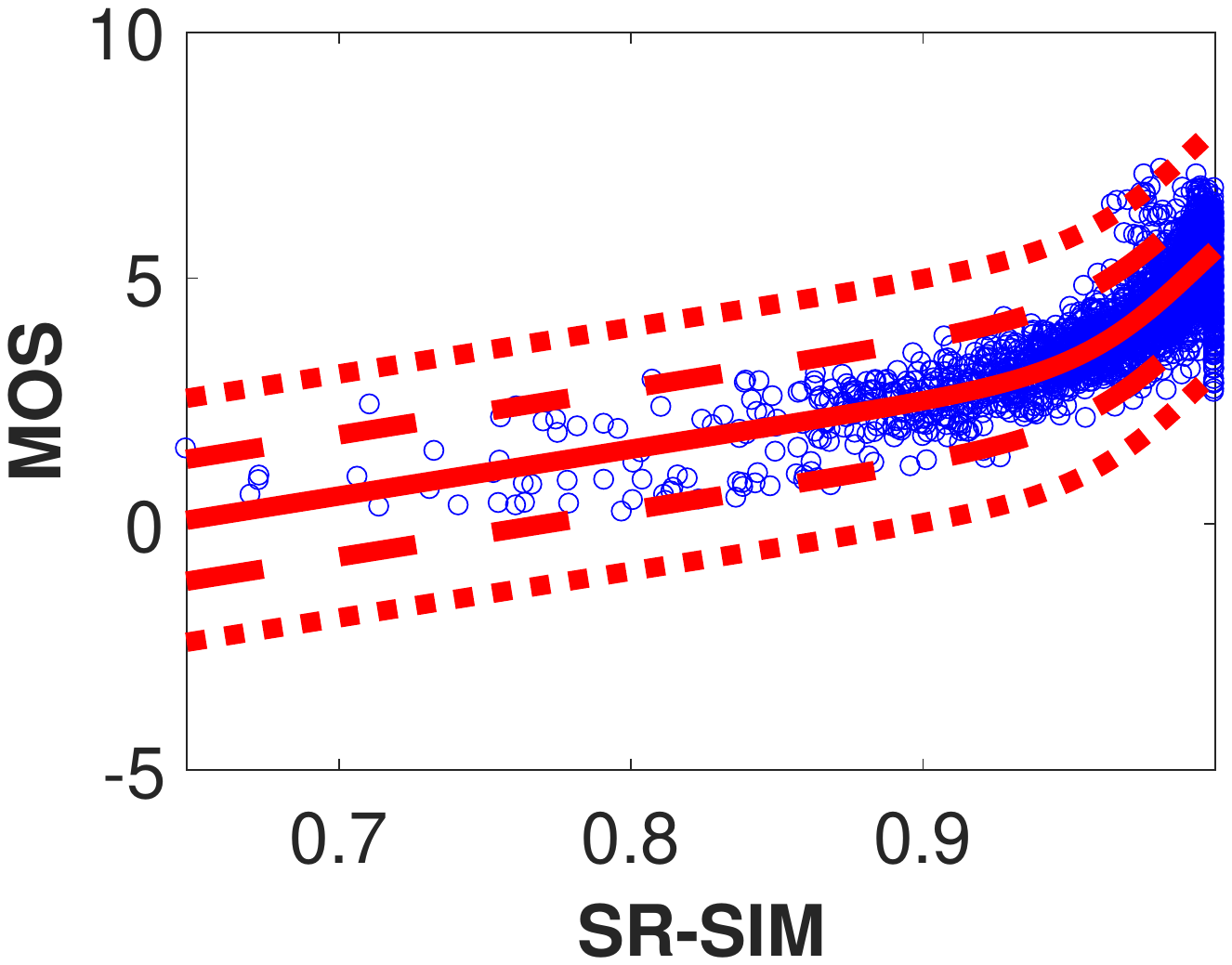}
  \vspace{0.03 cm}
  \centerline{\scriptsize{(f)TID13:SR-SIM} }
\end{minipage}
 \vspace{0.05cm}
\begin{minipage}[b]{0.31\linewidth}
  \centering
\includegraphics[width=\linewidth, trim= 40mm 90mm 40mm 90mm]{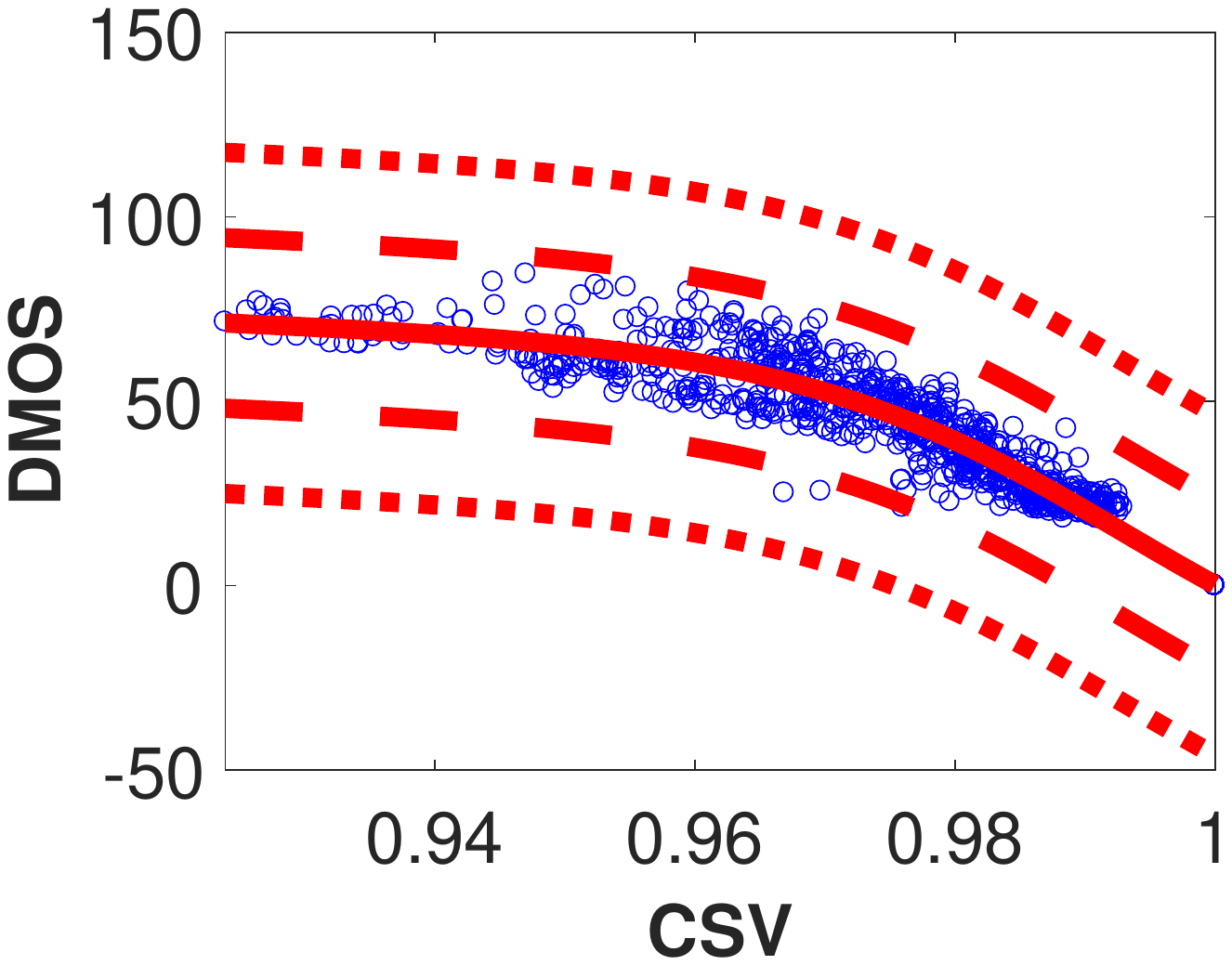}
  \vspace{0.03 cm}
  \centerline{\scriptsize{(g)LIVE:CSV} }
\end{minipage}
 \vspace{0.05cm}
\begin{minipage}[b]{0.31\linewidth}
  \centering
\includegraphics[width=\linewidth, trim= 40mm 90mm 40mm 90mm]{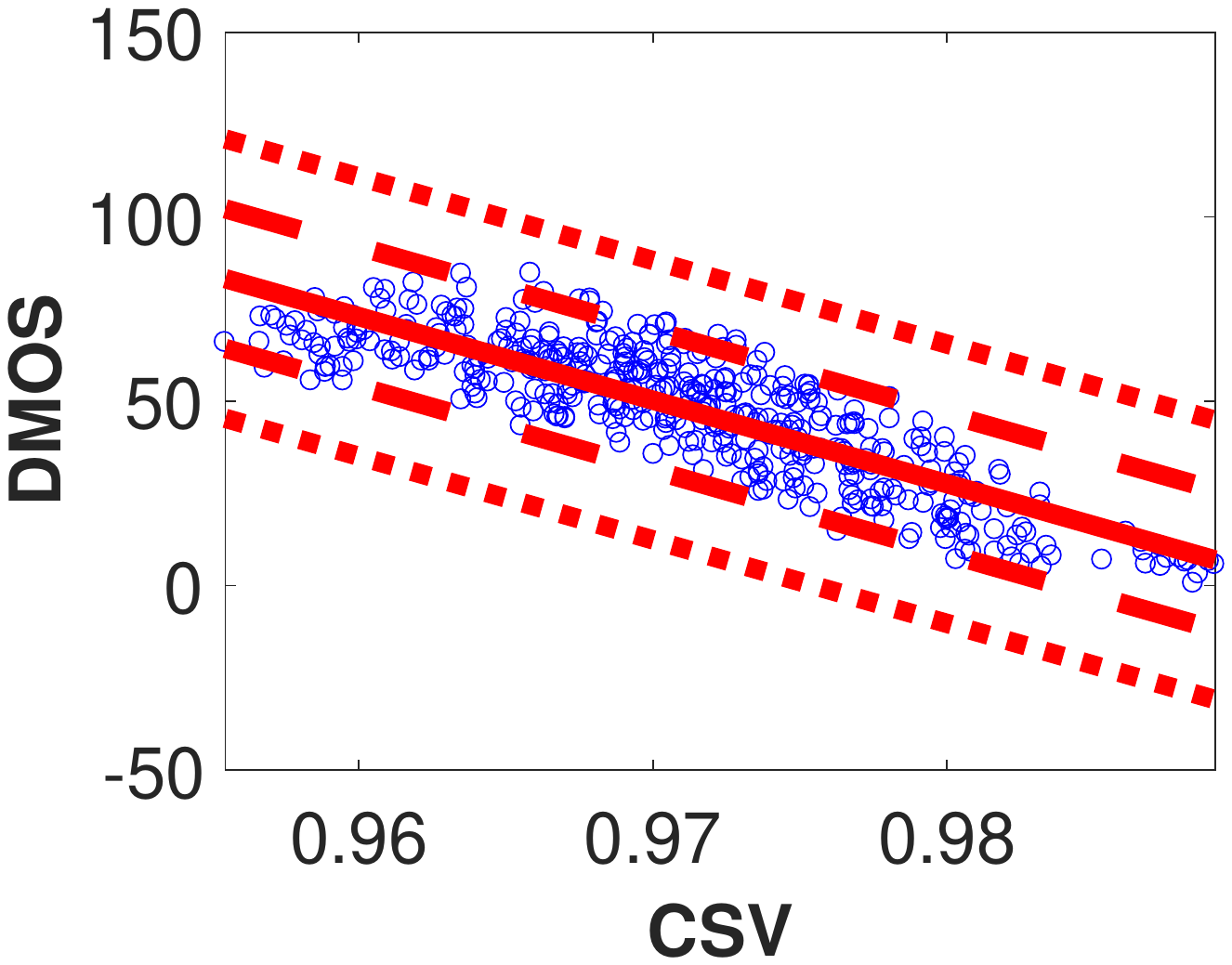}
  \vspace{0.03cm}
  \centerline{\scriptsize{(h)MULTI:CSV}}
\end{minipage}
 \vspace{0.05cm}
\begin{minipage}[b]{0.31\linewidth}
  \centering
\includegraphics[width=\linewidth, trim= 40mm 90mm 40mm 90mm]{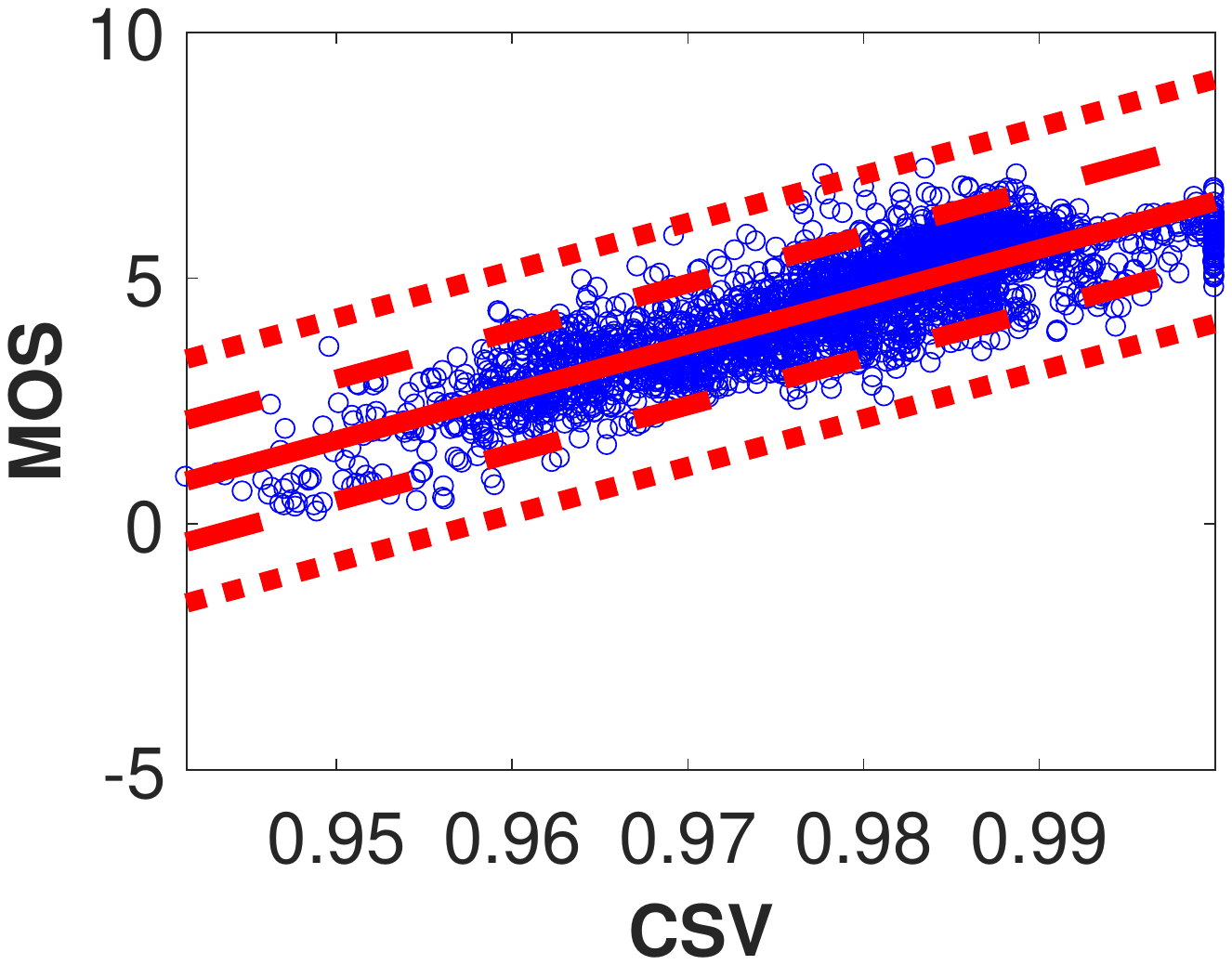}
  \vspace{0.03cm}
  \centerline{\scriptsize{(i)TID13:CSV}}
\end{minipage}
 \vspace{0.05cm}
\begin{minipage}[b]{0.31\linewidth}
  \centering
\includegraphics[width=\linewidth, trim= 40mm 90mm 40mm 90mm]{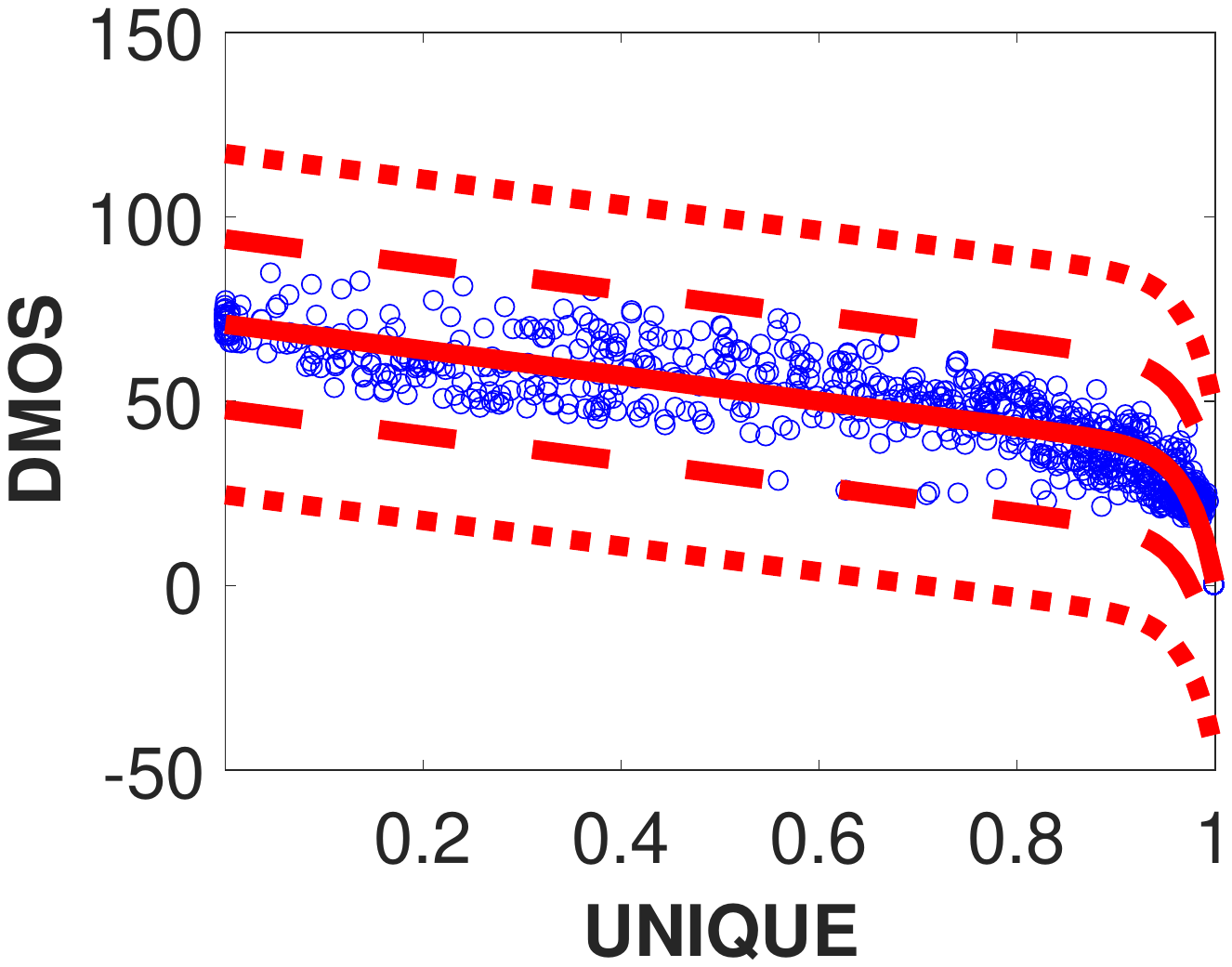}
  \vspace{0.03 cm}
  \centerline{\scriptsize{(j)LIVE:UNIQUE} }
\end{minipage}
\hfill 
 \vspace{0.05cm}
\begin{minipage}[b]{0.31\linewidth}
  \centering
\includegraphics[width=\linewidth, trim= 40mm 90mm 40mm 90mm]{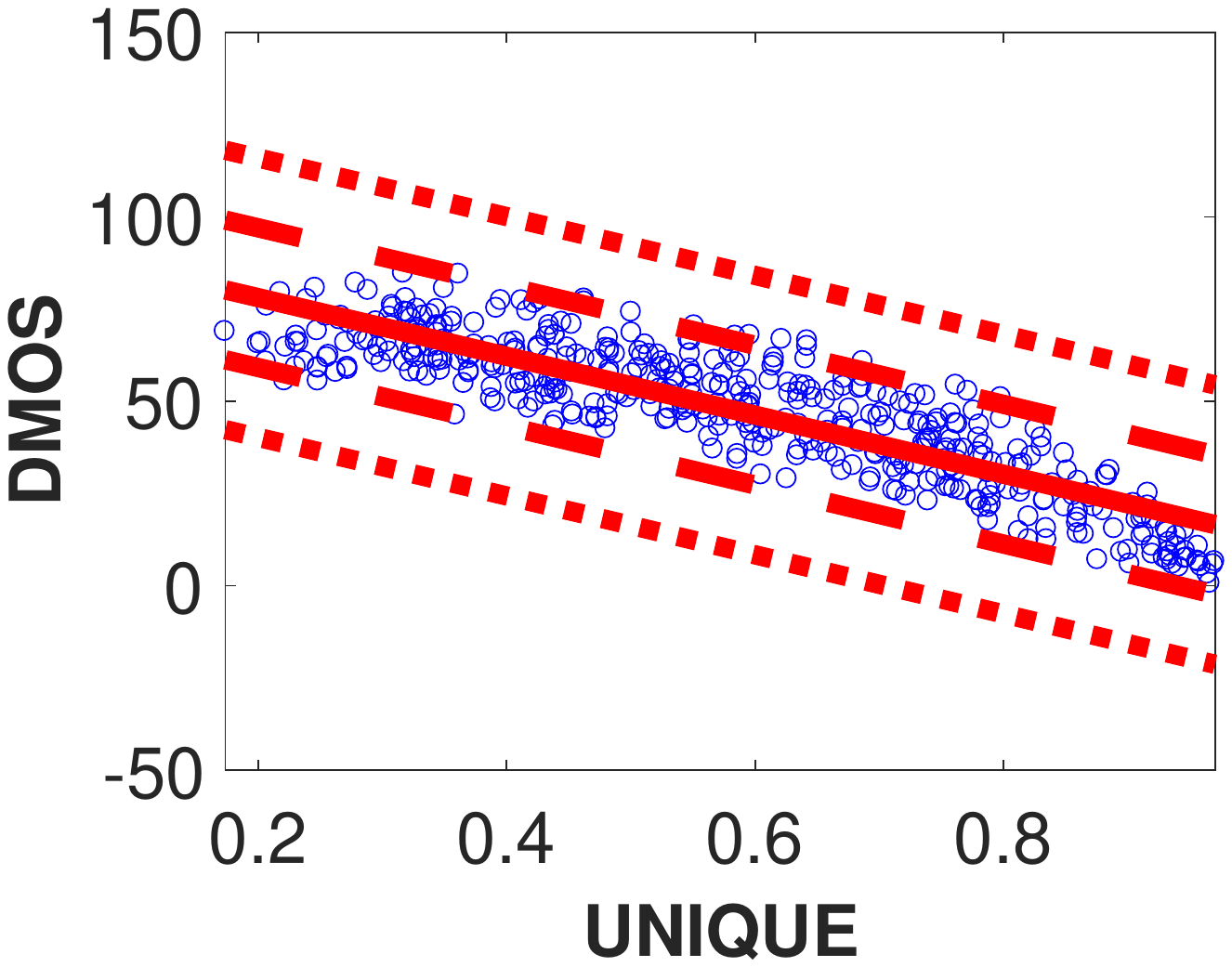}
  \vspace{0.03 cm}
  \centerline{\scriptsize{(k)MULTI:UNIQUE} }
\end{minipage}
\hfill 
 \vspace{0.05cm}
\begin{minipage}[b]{0.31\linewidth}
  \centering
\includegraphics[width=\linewidth, trim= 40mm 90mm 40mm 90mm]{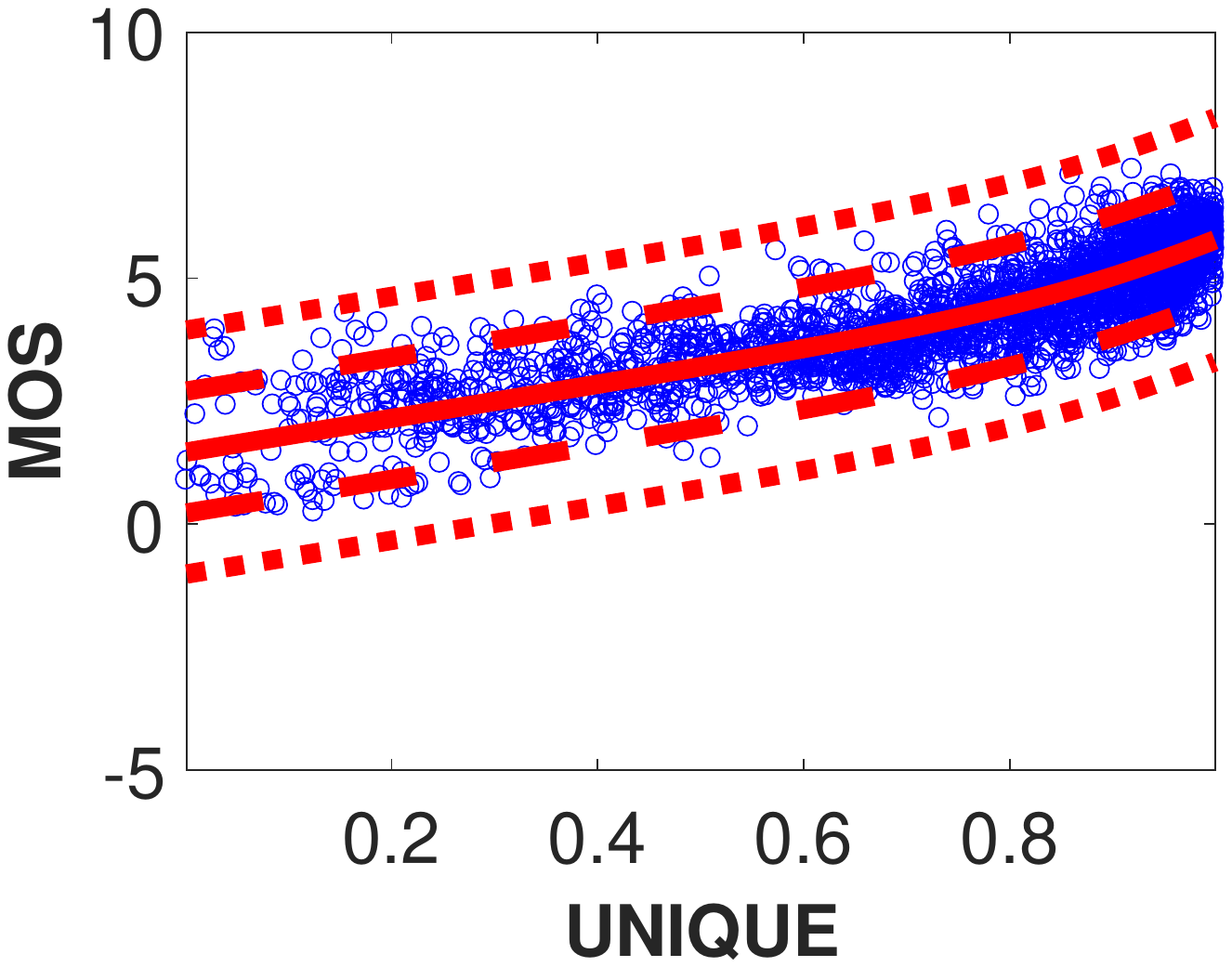}
  \vspace{0.03 cm}
  \centerline{\scriptsize{(l)TID13:UNIQUE}}
\end{minipage}
 \vspace{0.05cm}
 \begin{minipage}[b]{0.31\linewidth}
  \centering
\includegraphics[width=\linewidth, trim= 40mm 90mm 40mm 90mm]{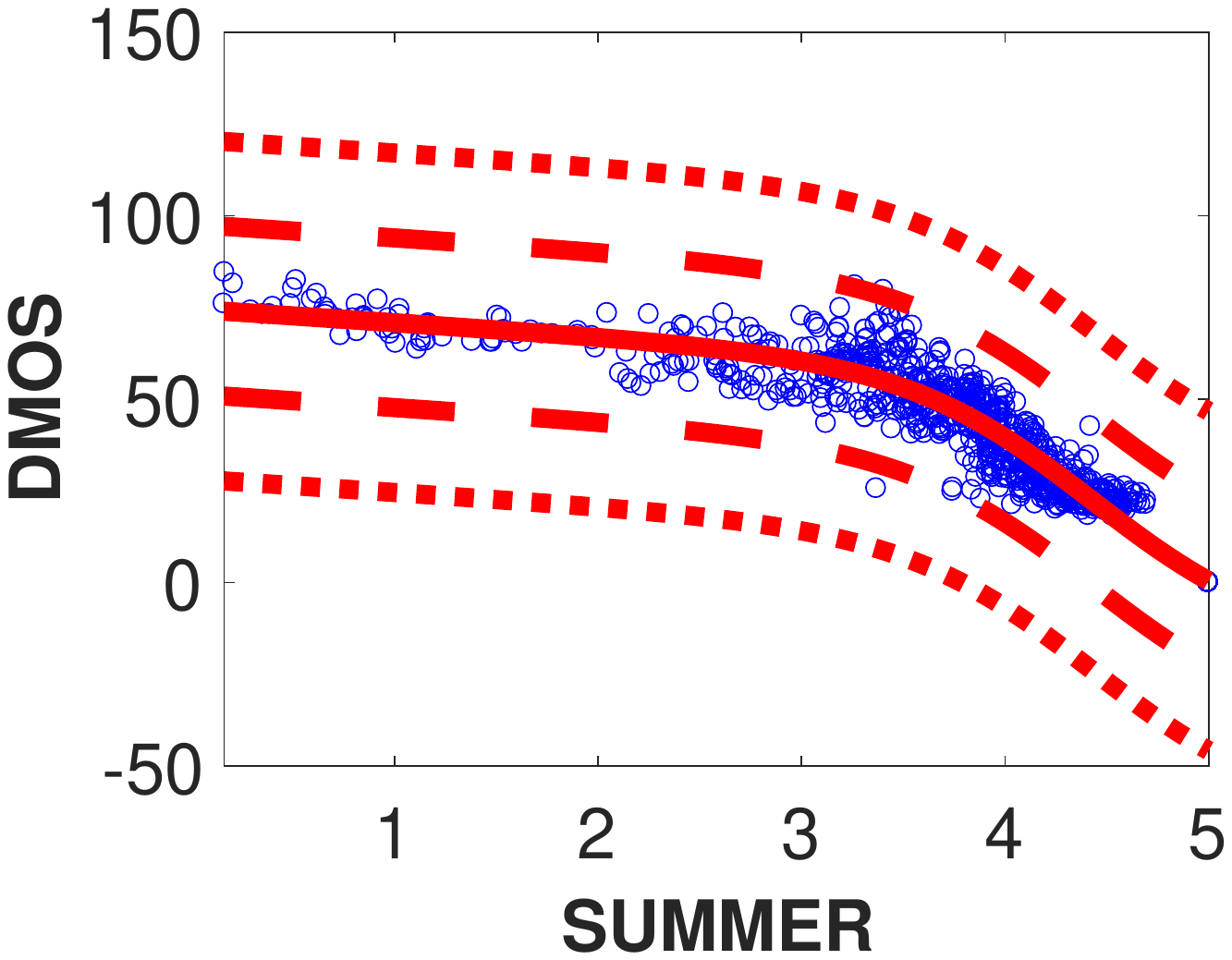}
  \vspace{0.03 cm}
  \centerline{\scriptsize{(m)LIVE:SUMMER} }
\end{minipage}
\hfill
 \vspace{0.05cm}
\begin{minipage}[b]{0.31\linewidth}
  \centering
\includegraphics[width=\linewidth, trim= 40mm 90mm 40mm 90mm]{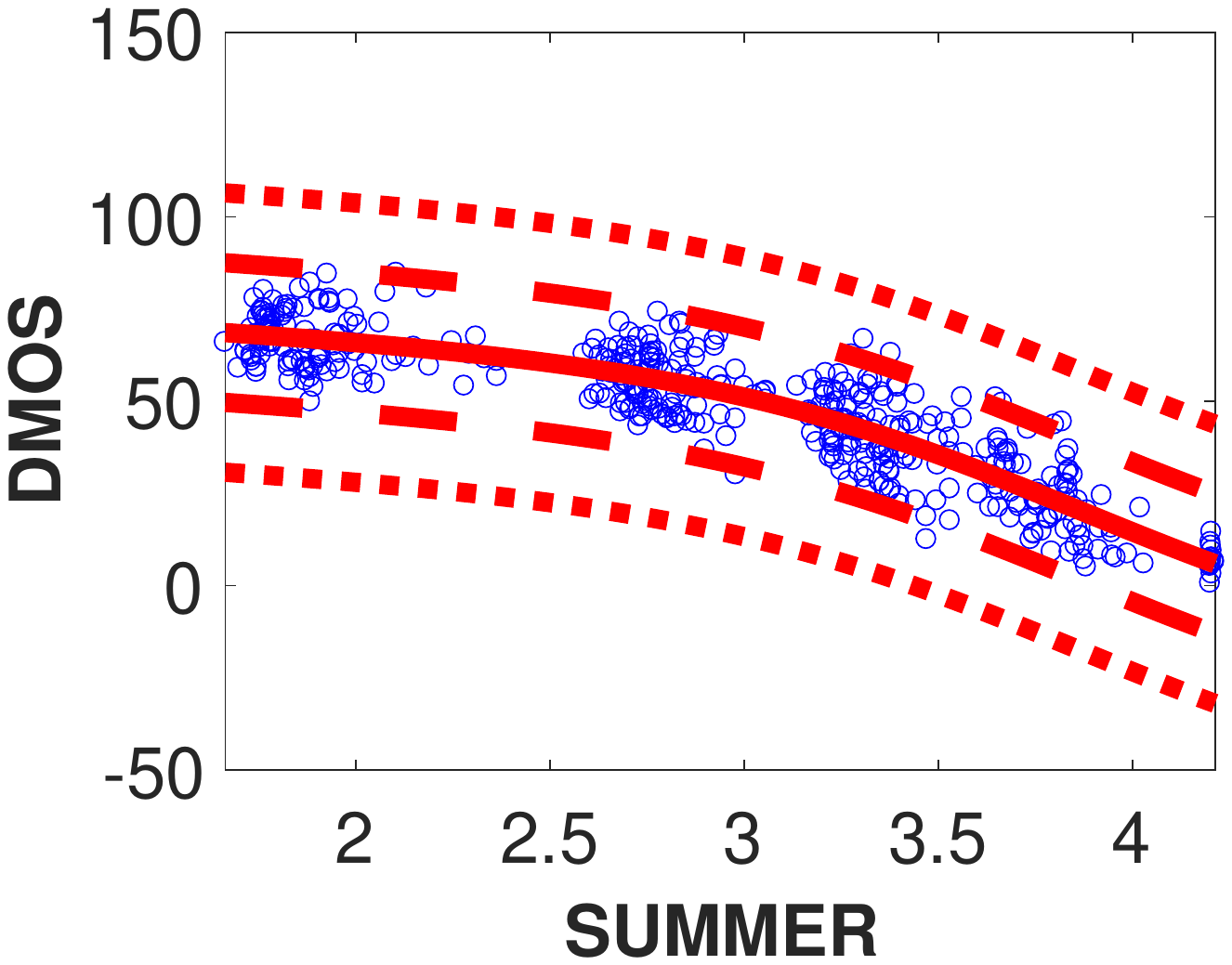}
  \vspace{0.03 cm}
  \centerline{\scriptsize{(n)MULTI:SUMMER} }
\end{minipage}
\hfill 
 \vspace{0.05cm}
\begin{minipage}[b]{0.31\linewidth}
  \centering
\includegraphics[width=\linewidth, trim= 40mm 90mm 40mm 90mm]{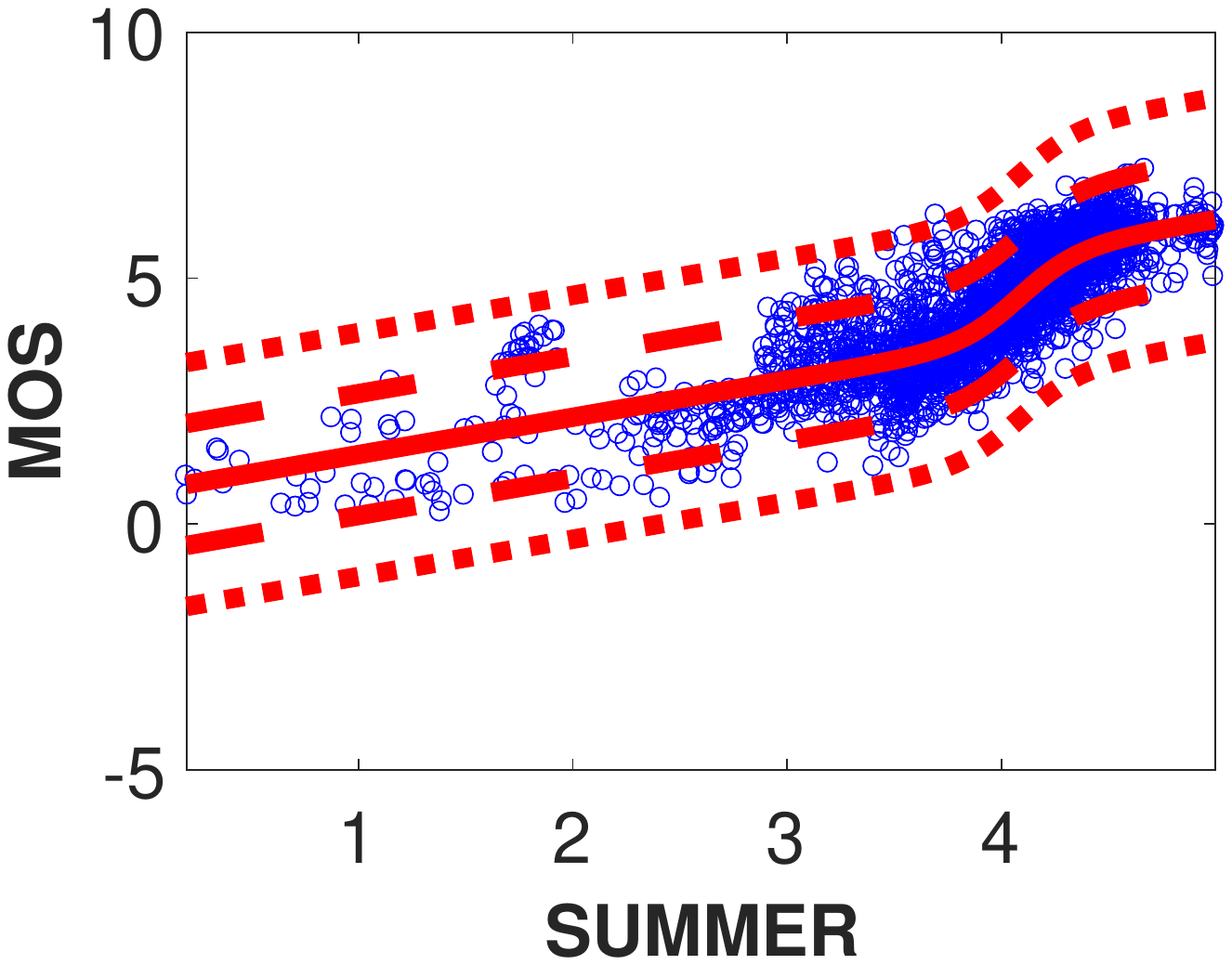}
  \vspace{0.03 cm}
  \centerline{\scriptsize{(o)TID13:SUMMER}}
\end{minipage}
\vspace{-0.5cm} 
 \caption{Scatter plots of best performing quality estimators.}
\label{fig:Scatter_all}
\end{figure}

\subsection{Distributional Difference and Scatter Plots}
\label{subsec:dist_scatter}
To analyze the difference between subjective and objective scores, we calculated the difference between normalized histograms of subjective scores and quality estimates as reported in Table \ref{tab_hist_dist}. Distribution of subjective scores is most similar to \texttt{SUMMER} and CSV in the LIVE and the TID13 databases and to UNIQUE and CSV in the MULTI database. To further analyze the relationship between subjective and objective scores, we selected quality estimators that are highlighted in at least two different databases in Table \ref{tab_results_all}, which include IW-SSIM, SR-SIM, CSV, UNIQUE, and \texttt{SUMMER}. We show the scatter plots of top quality estimators in Fig. \ref{fig:Scatter_all}. In the provided scatter plots, the x-axis corresponds to the quality estimates and the y-axis corresponds to the mean opinion scores (MOS) or differential mean opinion scores (DMOS). We plot the mapping function that is learned by the regression formulation as a red curve in the scatter plots. Moreover, we also plot two curves that are one standard deviation away with dashed lines and two curves that are two standard deviation away with dotted lines. An ideal quality estimator should be located on a linear curve with low deviation. In the LIVE database, IW-SSIM and SR-SIM have a steeper decrease close to maximum quality score. UNIQUE scores are spread out and they decrease more linearly whereas \texttt{SUMMER} and  CSV decrease monotonically. In the MULTI database, all methods follow a monotonically decreasing behavior. Even though certain algorithms including UNIQUE and CSV cover majority of objective scores between their minimum and maximum values continuously, we can observe that \texttt{SUMMER} scores are more clustered around certain values. In the TID13 database, IW-SSIM and SR-SIM follow a monotonically increasing behavior along with a steeper increase close to maximum quality score. \texttt{SUMMER} also monotonically increases but its steep increase is around high scores rather than max score. CSV and UNIQUE follow a relatively linear behavior compared to other methods. CSV has a limited quality score range utilization whereas UNIQUE is spread all over its score range.

\begin{figure}[htbp!]
\begin{minipage}[b]{0.28\linewidth}
  \centering
\includegraphics[width=\linewidth]{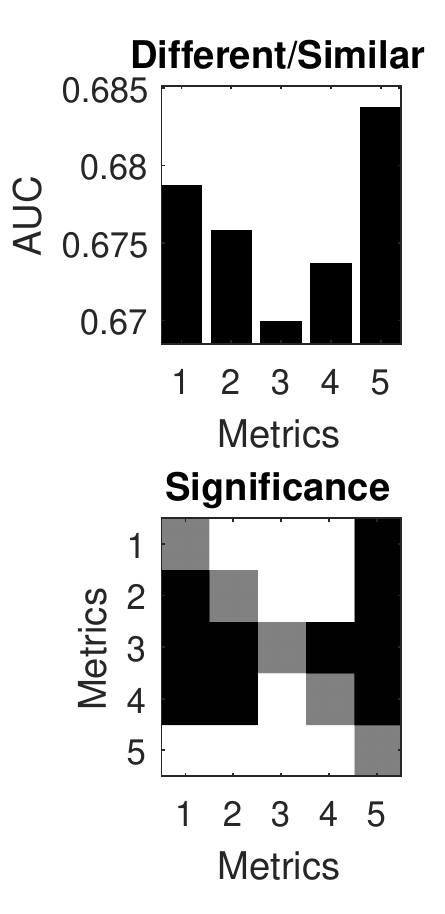}
  \vspace{-3.0 mm}
  \centerline{\scriptsize{(a) Different versus Similar (AUC)}}
\end{minipage}
 \vspace{0.05cm}
\hfill
\begin{minipage}[b]{0.28\linewidth}
  \centering
\includegraphics[width=\linewidth]{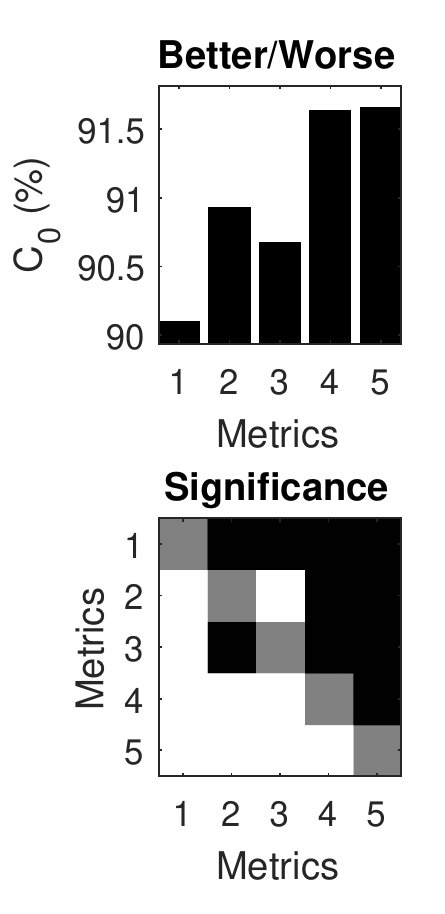}
  \vspace{-3.0 mm}
  \centerline{\scriptsize{(b)Better versus Worse ($C_0$)} }
\end{minipage}
 \vspace{0.05cm}
\hfill
\begin{minipage}[b]{0.28\linewidth}
  \centering
\includegraphics[width=\linewidth]{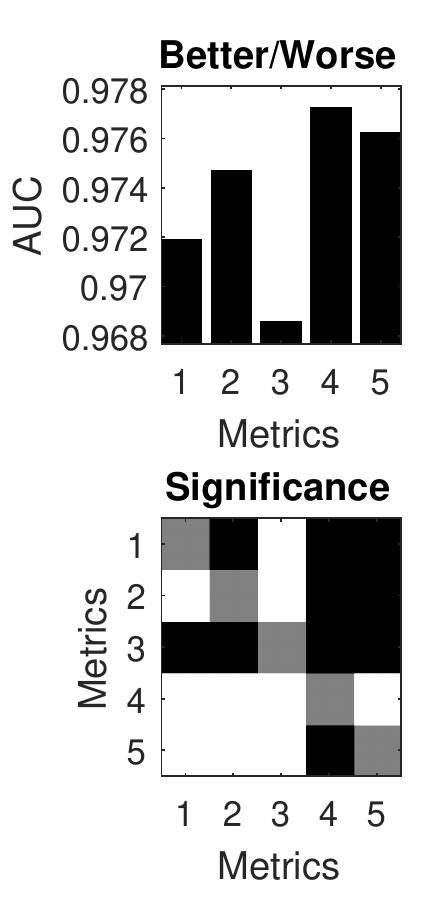}
  \vspace{-3.0 mm}
  \centerline{\scriptsize{(c) Better versus Worse (AUC)} }
\end{minipage}
\vspace{-1.0 mm}
\caption{The results of the statistical analysis tests for all three databases combined. Significance plots at the bottom show that the performance of the method in the row is either significantly better (white),worse (black), or equivalent (gray). Metric indices correspond to IW-SSIM (1), SR-SIM (2), CSV (3), UNIQUE (4), and SUMMER (5).}
\label{fig:more_stat_analysis}
\end{figure}

\subsection{Classification Performance}
\label{subsec:stat_analysis}
Previously, we tested the estimation performance of the image quality assessment algorithms. In this section, we test the classification performance of these assessment algorithms by utilizing the techniques introduced in \cite{Krasula2016}. The first analysis measures the capability of quality assessment algorithms to distinguish statistically different and similar pairs (Different versus Similar test). Absolute difference of the predicted scores should be larger for significantly different image pairs to achieve a high performance. We report the performance in terms of the area under Receiver Operating Characteristics Curve ($AUC$). The second analysis is performed on the significantly different pairs to measure the capability of the algorithms to identify the higher quality image and the lower quality image (Better versus Worse test). In addition to $AUC$, we also report the result for the second analysis in terms of classification percentage ($C_0$). We also provide statistical significance results corresponding to all reported metrics. In Fig.~\ref{fig:more_stat_analysis}, we report the $AUC$ and $C_0$ values of five top-performing quality estimators in the top row and statistical significance comparison in the bottom row. These results correspond to performances over all the databases combined. We observe that proposed method \texttt{SUMMER} significantly outperforms compared methods in almost all of the categories other than UNIQUE in the $AUC$ category.

\subsection{Computation Time}
\label{subsec:comp_time}
We measured the time required to obtain objective quality scores of all the images in the validation databases and computed the average processing time per image. In our analysis, we do not include the quality estimators that require an off-line training process. The computer used for these measurements has a 3.50 GHz Intel(R) Core(TM) i7-3770K CPU and a 32 GB  RAM. The average time per image for PSNR, SSIM, MS-SSIM, SR-SIM, and COHERENSI are all less than or equal to 0.05 seconds, which is followed by \texttt{SUMMER} with $0.07$ seconds. The average time required to obtain a quality score with other methods varies between $3$ to $26$ times of the time required by \texttt{SUMMER.} In its current implementation, spectral analysis and frequency-based weight extraction are performed over each color channel and scale sequentially. Therefore, we can reduce the computation time with a more efficient implementation that supports parallel processing of color channels and scales.


\section{Conclusion}
\label{sec:conclusion}
We analyzed the magnitude spectrum of error signals and extended this analysis with color channel utilization, multi-resolution representation, and  frequency-based weight extraction to obtain the quality assessment algorithm \texttt{SUMMER}. Based on our experiments, the proposed algorithm significantly outperforms the majority of compared image quality assessment algorithms. As shown in the validation, the relationship between objective and subjective scores is monotonic rather than linear. Therefore, to utilize \texttt{SUMMER} in practice for any kind of stimuli without regression, we need to enhance the algorithm to provide higher linearity and lower deviation. Color channel utilization contributed to the performance enhancement of \texttt{SUMMER} but we need to design bio-inspired algorithms that rely on visual system characteristics rather than solely depending on color channel values. In this study, we utilized the mean value of magnitude spectrums to estimate the objective quality. However, in addition to global statistics, we need to investigate the shape-based characteristics of error spectrums. With the proposed framework, a shape-based spectral signature can be obtained to not only estimate the quality but also to identify the distortion types.




\begin{thebibliography}{10}
\expandafter\ifx\csname url\endcsname\relax
  \def\url#1{\texttt{#1}}\fi
\expandafter\ifx\csname urlprefix\endcsname\relax\def\urlprefix{URL }\fi
\expandafter\ifx\csname href\endcsname\relax
  \def\href#1#2{#2} \def\path#1{#1}\fi

\bibitem{WL+95}
S.~Westen, R.~Lagendijk, J.~Biemond, Perceptual image quality based on a
  multiple channel {{HVS}} model, in: International Conference on Acoustics,
  Speech, and Signal Processing, Detroit, MI, USA, 1995.

\bibitem{Hegazy2014}
T.~Hegazy, G.~AlRegib, {COHERENSI: A new full-reference IQA index using error
  spectrum chaos}, in: IEEE Global Conference on Signal and Information
  Processing, 2014, pp. 965--969.

\bibitem{Wang2004}
Z.~Wang, A.~C. Bovik, H.~R. Sheikh, E.~P. Simoncelli, {Image quality
  assessment: From error visibility to structural similarity.}, IEEE
  Transactions on Image Processing 13~(4) (2004) 600--12.

\bibitem{Wang2003}
Z.~Wang, E.~P. Simoncelli, A.~C. Bovik, {Multi-scale structural similarity for
  image quality assessment}, the Thirty-Seventh Asilomar Conference on Signals,
  Systems and Computers 2 (2004) 9--13.

\bibitem{Sampat2009}
M.~Sampat, Z.~Wang, S.~Gupta, A.~Bovik, M.~Markey, {Complex wavelet structural
  similarity: A new image similarity index}, Image Processing, IEEE
  Transactions on 18~(11) (2009) 2385--2401.

\bibitem{Wang2011}
Z.~Wang, Q.~Li, {Information content weighting for perceptual image quality
  assessment.}, IEEE Transactions on Image Processing 20~(5) (2011) 1185--98.

\bibitem{EA+06}
K.~Egiazarian, J.~Astola, N.~Ponomarenko, V.~Lukin, F.~Battisti, M.~Carli, New
  full-reference quality metrics based on {{HVS}}, in: International Workshop
  on Video Processing and Quality Metrics, Scottsdale, AZ, USA, 2006.

\bibitem{PS+07}
N.~Ponomarenko, F.~Silvestri, K.~Egiazarian, M.~Carli, J.~Astola, V.~Lukin, On
  between-coefficient contrast masking of {{DCT}} basis functions, in:
  International Workshop on Video Processing and Quality Metrics, Scottdale,
  AZ, USA, 2007.

\bibitem{Ponomarenko2011}
N.~Ponomarenko, O.~Eremeev, L.~V., K.~Egiazarian, M.~Carli, Modified image
  visual quality metrics for contrast change and mean shift accounting.

\bibitem{Daly1992}
S.~Daly, Digital images and human vision, MIT Press, Cambridge, MA, USA, 1993,
  Ch. The Visible Differences Predictor: An Algorithm for the Assessment of
  Image Fidelity, pp. 179--206.

\bibitem{Zhang12}
L.~Zhang, H.~Li, {SR-SIM: A Fast and high performance IQA index based on
  spectral residual}, in: Image Processing (ICIP), 2012 19th IEEE International
  Conference on, 2012, pp. 1473--1476.

\bibitem{Venkata2000}
N.~Damera-Venkata, T.~D. Kite, W.~S. Geisler, B.~L. Evans, A.~C. Bovik, {Image
  quality assessment based on a degradation model.}, IEEE Transactions on Image
  Processing 9~(4) (2000) 636--50.

\bibitem{Chandler2007}
D.~M. Chandler, S.~S. Hemami, {VSNR: A Wavelet-based visual signal-to-noise
  ratio for natural images.}, IEEE Transactions on Image Processing 16~(9)
  (2007) 2284--98.

\bibitem{NiL+08}
A.~Ninassi, O.~{Le Meur}, P.~{Le Callet}, D.~Barba, {On the performance of
  human visual system based image quality assessment metric using wavelet
  domain}, in: Society of Photo-Optical Instrumentation Engineers (SPIE)
  Conference Series, Vol. 6806, 2008.

\bibitem{NaL+12}
M.~Narwaria, W.~Lin, I.~V. McLoughlin, S.~Emmanuel, L.~T. Chia, Fourier
  transform-based scalable image quality measure, IEEE Transactions on Image
  Processing 21~(8) (2012) 3364--3377.

\bibitem{Lambrecht2001}
C.~J.~V. Lambrecht, {Vision models and applications to image and video
  processing }, Kluwer Academic Publishers, 2001.

\bibitem{temel_15_persim}
D.~Temel, G.~AlRegib, {PerSIM: Multi-resolution image quality assessment in the
  perceptually uniform color domain}, in: IEEE International Conference on
  Image Processing, 2015, pp. 1682--1686.

\bibitem{temel_16_csv}
D.~Temel, G.~AlRegib, {CSV: Image quality assessment based on color, structure
  and visual system }, Signal Processing: Image Communication 48 (2016) 92 --
  103.

\bibitem{Tang2011}
H.~Tang, N.~Joshi, A.~Kapoor, Learning a blind measure of perceptual image
  quality, in: IEEE Conference on Computer Vision and Pattern Recognition,
  2011, pp. 305--312.

\bibitem{Mittal2012}
A.~Mittal, A.~K. Moorthy, A.~C. Bovik, {No-reference image quality assessment
  in the spatial domain.}, {IEEE Transactions on Image Processing} 21~(12)
  (2012) 4695--708.

\bibitem{Moorthy2010}
A.~K. Moorthy, A.~C. Bovik, A two-step framework for constructing blind image
  quality indices, IEEE Signal Processing Letters 17~(5) (2010) 513--516.

\bibitem{Saad2012}
M.~A. Saad, A.~C. Bovik, C.~Charrier, {Blind image quality assessment: A
  Natural scene statistics approach in the DCT domain}, IEEE Transactions on
  Image Processing 21~(8) (2012) 3339--3352.

\bibitem{Temel_UNIQUE}
D.~Temel, M.~Prabhushankar, G.~AlRegib, {UNIQUE: Unsupervised image quality
  estimation}, IEEE Signal Processing Letters 23~(10) (2016) 1414--1418.

\bibitem{Kang2014}
L.~Kang, P.~Ye, Y.~Li, D.~Doermann, {Convolutional neural networks for
  no-reference image quality assessment}, IEEE Conference on Computer Vision
  and Pattern Recognition (2014) 1733-- 1740.

\bibitem{QT+11}
M.~Qadri, K.~Tan, M.~Ghanbari, The impact of spatial masking in image quality
  meters, Global Journal of Computer Science and Technology 11 (2011) 69--75.

\bibitem{AD81}
D.~Albrecht, R.~{{De Valois}}, Striate cortex responses to periodic patterns
  with and without the fundamental harmonics, Journal of Physiology 319 (1981)
  497--514.

\bibitem{AP+87}
H.~Alphei, D.~Püschel, A.~Kohlrausch, Temporal and spectral masking effects of
  harmonic complex tones, in: Audio Engineering Society Convention 82, 1987.

\bibitem{VANDERSCHAAF1996}
A.~van~der Schaaf, J.~van Hateren, Modelling the power spectra of natural
  images: Statistics and information, Vision Research 36~(17) (1996) 2759 --
  2770.

\bibitem{Torralba2003}
A.~Torralba, A.~Oliva, Statistics of natural image categories, Network:
  Computation in Neural Systems 14~(3) (2003) 391--412, pMID: 12938764.

\bibitem{tid13}
N.~Ponomarenko, L.~Jin, O.~Ieremeiev, V.~Lukin, K.~Egiazarian, J.~Astola,
  B.~Vozel, K.~Chehdi, M.~Carli, F.~Battisti, C.-C.~J. Kuo, {Image database
  TID2013: Peculiarities, results and perspectives }, Signal Processing: Image
  Communication 30 (2015) 57 -- 77.

\bibitem{Sheikh2006b}
H.~R. Sheikh, M.~F. Sabir, A.~C. Bovik, A statistical evaluation of recent full
  reference image quality assessment algorithms, IEEE Transactions on Image
  Processing 15~(11) (2006) 3440--3451.

\bibitem{Jayaraman2012}
D.~Jayaraman, A.~Mittal, A.~K. Moorthy, A.~C. Bovik, {Objective quality
  assessment of multiply distorted images}, in: Asilomar Conference on Signals,
  Systems and Computers, 2012, pp. 1693--1697.

\bibitem{ITU-T}
\relax Telecommunication Standardization Sector~of International
  Telecommunication Union (ITU-T), {Methods, metrics and procedures for
  statistical evalution, qualification and comparison of objective quality
  prediction models}.

\bibitem{Kendal1945}
M.~Kendall, The advanced theory of statistics, in: Charles Griffin \& Company
  Limited, London, UK, 1945.

\bibitem{Zhang2011}
L.~Zhang, L.~Zhang, X.~Mou, D.~Zhang, {FSIM: A Feature similarity index for
  image quality qssessment.}, IEEE Transactions on Image Processing 20~(8)
  (2011) 2378--86.

\bibitem{Krasula2016}
L.~Krasula, K.~Fliegel, P.~L. Callet, M.~Klíma, On the accuracy of objective
  image and video quality models: New methodology for performance evaluation,
  in: International Conference on Quality of Multimedia Experience, 2016, pp.
  1--6.

\end{thebibliography}


\end{document}